\pdfoutput=1

\documentclass[10pt,twocolumn,letterpaper]{article}

\usepackage[pagenumbers]{wacv} 

\usepackage{graphicx}
\usepackage{makecell}
\usepackage{amsmath}
\usepackage{amssymb}
\usepackage{booktabs}

\usepackage{amsfonts,bm}
\usepackage{amsthm}
\usepackage[inline]{enumitem}
\usepackage{algorithm}
\usepackage{placeins}

\def\eqref#1{equation~\ref{#1}}

\def\1{\bm{1}}

\DeclareMathAlphabet{\mathsfit}{\encodingdefault}{\sfdefault}{m}{sl}
\SetMathAlphabet{\mathsfit}{bold}{\encodingdefault}{\sfdefault}{bx}{n}

\makeatletter
\@namedef{ver@everyshi.sty}{}
\makeatother
\usepackage{tikz}

\usepackage{pgfplots}
\usepackage{pgfplotstable}
\pgfplotsset{compat=1.7}
\usepackage{tikz}

\usepackage[export]{adjustbox}[2011/08/13]
\usepackage{dsfont}
\usepackage{mathtools}
\usepackage{caption}
\usepackage{subcaption}

\usepackage[pagebackref,breaklinks,colorlinks]{hyperref}

\usepackage[capitalize]{cleveref}
\crefname{section}{Sec.}{Secs.}
\Crefname{section}{Section}{Sections}
\Crefname{table}{Table}{Tables}
\crefname{table}{Tab.}{Tabs.}

\def\wacvPaperID{1056} 
\def\confName{WACV}
\def\confYear{2024}

\begin{document}

\title{Anchored Diffusion for Video Face Reenactment}

\author{
  Idan Kligvasser,
  Regev Cohen,
  George Leifman,
  Ehud Rivlin,
  and Michael Elad. \\ \\
  Verily AI
}
\maketitle

\begin{figure*}[ht!]
	\centering
	\captionsetup[subfigure]{labelformat=empty,justification=centering,aboveskip=1pt,belowskip=1pt}
    
    \begin{tabular}[c]{c c}
    
    \vspace{-2.5mm}
    
    \begin{subfigure}[t]{0.1\textwidth}
    \hspace{-.3mm}
	\centering
    \includegraphics[width=1\linewidth]{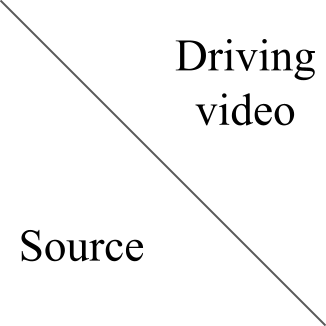}
	\end{subfigure}
	
	&
	
	\begin{subfigure}[t]{0.8\textwidth}
	\hspace{-.3mm}
    \centering
    \includegraphics[width=1\linewidth]{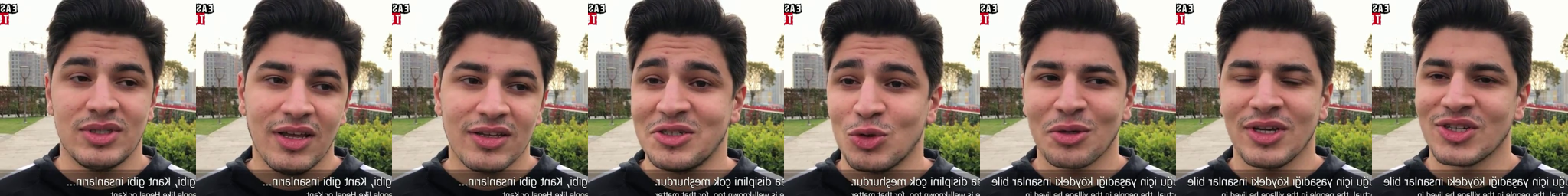}
	\end{subfigure}
	
	\\
    
    \vspace{-2.5mm}
    
    \begin{subfigure}[t]{0.1\textwidth}
    \hspace{-.3mm}
	\centering
    \includegraphics[width=1\linewidth]{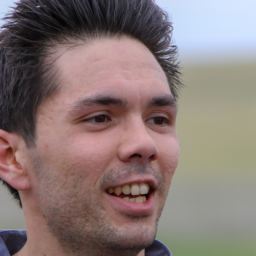}
	\end{subfigure}
	
	&
	
	\begin{subfigure}[t]{0.8\textwidth}
	\hspace{-.3mm}
    \centering
    \includegraphics[width=1\linewidth]{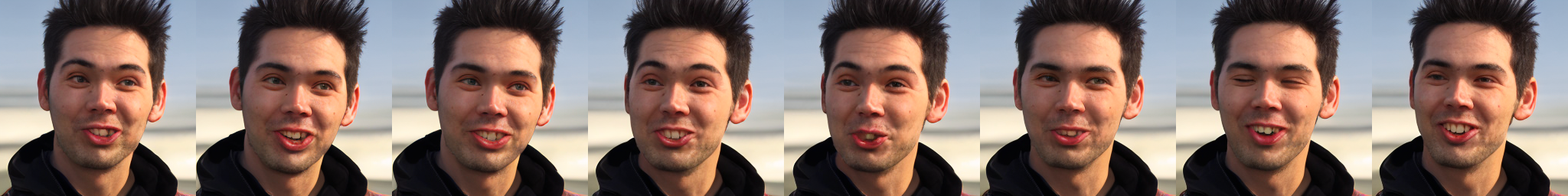}
	\end{subfigure}
	
	\\
	
	\vspace{-2.5mm}
	
	\begin{subfigure}[t]{0.1\textwidth}
    \hspace{-.3mm}
	\centering
    \includegraphics[width=1\linewidth]{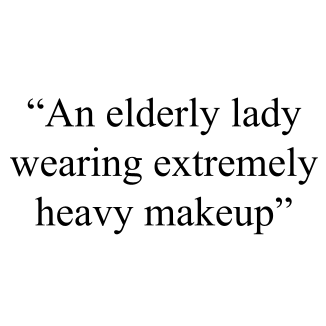}
	\end{subfigure}
	
	&
	
	\begin{subfigure}[t]{0.8\textwidth}
	\hspace{-.3mm}
    \centering
    \includegraphics[width=1\linewidth]{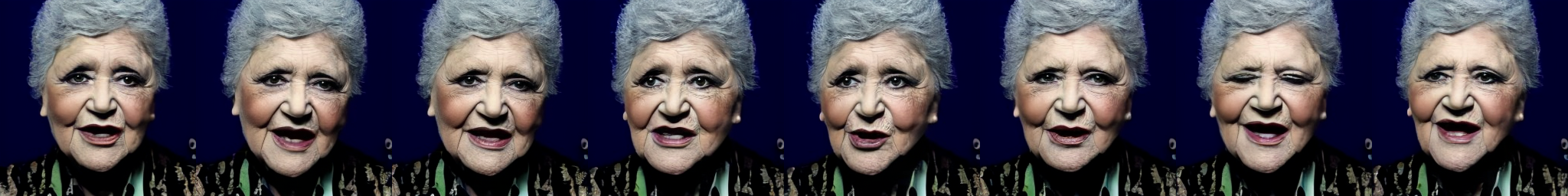}
	\end{subfigure}
	
	\\
	
	\vspace{-2.5mm}
	
	\begin{subfigure}[t]{0.1\textwidth}
    \hspace{-.3mm}
	\centering
    \includegraphics[width=1\linewidth]{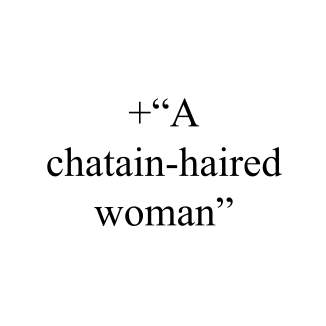}
	\end{subfigure}
	
	&
	
	\begin{subfigure}[t]{0.8\textwidth}
	\hspace{-.3mm}
    \centering
    \includegraphics[width=1\linewidth]{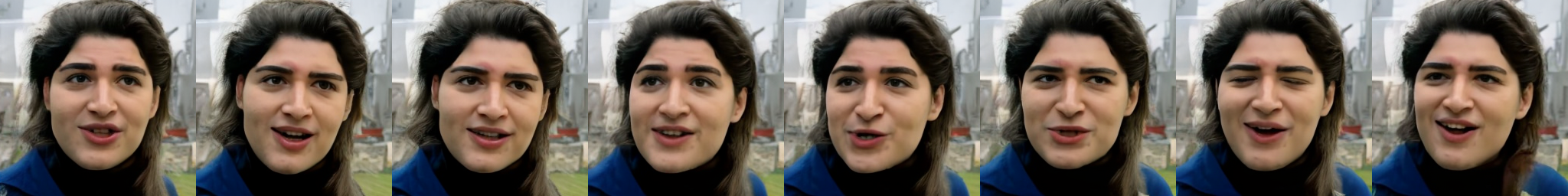}
	\end{subfigure}
	
	\\

	\end{tabular}
	
\caption{Sample results generated by \textit{Anchored Diffusion} for face reenactment given a driving video (top row), including image-to-video generation (second row), text-to-video generation (third row), and video editing (bottom row).}
\label{fig:figure1}
\end{figure*}

\begin{abstract}
Video generation has drawn significant interest recently, pushing the development of large-scale models capable of producing realistic videos with coherent motion. Due to memory constraints, these models typically generate short video segments that are then combined into long videos. The merging process poses a significant challenge, as it requires ensuring smooth transitions and overall consistency. In this paper, we introduce \textit{Anchored Diffusion}, a novel method for synthesizing relatively long and seamless videos. We extend Diffusion Transformers (DiTs) to incorporate temporal information, creating our sequence-DiT (sDiT) model for generating short video segments. Unlike previous works, we train our model on video sequences with random non-uniform temporal spacing and incorporate temporal information via external guidance, increasing flexibility and allowing it to capture both short and long-term relationships. Furthermore, during inference, we leverage the transformer architecture to modify the diffusion process, generating a batch of non-uniform sequences anchored to a common frame, ensuring consistency regardless of temporal distance. To demonstrate our method, we focus on face reenactment, the task of creating a video from a source image that replicates the facial expressions and movements from a driving video. Through comprehensive experiments, we show our approach outperforms current techniques in producing longer consistent high-quality videos while offering editing capabilities.

\end{abstract}

\section{Introduction}
Generative models have made remarkable strides in image synthesis, showcasing their ability to produce high-quality and diverse visuals through learning from extensive datasets~\cite{po2024state}.
A natural extension of this success is video generation, a field that has gained increasing attention in recent research~\cite{bar2024lumiere, liu2024sora, huang2024vbench}. Yet, it presents unique challenges due to the added complexity of capturing motion, temporal coherence, and the increased memory and computational requirements associated with processing sequences of frames. 

A common strategy to reduce the memory burden is generating short video segments and then combining them into a longer sequence~\cite{bar2024lumiere}. However, seamlessly merging these segments is challenging, as misalignment and inconsistencies can introduce boundary artifacts and temporal drift, degrading the quality and naturalness of the generated video.

To address the challenge of generating long, coherent videos, we introduce \textit{Anchored Diffusion}, a novel diffusion-based method. Our approach leverages the scalability and long-range dependency capabilities of Diffusion Transformers (DiTs), extending them to incorporate temporal information and temporal positional encoding. This forms our fundamental building block, \textit{sequence-DiT} (sDiT), designed for generating short video segments.

In contrast to previous approaches, we train our model using non-uniform video sequences with varying temporal distances between frames. This encourages the model to capture both short and long-term temporal relationships. Additionally, we guide generation with global signals that determine overall frame structure and per-frame temporal signals that dictate interactions between frames.

To achieve long video generation, we modify the diffusion process during inference. By exploiting the batch dimension, we generate multiple non-uniform sequences of the same scene, all sharing a common "anchor" frame. Throughout the diffusion process, we enforce the consistency of the tokens corresponding to the anchor frame across all sequences. This ensures all generated frames align with the anchor, regardless of their temporal distance, resulting in long, coherent videos with smooth transitions.

We showcase our method through its application in neural face reenactment, a prominent area within computer vision with significant advancements in applications like virtual reality, video conferencing, and digital entertainment. The goal here is to create videos from a single source image that realistically mimic the expressions and movements of a driving video, while preserving the source's identity. Current state-of-the-art methods~\cite{siarohin2019first, zhao2021sparse, zakharov2019few, hong2023implicit, agarwal2023audio, chung2018voxceleb2, karras2019style} often struggle with poor generalization and visual artifacts, particularly in extreme head poses or when the generated video length is significantly extended.

To address this, we first identified a lack of diverse, large-scale facial video datasets. In response, we curated a high-quality dataset of over 1M clips from over 53K identities, representing the largest publicly available facial video dataset to our knowledge. Leveraging this dataset and our novel anchored diffusion approach, we present a face reenactment method that mitigates artifacts, produces longer and more coherent videos, while offering versatile editing capabilities.
Comprehensive evaluations show that our approach outperforms current face reenactment techniques both qualitatively and quantitatively. By offering a robust solution to existing challenges, our work sets a new benchmark and opens up avenues for further research and applications in neural face reenactment and beyond.

\begin{figure*}[t!]
\centering
\includegraphics[width=2.0\columnwidth,center]{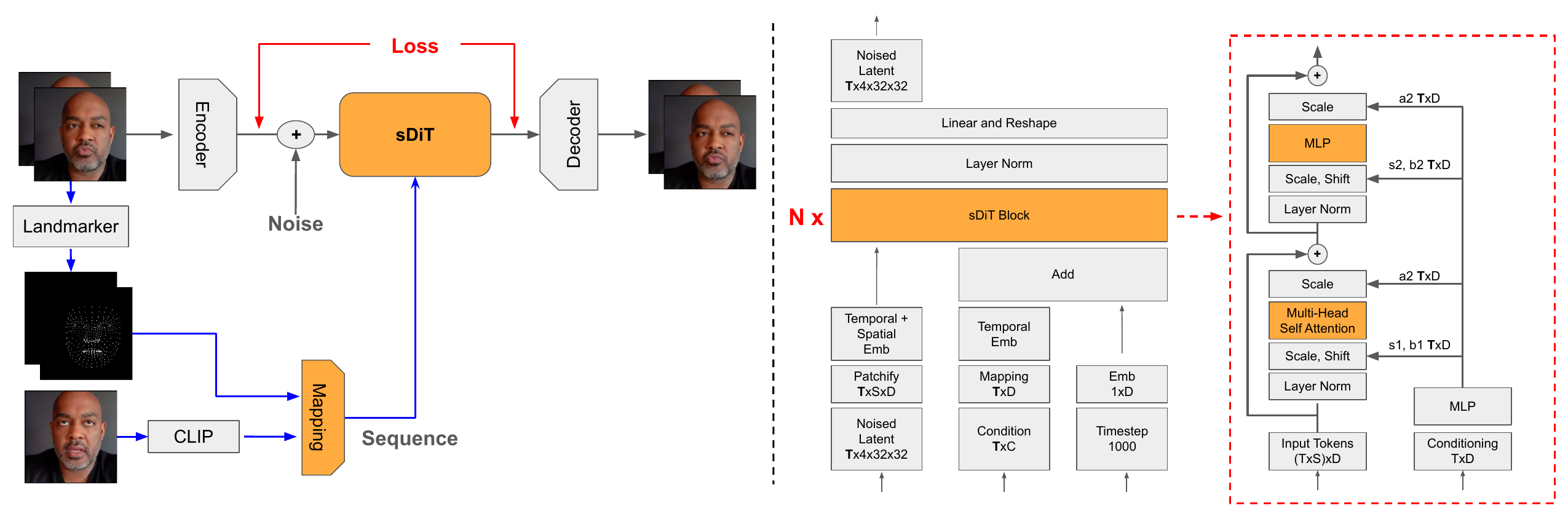}
\caption{\textbf{Scheme Overview.} \textit{Left}: Our video generation pipeline operates in latent space, where the sDiT denoiser is trained with per-frame guidance from CLIP embeddings and facial landmarks, using a weighted mean-square error loss to optimize the recovery of the driving video.
\textit{Right}: Our Sequence DiT (sDiT) architecture extends the DiT model for image generation to video generation by incorporating temporal dimensions and temporal positional encoding.}
\label{fig:architecture}
\end{figure*}

\section{Related Work}

\noindent \textbf{Video Generation with Diffusion Models.} Recently, substantial efforts have been made in training large-scale diffusion models on extensive datasets for video generation \cite{ho2022imagen, gupta2023photorealistic}, mostly using text guidance. A prominent approach for diffusion video generation involves "inflating" a pre-trained image model by adding temporal layers to its architecture and fine-tuning these layers, or optionally the entire model, on video data \cite{singer2022make, girdhar2023emu, yuan2024inflation}. VideoLDM \cite{blattmann2023align} and AnimateDiff \cite{guo2023animatediff} exemplify this approach by inflating StableDiffusion \cite{rombach2022high} and training only the newly-added temporal layers. The recent Lumiere \cite{bar2024lumiere} introduces a novel inflation scheme that includes learning to downsample and upsample the video in both space and time.
Our approach departs from these approaches. Instead of inflating existing models, we train our model from scratch using non-uniform video sequences from our newly curated dataset. Furthermore, we introduce temporal information through external signals that guide the diffusion process, offering increased flexibility. Finally, we propose a novel strategy for combining multiple sequences into one long, coherent video.

\noindent \textbf{Face Reenactment.}  Recent advancements in neural face reenactment have primarily employed image-driven strategies, which aim to capture expressions from a driving image and combine them with the identity from a source image. Several techniques \cite{wu2021f3a, yao2020mesh} utilize a 3D facial prior model to extract expression and identity codes from different faces to generate new ones. Other approaches \cite{zakharov2019few, zhao2021sparse} leverage facial landmarks detected by a pretrained model as anchors to transfer motion flow from driving face videos. As this can lead to accumulated errors, some methods \cite{siarohin2019first, wang2021one, zhao2022thin} have learned key points in an unsupervised manner, enhancing the representation of facial motion. In \cite{hong2023implicit}, the authors improve the quality of generation in ambiguous facial regions by using a memory-bank network. Despite these advances, these methods often struggle with cross-subject reenactment because facial landmarks retain the facial shape and identity geometry of the target face. To overcome these limitations, few works have adopted audio-driven strategies, as audio sequences lack facial identity information. Liang \textit{et al.} \cite{liang2022expressive} divide driving audio into characteristic root parts to precisely control lip shape, face pose, and facial expression. Agarwal \textit{et al.} \cite{agarwal2023audio} successfully employ both image-driven and audio-driven strategies, resulting in improved outcomes by leveraging the advantages of each approach. Despite these advancements, most existing approaches \cite{siarohin2019first, hong2022depth,hong2023implicit,bounareli2023hyperreenact,agarwal2023audio} rely on the Generative Adversarial Networks (GANs) framework for generation. GAN-based models often struggle to produce high-fidelity outputs when faced with limited training datasets or extreme head poses and extended video sequences.
In this work, we adopt the recently emerged diffusion model approach as a robust alternative to GANs for generating high-quality images and videos. Unlike GANs, diffusion models iteratively refine noisy images to create realistic outputs, offering more stable training dynamics, higher fidelity results and efficient editing capabilities.

\section{Method}
At the core of our work lie Diffusion Denoising Probabilistic Models~\cite{ho2020denoising, song2020denoising, song2020score, nichol2021improved, dhariwal2021diffusion, varshavsky2024semnatic}, which generate samples from a desired data distribution by iteratively refining random Gaussian noise until it transforms into a clean sample from the target distribution. These models can leverage side information, such as text prompts or  segmentation maps, to guide the generation and ensure the output aligns with the specified conditions. We specifically utilize Diffusion Transformers (DiTs), a class of of diffusion models that leverage the Transformer architecture, renowned for its scalability and ability to capture long-range dependencies, making them ideal choice for video generation.  

Our framework employs a diffusion transformer trained to generate short video sequences consisting of multiple frames. This generation process is guided by both global signals that dictate the high-level structure of the sequence and per-frame temporal information to ensure smooth, coherent transitions across the entire sequence. To generate long, temporally consistent videos at inference time, we leverage our model to produce a batch of video sequences of the same scene linked by a modified diffusion mechanism.  This mechanism, which we term \textit{anchored diffusion}, aligns and guides all generated sequences using the first sampled sequence as a reference. In the following sections, we provide an overview of our framework, detailing key architectural decisions, the training process, and our anchored inference diffusion approach.

\subsection{Guided Sequence-Diffusion (sDiT)}
Our model architecture is designed for generating video sequences $S=[F_1,F_2,\dots,F_T]$ of $T$ frames. We build upon the Diffusion Transformer (DiT) architecture~\cite{peebles2023scalable} which operates on sequences of spatial patches in latent space. The pre-trained AutoencoderKL~\cite{rombach2022high, kingma2013auto} is employed to encode input frames into this latent space and decode the output tokens back into pixel space. We extend DiTs to incorporate a temporal dimension of size $T$, as illustrated in Fig.~\ref{fig:architecture}. Furthermore, we introduce sinusoidal temporal positional encoding $[TPE_1,\dots.TPE_T]$ to facilitate the model's understanding of temporal order and relationships within the video sequence. The resulting block, termed sequence DiT (sDiT), serves as our core denoising model and it retains the scalability and efficiency of DiTs while extending their capabilities to the temporal domain. 

We guide the generation process with two types of control signals: global signals ($G$) determining the overall spatial structure of each frame, and local per-frame signals ($L_t$) dictating temporal relationships between frames. These signals are mixed via a small \textit{mapping} network to produce per-frame conditioning signals $(C_t)$, incorporated into our sDiT through adaptive layer normalization. This ensures high perceptual quality in both spatial and temporal dimensions. Unlike previous work, our approach incorporates temporal information through conditioning, offering greater flexibility in defining temporal control signals.

In this work we focus on video face reenactment, namely, the task of transferring facial expressions and head movements from a driving video $V=[F_1,F_2,\dots,F_N]$ onto a single source image $I_s$, creating a video $V_s=[I_1,I_2,\dots,I_N]$ that reenacts the source image while preserving its identity. Our global control signal is the CLIP~\cite{radford2021learning} representation of the source image, chosen for its ability to capture both spatial and semantic information. This use of CLIP not only facilitates high-quality reenactment but also provides editing capabilities, as demonstrated later. Our per-frame temporal signals consist of facial landmarks extracted from the driving video using a pre-trained MediaPipe model~\cite{lugaresi2019mediapipe}.

At training, only the sDiT and the mapping network are updated, while the landmark model, CLIP, and the autoencoder remain frozen. To create a diffusion model, we train the entire system on a denoising task in latent space. In each training iteration, we randomly select a non-uniform sequence $S=[F_{t_1},F_{t_2},\dots,F_{t_T}]$ from a driving video within our dataset. For our source image we select an additional frame from the same video that is furthest in time from the chosen sequence, effectively making it either the first or last frame of the entire video. Employing non-uniform sequences and a temporally distant source frame encourages the model to learn both short- and long-range temporal relationships. Next, following standard diffusion model training, Gaussian noise is added to the driving video, and the model is trained to denoise it conditioned on the control signals. We employ a weighted mean-squared error (MSE) loss function, assigning higher weights to facial landmark regions in the images. This prioritizes the generation of coherent frames and smooth transitions within the sequence.

\subsection{Anchored Diffusion at Inference Time}
While our ultimate goal is generating long videos, memory constraints limit our sDiT model to producing short segments. Although leveraging batch processing enables the generation of multiple related sequences, combining these into a single coherent video remains a challenge. One approach, MultiDiffusion~\cite{bar2024lumiere, bar2023multidiffusion}, involves generating overlapping sequences at inference and averaging the overlapping frames to produce a longer video. However, while ensuring consistency between adjacent sequences, this method may not maintain coherence across  temporally distant sequences, potentially resulting in inconsistencies in the overall generated video.

\begin{figure*}[t!]
\centering
\includegraphics[width=1.85\columnwidth,center]{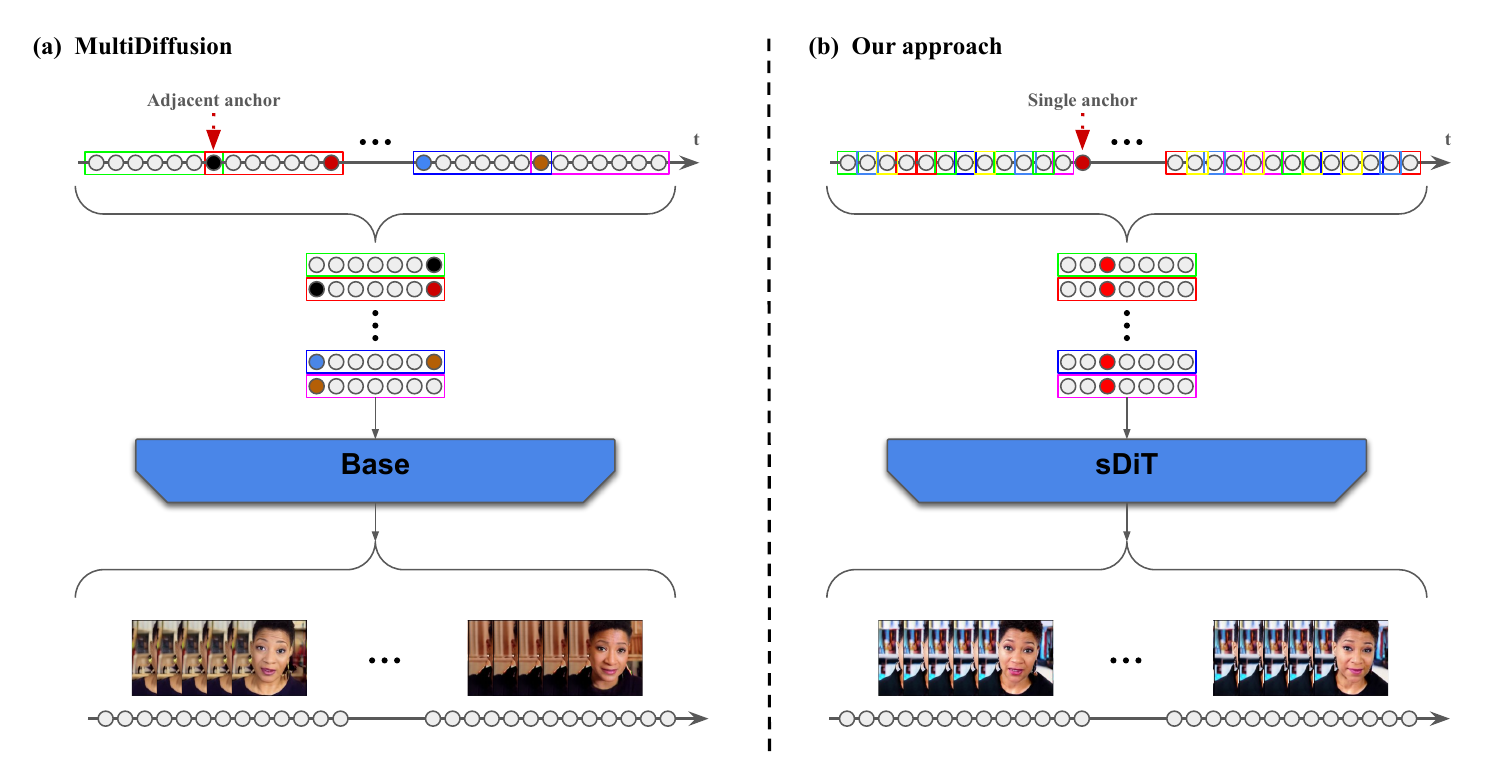}
\caption{\textbf{Anchored Diffusion.} We illustrate our strategy for merging multiple generated sequences into long videos, highlighting the main difference from a recent approach used in previous works. (a) Multidiffusion~\cite{bar2024lumiere, bar2023multidiffusion} generates multiple uniform sequences
with overlapping windows of adjacent anchor frames, achieving temporal consistency through averaging. (b) In contrast, our framework samples non-uniform sequences, with consistency between groups maintained by aligning all frames to a single frame shared across all groups.}
\label{fig:anchor}
\end{figure*}

\begin{figure*}
\centering
\captionsetup[subfigure]{labelformat=empty,justification=centering,aboveskip=1pt,belowskip=1pt}
    \begin{tabular}[c]{c c}
    \centering
    
    \vspace{-1.0mm}
    
    \begin{subfigure}[t]{0.15\textwidth}
    \centering
    \vspace{-0.75cm}
    \small{Driving video}
    \end{subfigure}
    
    &
    
    \hspace{-4.0mm}
    
    \begin{subfigure}[t]{0.825\textwidth}
    \centering
    \includegraphics[width=1\linewidth]{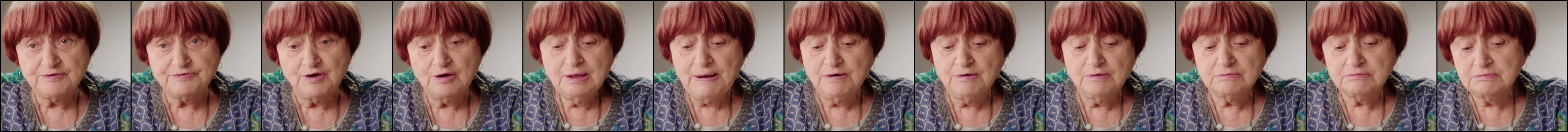}
    \end{subfigure}
    
    \\
    
    \vspace{-1.0mm}
    
    \begin{subfigure}[t]{0.15\textwidth}
    \centering
    \vspace{-0.75cm}
    \small{MultiDiffusion}
    \end{subfigure}
    
    &
    
    \hspace{-4.0mm}
    
    \begin{subfigure}[t]{0.825\textwidth}
    \centering
    \includegraphics[width=1\linewidth]{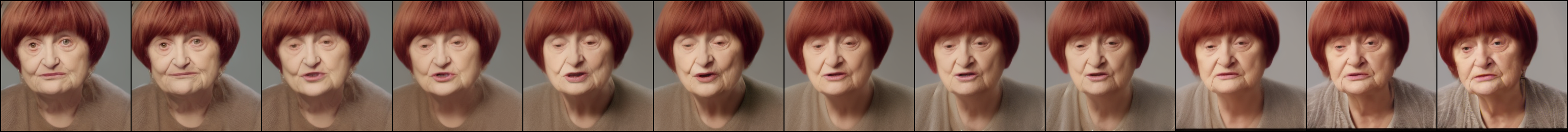}
    \end{subfigure}
    
    \\
    
    \begin{subfigure}[t]{0.15\textwidth}
    \centering
    \vspace{-0.75cm}
    \small{Anchored Diffusion}
    \end{subfigure}
    
    &
    
    \hspace{-4.0mm}
    
    \begin{subfigure}[t]{0.825\textwidth}
    \centering
    \includegraphics[width=1\linewidth]{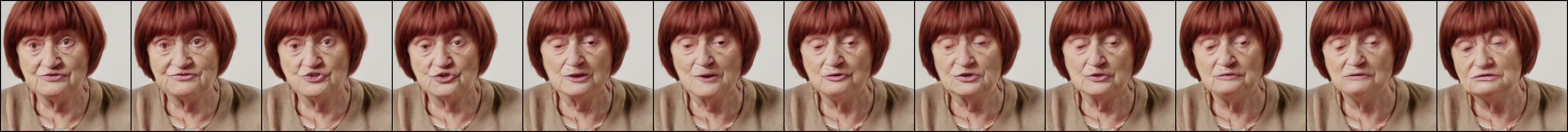}
    \end{subfigure}

    \end{tabular}
\caption{\textbf{Qualitative Consistency Comparison.} We use sDiT-XL model, capable of generating $4$ frames at once, to create a $12$-frame video. Multidiffusion fails to maintain consistency, as evident from the changing outfit of the person across the video. In contrast, our anchored diffusion demonstrates notable consistency throughout the video.}
\label{fig:drift}
\end{figure*}

\begin{figure}
\centering
\begin{tikzpicture}
\begin{axis}[
            xlabel={\# of frames},
            ylabel={Self-CSIM},
            grid=both,
            grid style={line width=.1pt, draw=gray!10},
            major grid style={line width=.2pt,draw=gray!50},
            width=0.7\columnwidth,
            height=0.525\linewidth,
            scale only axis=true,
            ylabel near ticks,
            xlabel near ticks,
            scaled ticks=true,
            legend pos=north west,
	        legend style={font=\small}
        ]
        
        \addplot[color=black, mark=*, line width=0.2mm,] coordinates{(12, 0.065)(16,0.078)(20,0.087)(24,0.104)(28,0.123)};
        \addplot[color=red, mark=*, line width=0.2mm,] coordinates{(12, 0.045)(16,0.062)(20,0.068)(24,0.073)(28,0.079)}; 
        
        \legend{MultiDiffusion, Anchored Diffusion};
        \end{axis}
    \end{tikzpicture}

\caption{\textbf{Consistency Evaluation.} Comparing our approach to Multidiffusion for generating long videos. We generated 50 self-reenactment videos per method and measured the average self cosine similarity (Self-CSIM), described in \ref{subsec:metrics}, between the generated and the driving video embeddings. Our method demonstrates superior consistency (lower values), with the margin further increasing as video length grows.}
\label{fig:consistency}
\end{figure}
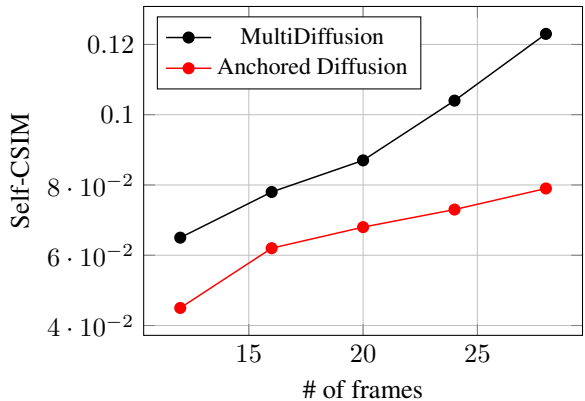

We propose an alternative method called anchored diffusion, depicted in Fig.~\ref{fig:anchor}. During inference, we first sample from the driving video a batch of non-uniform sequences with a shared \textit{anchor}, chosen as the central frame:
\begin{equation}
            \textbf{S}=\begin{pmatrix}
                S_1 \\
                S_2 \\
                \vdots \\
                S_B
            \end{pmatrix} = 
            \begin{pmatrix}
                F_{t_{11}} & \dots & F_{t_\texttt{anchor}} & \dots & F_{t_{1T}} \\
                F_{t_{21}} & \dots & F_{t_\texttt{anchor}} & \dots & F_{t_{2T}} \\
                \vdots & \ddots & F_{t_\texttt{anchor}} & \ddots & \vdots \\
                F_{t_{B1}} & \dots & F_{t_\texttt{anchor}} & \dots & F_{t_{BT}}
            \end{pmatrix}.
\end{equation}
We apply the landmark model on the above to yield a batch of our per-frame control signals. Then, we initiate the diffusion process by generating a batch of token sequences from pure noise:
\begin{equation}
            \textbf{Q}'=\begin{pmatrix}
                Q'_1 \\
                Q'_2 \\
                \vdots \\
                Q'_B
            \end{pmatrix} = 
            \begin{pmatrix}
                q_{t_{11}} & \dots & q_{1\texttt{anchor}} & \dots & q_{t_{1T}} \\
                q_{t_{21}} & \dots & q_{2\texttt{anchor}} & \dots & q_{t_{2T}} \\
                \vdots & \ddots & \vdots & \ddots & \vdots \\
                q_{t_{B1}} & \dots & q_{B\texttt{anchor}} & \dots & q_{t_{BT}}
            \end{pmatrix}.
\end{equation}
To enforce consistency of the central frame across all sequences, we override the corresponding tokens in other sequences with the tokens of the central frame in the first generated sequence at each diffusion step:
\begin{equation}
            \tilde{\textbf{Q}}=\begin{pmatrix}
                \tilde{Q}_1 \\
                \tilde{Q}_2 \\
                \vdots \\
                \tilde{Q}_B
            \end{pmatrix} = 
            \begin{pmatrix}
                q_{t_{11}} & \dots & \textcolor{blue}{q_{1\texttt{anchor}}} & \dots & q_{t_{1T}} \\
                q_{t_{21}} & \dots & \textcolor{blue}{q_{1\texttt{anchor}}} & \dots & q_{t_{2T}} \\
                \vdots & \ddots & \vdots & \ddots & \vdots \\
                q_{t_{B1}} & \dots & \textcolor{blue}{q_{1\texttt{anchor}}} & \dots & q_{t_{BT}}
            \end{pmatrix}.
            \label{eq:override}
\end{equation}
It is important to note that this override process is performed in all layers throughout the model architecture, ensuring consistency across all hierarchical representations of the video.
Thus, the anchor tokens are generated while attending to the other tokens within the first sequence of the batch. Due to the autoregressive nature of transformers, we can modify the diffusion mechanism to ensure all generated tokens attend to and align with the anchor tokens, regardless of their relative temporal distance. This promotes both short- and long-term consistency, as we show both qualitatively and quantitatively in Figures \ref{fig:drift} and \ref{fig:consistency} respectively.

Upon completion of the diffusion process, we decode the final output tokens and reorder the resulting frames chronologically to construct the final, seamless long video.
This technique, summarized in Algorithm~\ref{alg:anchor}, enables us to overcome memory constraints and generate extended videos with smooth transitions.

\begin{algorithm}
\caption{Anchored Diffusion for Face Reenactment}
\label{alg:anchor}
\small
\textbf{Input:} Source image $F_s$, Driving Video $V=[F_1,F_2,\dots,F_N]$.
\begin{enumerate}[itemindent=*]
    \item Sample non-uniform sequences with a shared anchor frame:
    $\textbf{S}\leftarrow\texttt{NonUniformSampling}(V)$
    \item Guidance Signals:
    \begin{itemize}
        \item $G\leftarrow\texttt{CLIP}(F_s)$.
        \item $\textbf{L} \leftarrow \texttt{FacialLandmark}(\textbf{S})$.
        \item $\textbf{C} \leftarrow \texttt{Mapping}(G,\textbf{L})$.
    \end{itemize}
    \item For each diffusion step $k$:
    \begin{itemize}
        \item $\textbf{Q}'_{k-1}\leftarrow\texttt{sDiT}(\textbf{C},\textbf{Q}_{k})$.
        \item $\tilde{\textbf{Q}}_{k-1}\leftarrow\texttt{Override}(\textbf{Q}'_{k-1})$ as in (\ref{eq:override}).
        \item $\textbf{Q}_{k-1}\leftarrow\texttt{DiffusionUpdateStep}(\tilde{\textbf{Q}}_{k-1})$.
    \end{itemize}
    \item $\textbf{F}_s\leftarrow\texttt{Decode}(\tilde{\textbf{Q}}_0)$.
    \item $V_s\leftarrow\texttt{Reorder}(\textbf{F}_s)$.
\end{enumerate}
\textbf{Output:} Generated Video $V_s$.
\end{algorithm}

{}

\subsection{Data Curation}
Previous works often extend pre-trained image generative models by adding layers to capture temporal information, training only these new components. In contrast, we train our sDiT model from scratch, thereby requiring a large-scale, high-quality video dataset.
In addition to utilizing small public video datasets such as CelebV-HQ~\cite{zhu2022celebv} and RAVDESS~\cite{livingstone2018ryerson}, we curated a novel, reproducible dataset through the following data collection process:
\begin{enumerate}[parsep=-2pt, leftmargin=*]
    \item \textbf{Query Creation}: We first retrieved videos with diverse facial content using queries that included YouTube channels and celebrity names.
    \item \textbf{Video Selection}: For each query, we selected only the top $20$ results and excluded videos with a resolution lower than $720p$.
    \item \textbf{Face Detection}: In each video, we performed face detection, extracting bounding boxes for stable detections larger than 400 pixels in segments longer than 2 seconds.
    \item \textbf{Segment Validation}: To ensure consistency, we measured the maximum ArcFace~\cite{deng2019arcface} and CLIP~\cite{radford2021learning} distances from the first frame to the rest of the video, discarding segments with significant distance variations.
\end{enumerate}

\noindent Through this extensive process, we curated a high-quality dataset, termed \textit{ReenactFaces-1M}, consisting of 1,006,257 video segments with an average length of 3.29 seconds and an average resolution of $745p$. For testing purposes, we excluded 1k videos from this dataset and added 1k randomly selected images from the FFHQ~\cite{karras2019style} dataset, specifically for the tasks of self and cross identity reenactment. We named this test set \textit{ReenactFaces-Test-1K}. Please refer the supplementary material for additional statistic information.

\section{Experiments}
We thoroughly evaluate our method across various applications, with ablation studies to justify design choices.

\subsection{Implementation Details}
We train two base sDiT-XL models, each with a patch size of 2x2, capable of generating sequences of $4$ and $8$ frames respectively. The mapping network consists of $4$ residual blocks. Following DiT's training scheme \cite{peebles2023scalable}, we train the models for $1$ million steps on our training dataset, using a batch size of 16 samples and the AdamW \cite{loshchilov2017decoupled} optimizer with a cosine learning rate scheduler starting at a base rate of $6.4 \cdot 10^{-5}$. To ensure better stability, the learning rate for the mapping network is set to be $10$ times smaller than the global learning rate.
We employ a weighted MSE training loss function to prioritize accurate reconstruction of facial expression, namely, facial landmarks around the mouth and eyes. These expressive landmark pixels are assigned a weight of $(1 + \lambda_{\text{ex}})$, setting $\lambda_{\text{ex}}=1$ specifically, while other pixels receive a weight of~$1$.
A detailed description of model components is provided in the supplementary material.

\subsection{Metrics}
\label{subsec:metrics}
We evaluate the performance of the examined algorithms across the following aspects:

\noindent\textbf{Fidelity.}
We measure generation realism using the Fréchet Inception Distance (FID)~\cite{heusel2017gans} and the Fréchet Video Distance (FVD)~\cite{unterthiner2019fvd}, standard metrics for generative models. Additionally, as FID tends not to capture distortion levels \cite{jayasumana2024rethinking}, we measure the generation quality using a non-reference image quality assessment model HyperIQA~\cite{su2020blindly}.  


\noindent\textbf{Motion.} We assess motion transfer by extracting $478$ XYZ facial points from the generated and driving videos using MediaPipe~\cite{lugaresi2019mediapipe}. We then compute the MSE between corresponding facial points (LMSE), and specifically for expressive points around the eyes and mouth (Expressive-LMSE). 

\noindent \textbf{Consistency.} To examine scene preservation across the video, we compute the minimum cosine similarity (CSIM) of the source and generated embeddings based on CLIP. This in contrast to previous works that rely on the ArcFace~\cite{deng2019arcface} embedding space, which is invariant to background, subject's haircut, outfit, etc. 
Lastly, we measure the consistency of the generated video in self-reenactment by computing distance between the minimum cosine similarity observed in the generated video embeddings and that observed in the corresponding driving video (Self-CSIM).

\subsection{Face Reenactment}
For the task of face reenactment, we compare our method with the state-of-the-art approaches. To ensure a fair comparison, we use the official pre-trained models of FOMM~\cite{siarohin2019first}, DaGAN~\cite{hong2022depth}, and MCNET~\cite{hong2023implicit}, sourced from their respective open-source implementations.

\subsubsection{Same-Identity Reenactment}
First, we perform a self-reenactment task where the source frame and the driving video are of the same person. Specifically, for each video, we select a random $8$-frame sequence in its original order as the driving video and the frame furthest in time from this sequence as the source image. This setup is considered straightforward as the source image already contains comprehensive information related to the desired generated video. Table~\ref{tab:quantitative_combined} presents quantitative results showing we outperform competing approaches in nearly all aspects for same-identity reenactment, significantly improving image quality while preserving fine motion.

\begin{table*}[ht]
\small
\centering
\begin{tabular}{l c c c c}
Metric & FOMM & DaGAN & MCNET & Ours \\ \hline
FID $\downarrow$             & $62.3 / 71.0$ & $56.4 / 59.9$ & $51.8 / 61.4$ & $\mathbf{38.2} / \mathbf{34.4}$ \\
FVD $\downarrow$             & $200 / 297$ & $163 / 287$ & $167 / 291$ & $\mathbf{111} / \mathbf{236}$ \\
HyperIQA $\uparrow$          & $37.9 / 36.5$ & $39.6 / 37.0$ & $38.9 / 37.3$ & $\mathbf{50.4} / \mathbf{52.1}$ \\ \hline
LMSE $\downarrow$            & $10.0 / 15.4$ & $11.0 / 14.8$ & $9.4 / 13.0$ & $\mathbf{8.92} / \mathbf{9.61}$ \\
Expressive-LMSE $\downarrow$ & $10.8 / 15.3$ & $12.3 / 16.4$ & $13.2 / 13.2$ & $\mathbf{9.07} / \mathbf{9.39}$ \\ \hline
CSIM $\uparrow$              & $0.74 / 0.58$ & $0.77 / 0.63$ & $0.78 / 0.58$ & $\mathbf{0.83} / \mathbf{0.74}$ \\
Self-CSIM $\downarrow$       & $\mathbf{0.03} / \mathbf{0.03}$ & $0.04 / 0.04$ & $\mathbf{0.03} / \mathbf{0.03}$ & $0.04 / 0.04$ \\ 
\end{tabular}
\caption{\textbf{Quantitative Results.} Comparisons with the competing methods \cite{siarohin2019first, hong2022depth, hong2023implicit} on the same- and cross-identity reenactment using our Records-Test-5K dataset. Metrics are marked with $\uparrow$ (higher is better)  or $\downarrow$ (lower is better), values are presented as x/y, representing same-identity and cross-identity reenactment results, respectively. 
Our method surpasses previous methods in nearly every aspect.}
\label{tab:quantitative_combined}
\end{table*}

\subsubsection{Cross-Identity Reenactment}
\begin{figure*}
	\centering
	\captionsetup[subfigure]{labelformat=empty,justification=centering,aboveskip=1pt,belowskip=1pt}
    
    \begin{tabular}[c]{c c}
    
    \vspace{-2.5mm}
    
    \begin{subfigure}[t]{0.1\textwidth}
    \hspace{-.3mm}
	\centering
    Source
	\end{subfigure}
	
	&
	
	\begin{subfigure}[t]{0.8\textwidth}
    \hspace{-.3mm}
	\centering
    Driving video
	\end{subfigure}
	
	\\
	
    \vspace{-2.5mm}
    
    \begin{subfigure}[t]{0.1\textwidth}
    \hspace{-.3mm}
	\centering
    \includegraphics[width=1\linewidth]{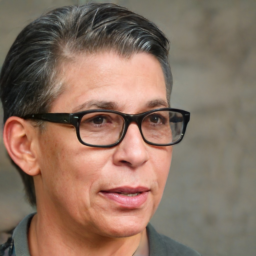}
	\end{subfigure}
	
	&
	
	\begin{subfigure}[t]{0.1\textwidth}
	\hspace{-.3mm}
    \centering
    \includegraphics[width=1\linewidth]{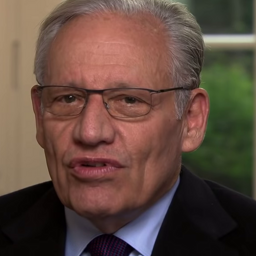}
	\end{subfigure}
	
	\begin{subfigure}[t]{0.1\textwidth}
	\hspace{-.3mm}
    \centering
    \includegraphics[width=1\linewidth]{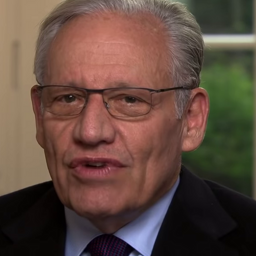}
	\end{subfigure}
	
	\begin{subfigure}[t]{0.1\textwidth}
	\hspace{-.3mm}
    \centering
    \includegraphics[width=1\linewidth]{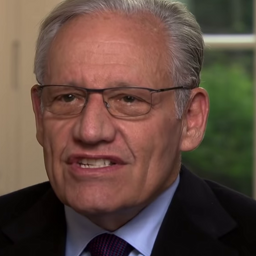}
	\end{subfigure}
	
	\begin{subfigure}[t]{0.1\textwidth}
	\hspace{-.3mm}
    \centering
    \includegraphics[width=1\linewidth]{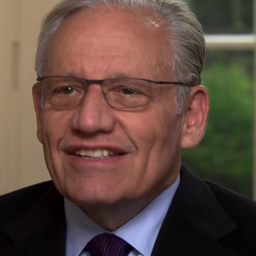}
	\end{subfigure}
	
	\begin{subfigure}[t]{0.1\textwidth}
	\hspace{-.3mm}
    \centering
    \includegraphics[width=1\linewidth]{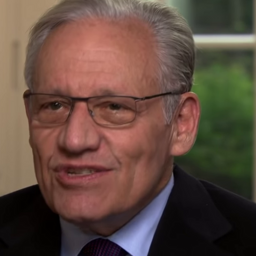}
	\end{subfigure}
	
	\begin{subfigure}[t]{0.1\textwidth}
	\hspace{-.3mm}
    \centering
    \includegraphics[width=1\linewidth]{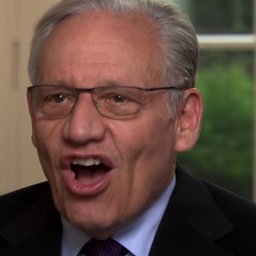}
	\end{subfigure}
	
	\begin{subfigure}[t]{0.1\textwidth}
	\hspace{-.3mm}
    \centering
    \includegraphics[width=1\linewidth]{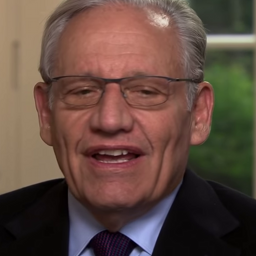}
	\end{subfigure}
	
	\begin{subfigure}[t]{0.1\textwidth}
	\hspace{-.3mm}
    \centering
    \includegraphics[width=1\linewidth]{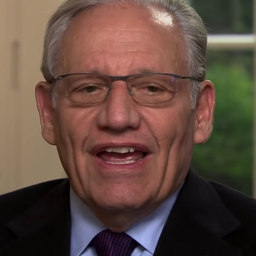}
	\end{subfigure}

	\\
	
	\vspace{-2.5mm}
	
	\begin{subfigure}[t]{0.1\textwidth}
    \hspace{-.3mm}
	\centering
    \includegraphics[width=1\linewidth]{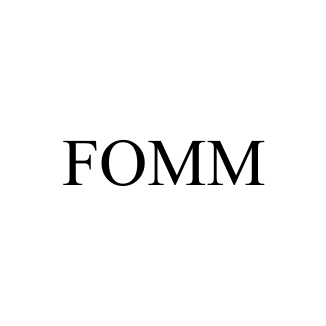}
	\end{subfigure}
	
	&
	
	\begin{subfigure}[t]{0.1\textwidth}
	\hspace{-.3mm}
    \centering
    \includegraphics[width=1\linewidth]{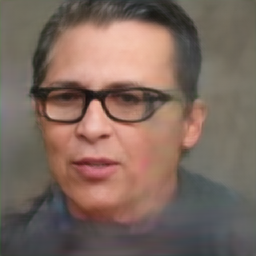}
	\end{subfigure}
	
	\begin{subfigure}[t]{0.1\textwidth}
	\hspace{-.3mm}
    \centering
    \includegraphics[width=1\linewidth]{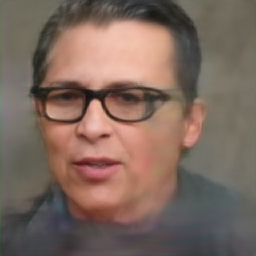}
	\end{subfigure}
	
	\begin{subfigure}[t]{0.1\textwidth}
	\hspace{-.3mm}
    \centering
    \includegraphics[width=1\linewidth]{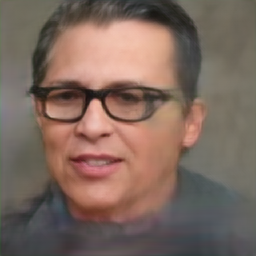}
	\end{subfigure}
	
	\begin{subfigure}[t]{0.1\textwidth}
	\hspace{-.3mm}
    \centering
    \includegraphics[width=1\linewidth]{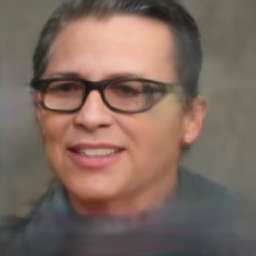}
	\end{subfigure}
	
	\begin{subfigure}[t]{0.1\textwidth}
	\hspace{-.3mm}
    \centering
    \includegraphics[width=1\linewidth]{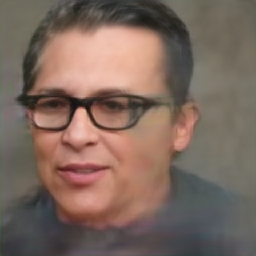}
	\end{subfigure}
	
	\begin{subfigure}[t]{0.1\textwidth}
	\hspace{-.3mm}
    \centering
    \includegraphics[width=1\linewidth]{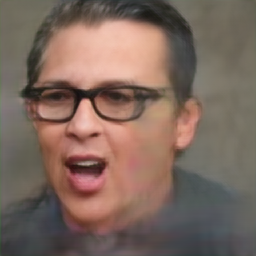}
	\end{subfigure}
	
	\begin{subfigure}[t]{0.1\textwidth}
	\hspace{-.3mm}
    \centering
    \includegraphics[width=1\linewidth]{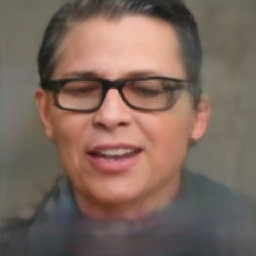}
	\end{subfigure}
	
	\begin{subfigure}[t]{0.1\textwidth}
	\hspace{-.3mm}
    \centering
    \includegraphics[width=1\linewidth]{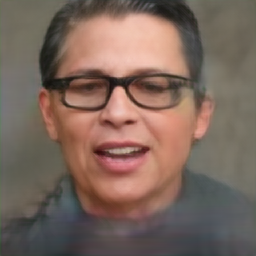}
	\end{subfigure}

	\\
	
	\vspace{-2.5mm}
	
	\begin{subfigure}[t]{0.1\textwidth}
    \hspace{-.3mm}
	\centering
    \includegraphics[width=1\linewidth]{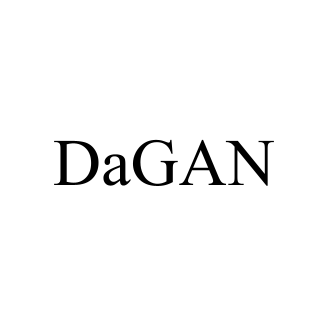}
	\end{subfigure}
	
	&
	
	\begin{subfigure}[t]{0.1\textwidth}
	\hspace{-.3mm}
    \centering
    \includegraphics[width=1\linewidth]{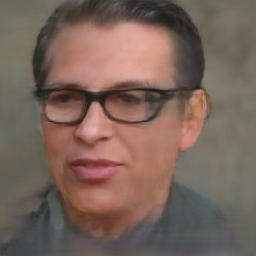}
	\end{subfigure}
	
	\begin{subfigure}[t]{0.1\textwidth}
	\hspace{-.3mm}
    \centering
    \includegraphics[width=1\linewidth]{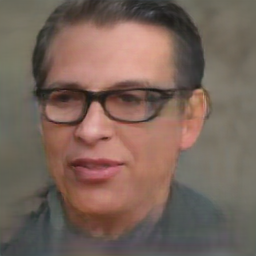}
	\end{subfigure}
	
	\begin{subfigure}[t]{0.1\textwidth}
	\hspace{-.3mm}
    \centering
    \includegraphics[width=1\linewidth]{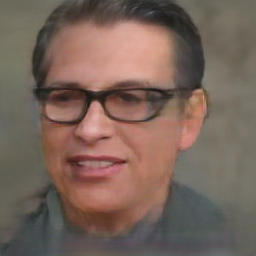}
	\end{subfigure}
	
	\begin{subfigure}[t]{0.1\textwidth}
	\hspace{-.3mm}
    \centering
    \includegraphics[width=1\linewidth]{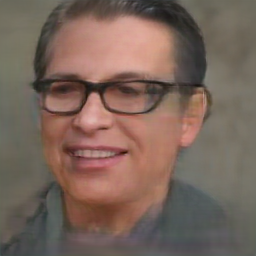}
	\end{subfigure}
	
	\begin{subfigure}[t]{0.1\textwidth}
	\hspace{-.3mm}
    \centering
    \includegraphics[width=1\linewidth]{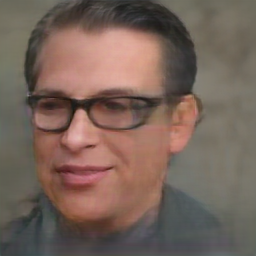}
	\end{subfigure}
	
	\begin{subfigure}[t]{0.1\textwidth}
	\hspace{-.3mm}
    \centering
    \includegraphics[width=1\linewidth]{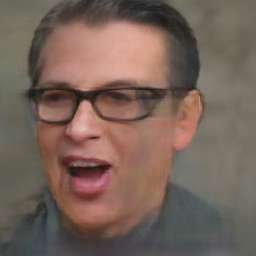}
	\end{subfigure}
	
	\begin{subfigure}[t]{0.1\textwidth}
	\hspace{-.3mm}
    \centering
    \includegraphics[width=1\linewidth]{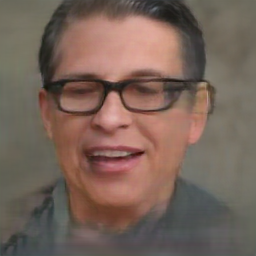}
	\end{subfigure}
	
	\begin{subfigure}[t]{0.1\textwidth}
	\hspace{-.3mm}
    \centering
    \includegraphics[width=1\linewidth]{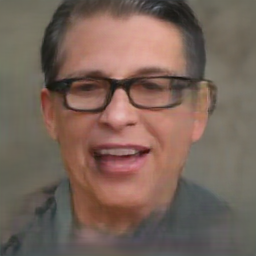}
	\end{subfigure}

	\\
	
	\vspace{-2.5mm}
	
	\begin{subfigure}[t]{0.1\textwidth}
    \hspace{-.3mm}
	\centering
    \includegraphics[width=1\linewidth]{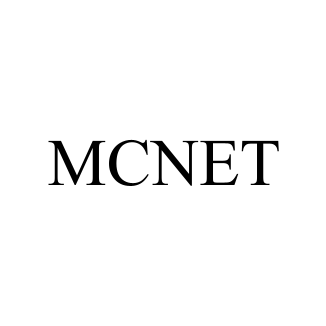}
	\end{subfigure}
	
	&
	
	\begin{subfigure}[t]{0.1\textwidth}
	\hspace{-.3mm}
    \centering
    \includegraphics[width=1\linewidth]{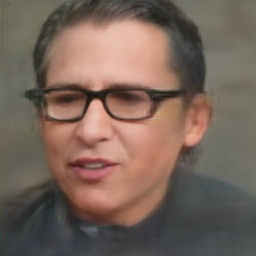}
	\end{subfigure}
	
	\begin{subfigure}[t]{0.1\textwidth}
	\hspace{-.3mm}
    \centering
    \includegraphics[width=1\linewidth]{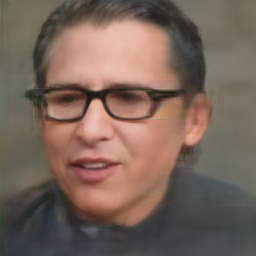}
	\end{subfigure}
	
	\begin{subfigure}[t]{0.1\textwidth}
	\hspace{-.3mm}
    \centering
    \includegraphics[width=1\linewidth]{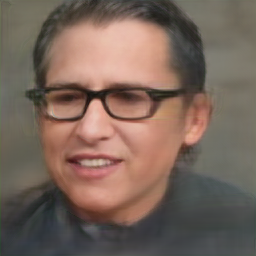}
	\end{subfigure}
	
	\begin{subfigure}[t]{0.1\textwidth}
	\hspace{-.3mm}
    \centering
    \includegraphics[width=1\linewidth]{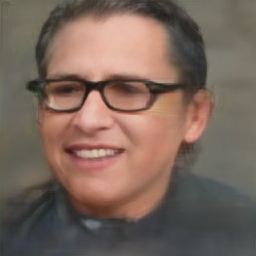}
	\end{subfigure}
	
	\begin{subfigure}[t]{0.1\textwidth}
	\hspace{-.3mm}
    \centering
    \includegraphics[width=1\linewidth]{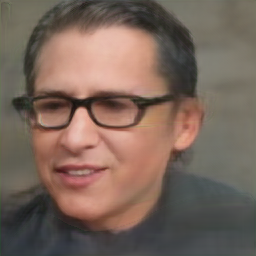}
	\end{subfigure}
	
	\begin{subfigure}[t]{0.1\textwidth}
	\hspace{-.3mm}
    \centering
    \includegraphics[width=1\linewidth]{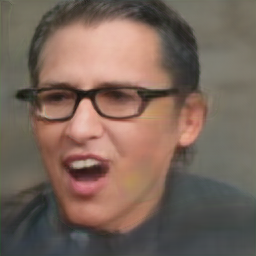}
	\end{subfigure}
	
	\begin{subfigure}[t]{0.1\textwidth}
	\hspace{-.3mm}
    \centering
    \includegraphics[width=1\linewidth]{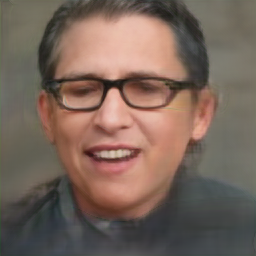}
	\end{subfigure}
	
	\begin{subfigure}[t]{0.1\textwidth}
	\hspace{-.3mm}
    \centering
    \includegraphics[width=1\linewidth]{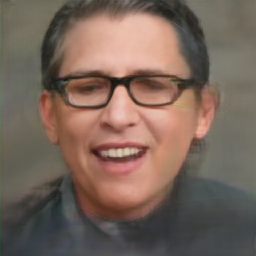}
	\end{subfigure}

	\\
	
	\begin{subfigure}[t]{0.1\textwidth}
    \hspace{-.3mm}
	\centering
    \includegraphics[width=1\linewidth]{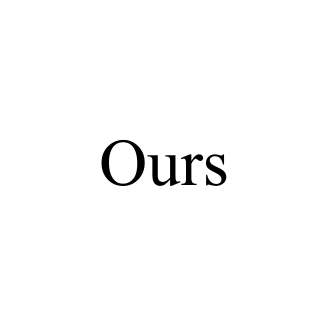}
	\end{subfigure}
	
	&
	
	\begin{subfigure}[t]{0.1\textwidth}
	\hspace{-.3mm}
    \centering
    \includegraphics[width=1\linewidth]{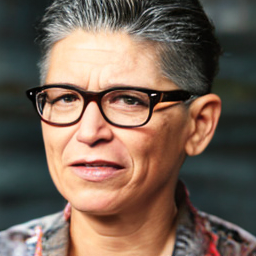}
	\end{subfigure}
	
	\begin{subfigure}[t]{0.1\textwidth}
	\hspace{-.3mm}
    \centering
    \includegraphics[width=1\linewidth]{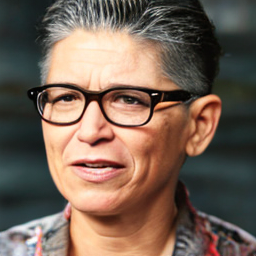}
	\end{subfigure}
	
	\begin{subfigure}[t]{0.1\textwidth}
	\hspace{-.3mm}
    \centering
    \includegraphics[width=1\linewidth]{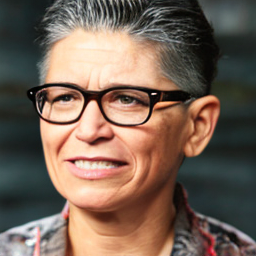}
	\end{subfigure}
	
	\begin{subfigure}[t]{0.1\textwidth}
	\hspace{-.3mm}
    \centering
    \includegraphics[width=1\linewidth]{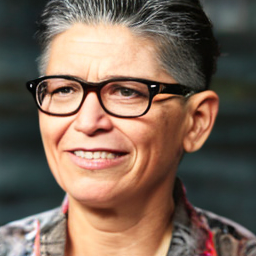}
	\end{subfigure}
	
	\begin{subfigure}[t]{0.1\textwidth}
	\hspace{-.3mm}
    \centering
    \includegraphics[width=1\linewidth]{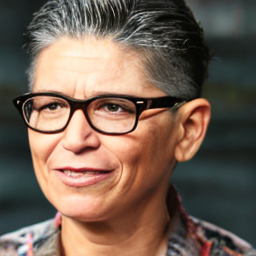}
	\end{subfigure}
	
	\begin{subfigure}[t]{0.1\textwidth}
	\hspace{-.3mm}
    \centering
    \includegraphics[width=1\linewidth]{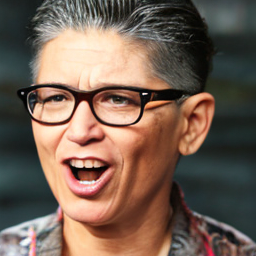}
	\end{subfigure}
	
	\begin{subfigure}[t]{0.1\textwidth}
	\hspace{-.3mm}
    \centering
    \includegraphics[width=1\linewidth]{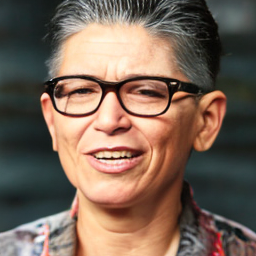}
	\end{subfigure}
	
	\begin{subfigure}[t]{0.1\textwidth}
	\hspace{-.3mm}
    \centering
    \includegraphics[width=1\linewidth]{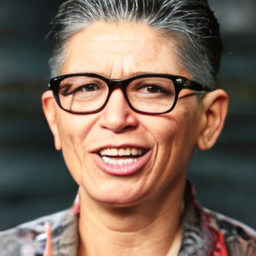}
	\end{subfigure}
	
	\\

\end{tabular}
	
\caption{\textbf{Qualitative Results.} Comparisons with the competing methods~\cite{siarohin2019first, hong2022depth, hong2023implicit} for cross-identity reenactment, showing our approach achieves superior image quality and motion consistency. Additional visual results can be found in the supplementary material.}
\label{fig:reenactment}
\end{figure*}

To demonstrate effectiveness in real-world scenarios with diverse identities, we perform cross-identity reenactment, pairing $5$K driving videos from our test set with $5$K random FFHQ images. Despite the increased challenge of significant source-target variations, our method maintains superior performance in preserving fine motion details and overall scene consistency, while also achieving high-quality video generation, as shown in Table~\ref{tab:quantitative_combined}. Qualitative comparisons (Fig.~\ref{fig:reenactment}) further highlight our ability to produce artifact-free frames that faithfully preserve source identity and transfer target facial expressions and poses.

\subsection{Text-to-Video and Semantic Editing}
While our primary use of CLIP is to capture scene structure and identity, its versatility unlocks additional capabilities beyond standard reenactment.
Notably, using text as our source enables text-to-video generation, where scenes created from textual descriptions mirror the motion of driving videos. Additionally, we can combine CLIP embeddings of a source image with those of text to perform facial video editing, encompassing subtle to major transformations in appearance, narrative, or content. This contrasts with face reenactment, which focuses on generating new videos from minimal inputs. We exemplify these extended capabilities in Fig.~\ref{fig:figure1} and the supplementary material.

\subsection{Ablation Studies}
In this section, we conduct an ablation study focusing on key guidance mechanisms within our diffusion process.

\subsubsection{Scene Recognition}
First, we conduct an experiment to determine the most suitable encoder for capturing global video characteristics.
While ArcFace~\cite{deng2019arcface} is commonly used for identity encoding, we aim for an encoder capable of distinguishing between scenes, even when the same person appears but with different backgrounds or outfits. CLIP~\cite{radford2021learning} is a natural choice for this task due to its ability to encode both semantic information and identity.
We further experiment with augmenting the CLIP embeddings by training a small MLP with $5$ hidden layers on top, using different discriminative losses: center loss \cite{wen2016discriminative} (MLP-Centers), ArcFace loss~\cite{deng2019arcface} (MLP-ArcFace), and Focal loss \cite{lin2017focal} (MLP-Focal). Table \ref{tab:video_separability} reports the Davies–Bouldin~\cite{davies1979cluster} and Calinski–Harabasz~\cite{calinski1974dendrite} indices for clustering embeddings from $32768$ frames across $512$ videos. CLIP demonstrates superior clustering, with data points more spread out between clusters than within them. The supplementary material includes t-SNE projections further showing CLIP's ability to distinguish between scenes.

\begin{table}[ht]
\small
\centering
\begin{tabular}{l c c}
Method                  & Davies-Bouldin $\downarrow$ & Calinski-Harabasz $\uparrow$ \\ \hline
ArcFace                 & 1.516          & 113 \\
CLIP                    & \textbf{0.611} & \textbf{641} \\
MLP-Centers & 0.819          & 411 \\
MLP-ArcFace & 0.810          & 413  \\
MLP-Focal  & 0.673          & 493  \\    
\end{tabular}
\caption{\textbf{Scene Recognition.} The Davies–Bouldin and Calinski–Harabasz indices for clustering $32768$ frames from $512$ videos. CLIP embeddings demonstrate superior clustering performance compared to all other representations.}
\label{tab:video_separability}
\vspace{-15pt}
\end{table}

\subsubsection{Impact of Guidance Signals}
We investigate the imact of each guidance element on the diffusion performance, focusing on image denoising of varying noise values according to the diffusion steps. We trained four DiT-B/4 models on FFHQ~\cite{karras2019style} for face denosing with different guidance: none (baseline), MediaPipe landmarks~\cite{lugaresi2019mediapipe} (Landmark), CLIP embeddings~\cite{radford2021learning} (CLIP), and both combined (Landmark + CLIP).

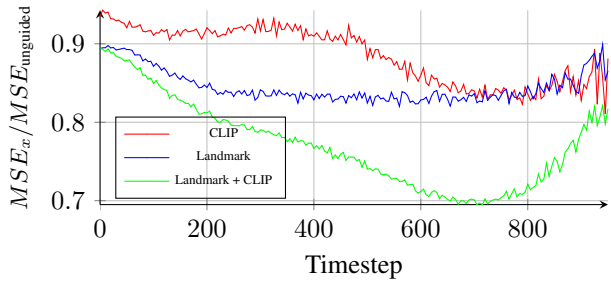
\begin{figure}
\centering
\begin{tikzpicture}
\begin{axis}[
            xlabel={Timestep},
            ylabel={\small $MSE_{x} / MSE_{\text{unguided}}$},
            grid=both,
            grid style={line width=.1pt, draw=gray!10},
            major grid style={line width=.2pt,draw=gray!50},
            width=1.0\columnwidth,
            height=0.5\linewidth,
            axis lines=left,
            ylabel near ticks,
            xlabel near ticks,
            scaled ticks=true,
            clip=false,
            scaled ticks=true,
            legend pos=south west,
            legend style={font=\tiny, fill=none},
        ]
        
        \addplot[
            color=red,
            mark=*,
            mark size=0.01pt,
        ] 
        table[col sep=comma, x=timesteps, y=mclip] {figures/mse-reconstruction/noise_exploration_v2.csv};
        
        \addplot[
            color=blue,
            mark=*,
            mark size=0.01pt,
        ]
        table[col sep=comma, x=timesteps, y=mlandmark] {figures/mse-reconstruction/noise_exploration_v2.csv};
        
        \addplot[
            color=green,
            mark=*,
            mark size=0.01pt,
        ] 
        table[col sep=comma, x=timesteps, y=mlandmark_clip] {figures/mse-reconstruction/noise_exploration_v2.csv};
        
        \legend{CLIP, Landmark, Landmark + CLIP};
        \end{axis}
    \end{tikzpicture}

\caption{\textbf{Impact of Guidance.}
MSE improvement of guided diffusion models over the unguided baseline across timesteps, demonstrating the impact of different guidance signals on face denoising performance.
Landmark guidance is most effective in early stages, while the combination of landmark and CLIP guidance yields the best overall reconstruction.
}
\label{fig:mse}
\end{figure}

Fig.~\ref{fig:mse} shows the recovery improvement of each guided model over the baseline across timesteps. Interestingly, landmark guidance is most beneficial in early, high-uncertainty stages, suggesting positional information about the face is is more important than  semantic properties. As expected, the landmark-CLIP combination provides the best overall performance. Moreover, all models exhibit the most significant improvements in middle stages, likely due to the noise level being most amenable to guidance in this range.

\section{Limitations and Broader Impact}
While our method shows strong potential for generating videos, its demonstration is limited to face reenactment using facial landmarks as the primary control signal. Future work could explore several promising directions. First, we can expand the types of guidance beyond facial landmarks, incorporating segmentation maps, optical flow, depth maps, and other modalities. Second, we can apply our approach to diverse domains like image-to-video, inverse problems, cinemagraphs, and special effects. Finally, our method is limited to a single scene or shot, but we envision using multiple or moving anchors to enable multi-scene video generation.

Recognizing the potential for misuse of our work, we advocate for responsible use and the development of detection mechanisms to identify manipulated or misleading content. To mitigate potential harm, we will implement strict access control measures, limiting access to our models and datasets exclusively to authorized research purposes.

\section{Conclusion}
\label{sec:Conclusion}
We introduced Anchored Diffusion for generating long, coherent videos. We presented sDiT, a direct extension of DiTs to video generation, incorporating temporal information through guidance and trained using a novel strategy based on random non-uniform video sequences. 
Leveraging this training strategy and the unique structure of Transformers, we developed an inference mechanism generating multiple aligned video sequences of the same scene, ensuring consistency and smooth motion.
We demonstrated state-of-the-art results in face reenactment, aided by a newly curated, large-scale facial video dataset.
Our approach offers improved video fidelity, temporal consistency, and editing capabilities, opening new avenues for video generation.

\appendix
\section*{\Large Supplementary Material}
\section{Artistic Reenactment}
We introduce an additional application: artistic reenactment, which involves transferring facial expressions and movements from a driving video to a target artistic portrait. To address the domain gap between our curated training data clips and the desired artistic domain, we incorporate 25,000 artistic images from the Artstation-Artistic-face-HQ (AAHQ)~\cite{liu2021blendgan} dataset into our training scheme. Geometric transformations are applied to each sample to conceptualize a series of video clips.

\begin{figure*}
\centering
\captionsetup[subfigure]{labelformat=empty,justification=centering,aboveskip=1pt,belowskip=1pt}
    
    \begin{tabular}[c]{c c}
    
    \vspace{-1.mm}
    
    \begin{subfigure}[t]{0.075\textwidth}
    \centering
    \includegraphics[width=1\linewidth]{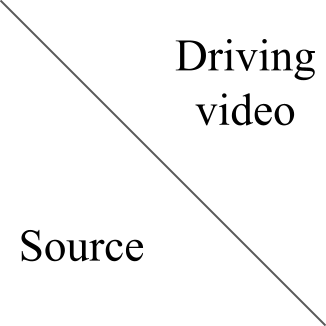}
    \end{subfigure}
	
    &
	
    \begin{subfigure}[t]{0.9\textwidth}
    \centering
    \includegraphics[width=1\linewidth]{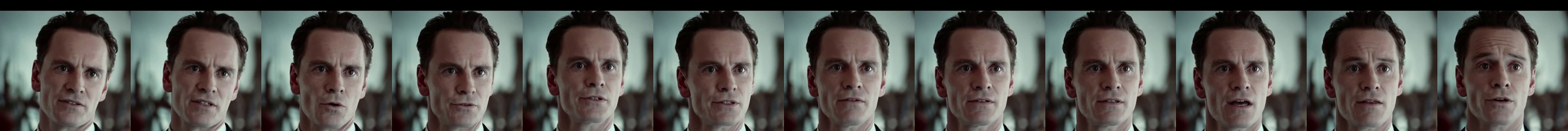}
    \end{subfigure}
	
    \\
    
    \vspace{-1.mm}
    
    \begin{subfigure}[t]{0.075\textwidth}
    \centering
    \includegraphics[width=1\linewidth]{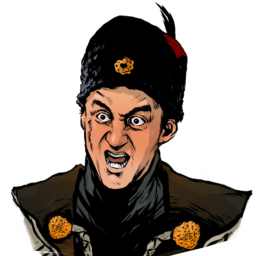}
    \end{subfigure}
	
    &
	
    \begin{subfigure}[t]{0.9\textwidth}
    \centering
    \includegraphics[width=1\linewidth]{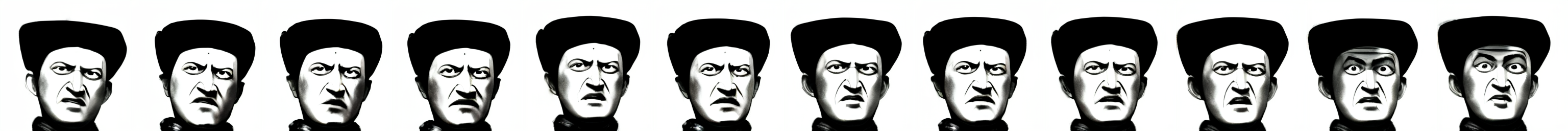}
    \end{subfigure}

    \\
    
    \vspace{-1.mm}
    
    \begin{subfigure}[t]{0.075\textwidth}
    \centering
    \includegraphics[width=1\linewidth]{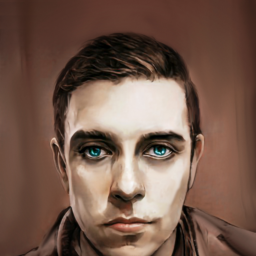}
    \end{subfigure}
	
    &
	
    \begin{subfigure}[t]{0.9\textwidth}
    \centering
    \includegraphics[width=1\linewidth]{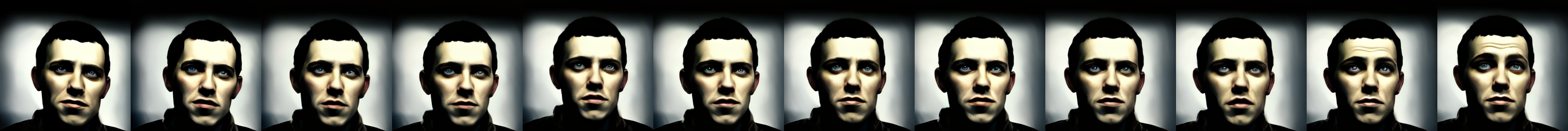}
    \end{subfigure}

    \\
    
    \vspace{-1.mm}
    
    \begin{subfigure}[t]{0.075\textwidth}
    \centering
    \includegraphics[width=1\linewidth]{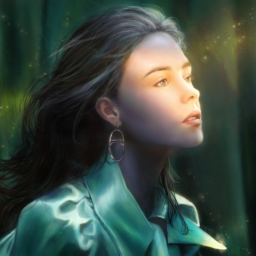}
    \end{subfigure}
	
    &
	
    \begin{subfigure}[t]{0.9\textwidth}
    \centering
    \includegraphics[width=1\linewidth]{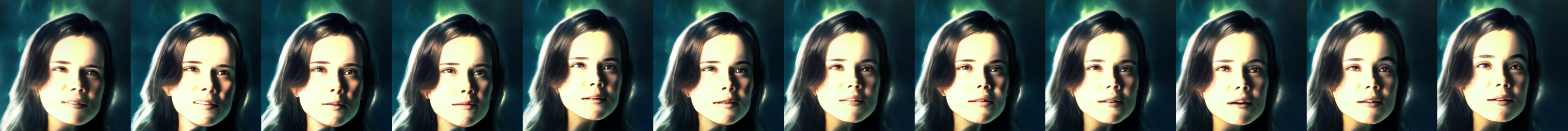}
    \end{subfigure}

    \\
    
    \vspace{-1.mm}
    
    \begin{subfigure}[t]{0.075\textwidth}
    \centering
    \includegraphics[width=1\linewidth]{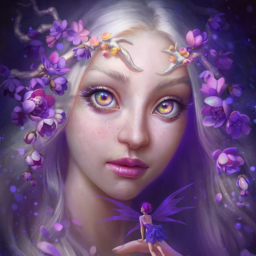}
    \end{subfigure}
	
    &
	
    \begin{subfigure}[t]{0.9\textwidth}
    \centering
    \includegraphics[width=1\linewidth]{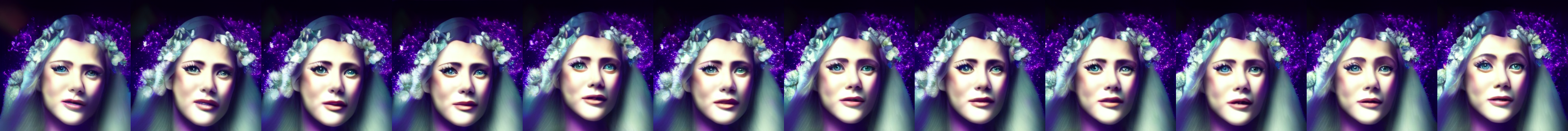}
    \end{subfigure}

    \end{tabular}
	
\caption{\textbf{Artistic Reenactment.} Results of the artistic reenactment process.}
\label{fig:art_sm1}
\end{figure*}

\begin{figure*}
    \centering
    \captionsetup[subfigure]{labelformat=empty,justification=centering,aboveskip=1pt,belowskip=1pt}
        
        \begin{tabular}[c]{c c}
        
        \vspace{-1.mm}
        
        \begin{subfigure}[t]{0.075\textwidth}
        \centering
        \includegraphics[width=1\linewidth]{figures/artistic/corner.png}
        \end{subfigure}
        
        &
        
        \begin{subfigure}[t]{0.9\textwidth}
        \centering
        \includegraphics[width=1\linewidth]{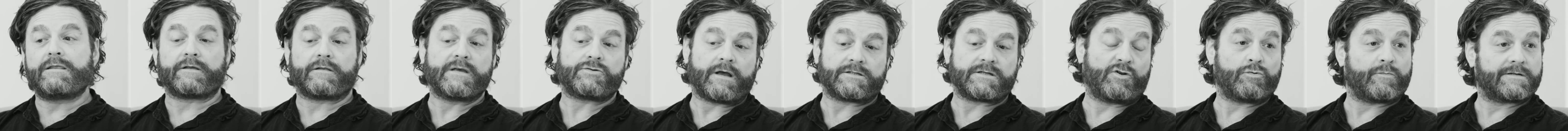}
        \end{subfigure}
        
        \\
        
        \vspace{-1.mm}
        
        \begin{subfigure}[t]{0.075\textwidth}
        \centering
        \includegraphics[width=1\linewidth]{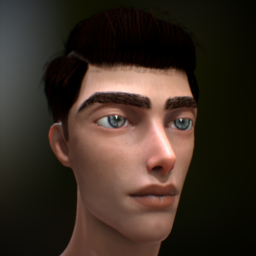}
        \end{subfigure}
        
        &
        
        \begin{subfigure}[t]{0.9\textwidth}
        \centering
        \includegraphics[width=1\linewidth]{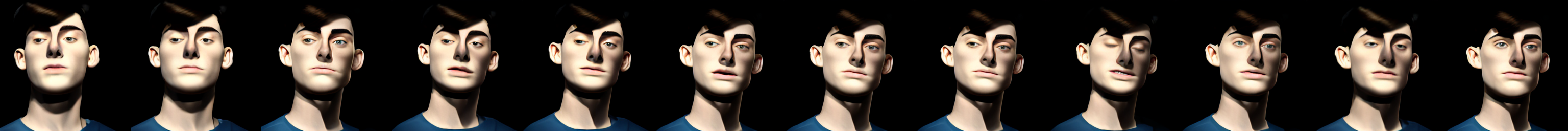}
        \end{subfigure}

        \\
        
        \vspace{-1.mm}
        
        \begin{subfigure}[t]{0.075\textwidth}
        \centering
        \includegraphics[width=1\linewidth]{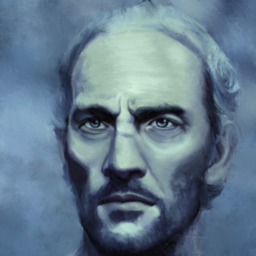}
        \end{subfigure}
        
        &
        
        \begin{subfigure}[t]{0.9\textwidth}
        \centering
        \includegraphics[width=1\linewidth]{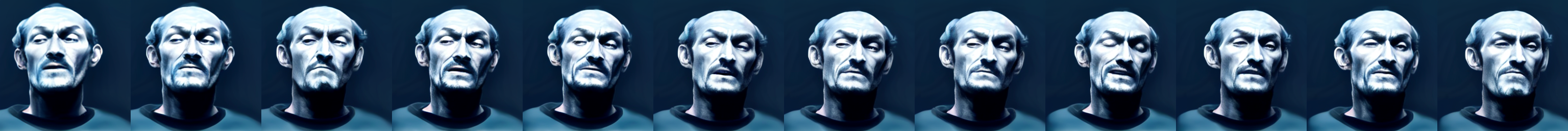}
        \end{subfigure}

        \\
        
        \vspace{-1.mm}
        
        \begin{subfigure}[t]{0.075\textwidth}
        \centering
        \includegraphics[width=1\linewidth]{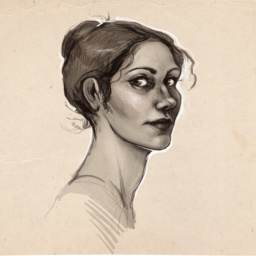}
        \end{subfigure}
        
        &
        
        \begin{subfigure}[t]{0.9\textwidth}
        \centering
        \includegraphics[width=1\linewidth]{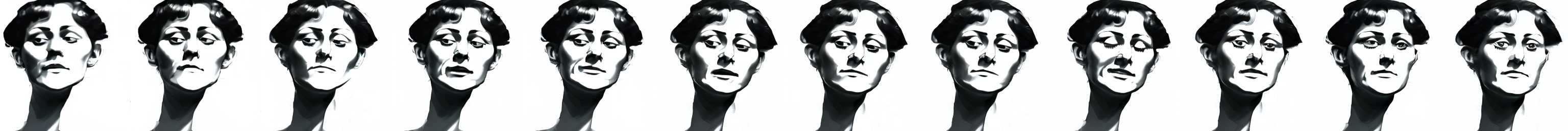}
        \end{subfigure}

        \\
        
        \vspace{-1.mm}
        
        \begin{subfigure}[t]{0.075\textwidth}
        \centering
        \includegraphics[width=1\linewidth]{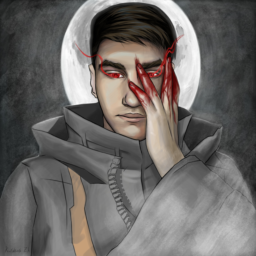}
        \end{subfigure}
        
        &
        
        \begin{subfigure}[t]{0.9\textwidth}
        \centering
        \includegraphics[width=1\linewidth]{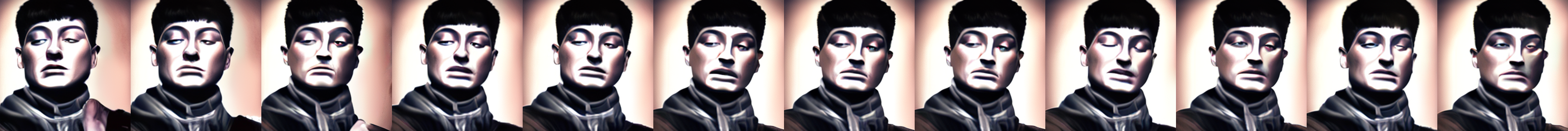}
        \end{subfigure}

        \end{tabular}
        
    \caption{\textbf{Artistic Reenactment.} Additional results of the artistic reenactment process.}
    \label{fig:art_sm2}
    \end{figure*}

Interestingly, as demonstrated in Fig.~\ref{fig:art_sm1} and Fig.~\ref{fig:art_sm2}, although the geometric transformations do not include detailed movement information, such as different poses or eye closure, the model successfully learns to reenact artistic video clips. This success is attributed to the integration with real-world clips, which enables the model to effectively bridge the gap between artistic and realistic domains.

\section{Dataset Statistics}
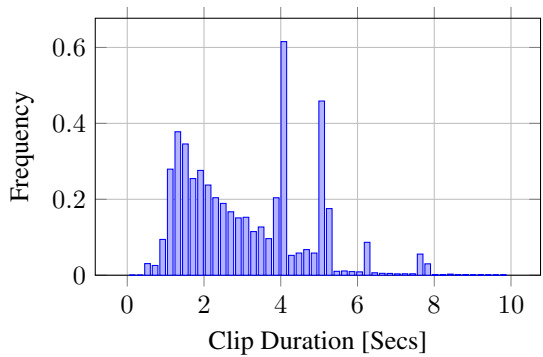
\begin{figure}
    \centering
    \begin{tikzpicture}
        \begin{axis}[
            ybar,
            width=0.9\columnwidth,
            height=0.6\linewidth,
            xlabel={Clip Duration [Secs]},
            ylabel={Frequency},
            ymin=0,
            grid=major,
            bar width=0.075cm,
            ]
            \pgfplotstableread[col sep=comma]{figures/data-statics/csv/hist_duration.csv}\datatable
            \addplot table[x=bin_edge, y=frequencies] {\datatable};
        \end{axis}
    \end{tikzpicture}
    \caption{\textbf{Clip Duration.} This histogram shows the distribution of clip durations in our dataset, with an average duration of $3.29$ seconds and a standard deviation of $2.07$ seconds.}
    \label{fig:duration_hist}
\end{figure}

In this work, we introduce \textit{ReenactFaces-1M}, a large-scale, high-quality, and diverse video dataset. \textit{ReenactFaces-1M} comprises $1,006,257$ video segments, each with an average length of $3.29$ seconds, totaling over $920$ hours of footage. The dataset exhibits an average resolution of $745$ pixels, making it a valuable resource for various applications in video analysis and facial recognition research. To further understand the characteristics of our dataset, we provide an analysis of important data statistics:

\begin{itemize}[leftmargin=*]
  \item Figure \ref{fig:duration_hist} shows the distribution of clip durations in our dataset, with an average duration of $3.29$ seconds and a standard deviation of $2.07$ seconds.
  \item Figure \ref{fig:hyper_hist} shows the distribution of clip HyperIQA \cite{su2020blindly} scores in our dataset, with an average duration of $51.5$ and a standard deviation of $10.72$.
  \item Figure \ref{fig:resolution_hist} shows the distribution of clip resolution in our dataset, with an average duration of $745.1$ and a standard deviation of $247.8$.
  \item Figure \ref{fig:facial_size_hist} depicts the distribution of the face height ratio relative to the total clip height and the face width ratio relative to the total clip width. The face width ratio has a mean of $0.45$ and a standard deviation of $0.05$, while the face height ratio has a mean of $0.53$ and a standard deviation of $0.07$.
  
\end{itemize}

\begin{figure}
    \centering
    \begin{tikzpicture}
        \begin{axis}[
            ybar,
            width=0.9\columnwidth,
            height=0.6\linewidth,
            xlabel={HyperIQA},
            ylabel={Frequency},
            ymin=0,
            grid=major,
            bar width=0.075cm,
            ]
            \pgfplotstableread[col sep=comma]{figures/data-statics/csv/hist_hyper.csv}\datatable
            \addplot table[x=bin_edge, y=frequencies] {\datatable};
        \end{axis}
    \end{tikzpicture}
    \caption{\textbf{Clip HyperIQA.} This histogram shows the distribution of clip HyperIQA scores in our dataset, with an average duration of $51.5$ and a standard deviation of $10.72$.}
    \label{fig:hyper_hist}
\end{figure}
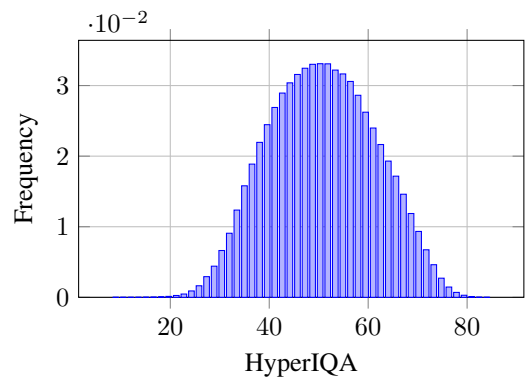
\begin{figure}
    \centering
    \begin{tikzpicture}
        \begin{axis}[
            ybar,
            width=0.9\columnwidth,
            height=0.6\linewidth,
            xlabel={Resolution [Pixels]},
            ylabel={Frequency},
            ymin=0,
            grid=major,
            bar width=0.075cm,
            ]
            \pgfplotstableread[col sep=comma]{figures/data-statics/csv/hist_resolution.csv}\datatable
            \addplot table[x=bin_edge, y=frequencies] {\datatable};
        \end{axis}
    \end{tikzpicture}
    \caption{\textbf{Clip Resolution.} This histogram shows the distribution of clip resolution in our dataset, with an average duration of $745.1$ and a standard deviation of $247.8$.}
    \label{fig:resolution_hist}
\end{figure}
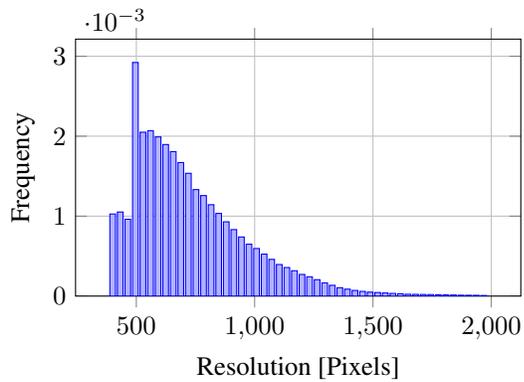
\begin{figure}
    \centering
    \begin{tikzpicture}
        \begin{axis}[
            ybar,
            width=0.9\columnwidth,
            height=0.6\linewidth,
            xlabel={Ratio},
            ylabel={Frequency},
            ymin=0,
            grid=major,
            bar width=0.075cm,
            legend style={font=\small},
            ]
            \pgfplotstableread[col sep=comma]{figures/data-statics/csv/hist_face_height.csv}\datatable
            \addplot table[x=bin_edge, y=frequencies] {\datatable};
            \pgfplotstableread[col sep=comma]{figures/data-statics/csv/hist_face_width.csv}\datatable
            \addplot table[x=bin_edge, y=frequencies] {\datatable};
            \legend{Height,Width}
        \end{axis}
    \end{tikzpicture}
    \caption{\textbf{Facial Ratio.} The histogram depicts the distribution of the face height ratio relative to the total clip height and the face width ratio relative to the total clip width. 
    The face width ratio has a mean of $0.45$ and a standard deviation of $0.05$, while the face height ratio has a mean of $0.53$ and a standard deviation of $0.07$.
    }
    \label{fig:facial_size_hist}
\end{figure}
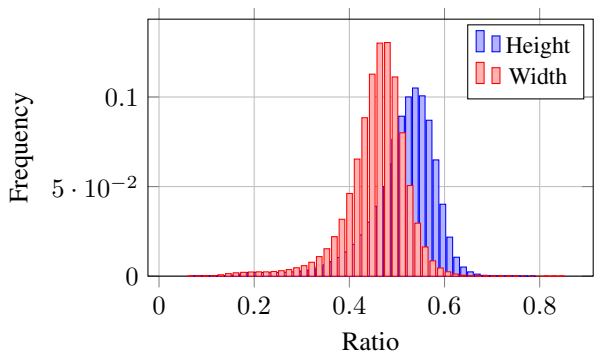

\section{Face Reenactment - Extended Results}

\subsection{Analysis of Scene Recognition}
\begin{figure*}
	\centering
	\captionsetup[subfigure]{labelformat=empty,justification=centering,aboveskip=1pt,belowskip=1pt}
    
    \begin{tabular}[c]{c c}
    
    
        \begin{subfigure}[t]{0.4\textwidth}
            \centering
            \includegraphics[width=1\linewidth]{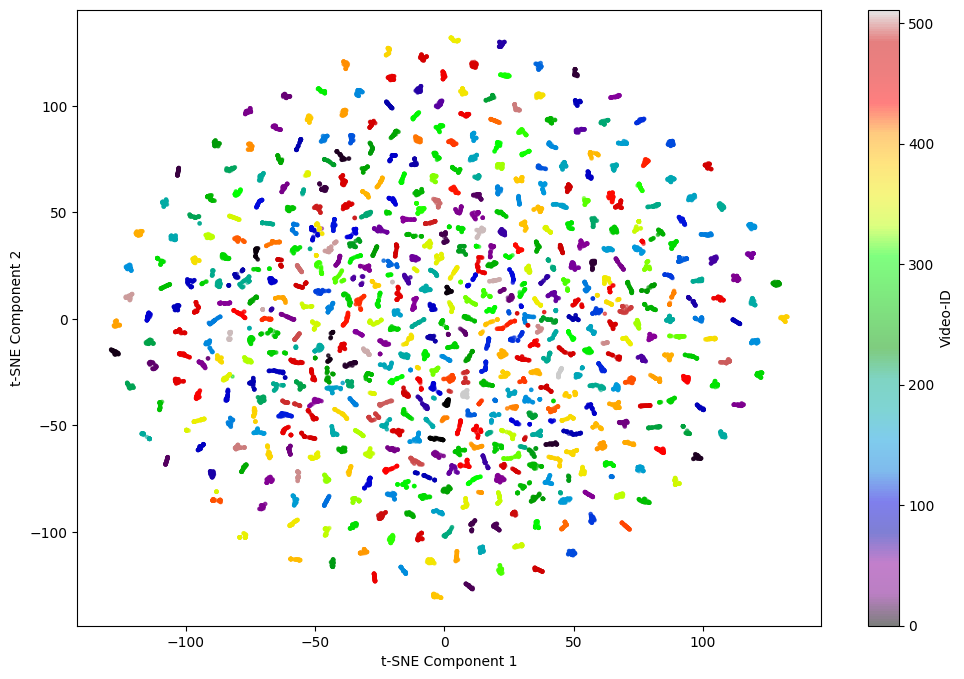}
            ArcFace
        \end{subfigure}

	&
	
	\begin{subfigure}[t]{0.4\textwidth}
            \centering
             \includegraphics[width=1\linewidth]{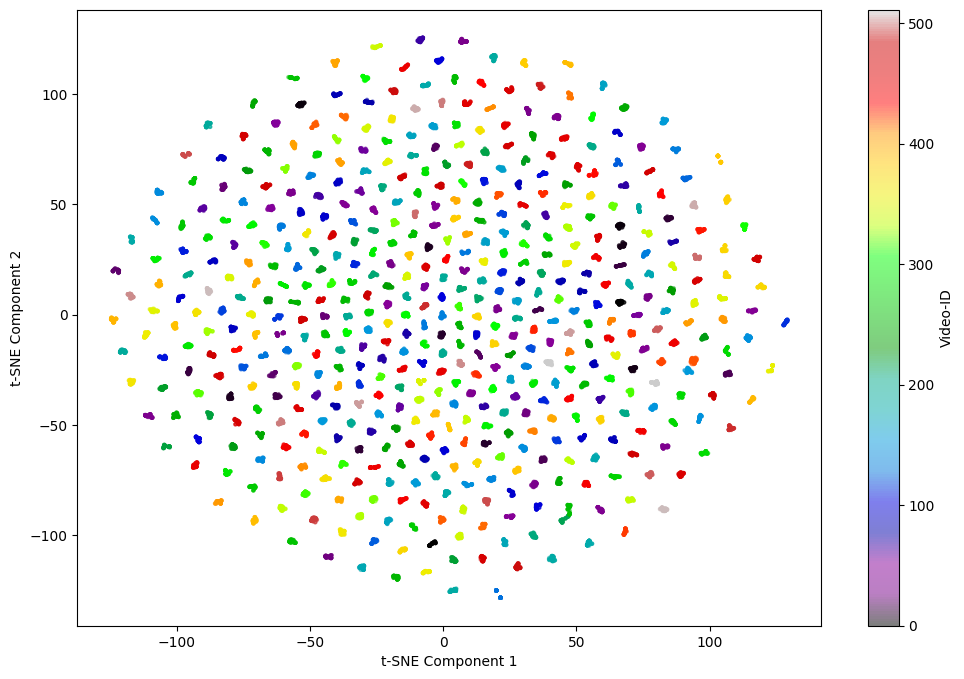}
             CLIP
	\end{subfigure}

	\end{tabular}
	
\caption{\textbf{Scene Recognition.} t-SNE 2D projection of ArcFace Embeddings (Left) and CLIP embeddings (Right).}
\label{fig:tsne}
\end{figure*}

To assess the scene recognition capabilities of our approach, we analyzed both ArcFace and CLIP embeddings of video frames. Figure~\ref{fig:tsne} presents t-SNE visualizations of these embeddings, where each point represents a frame and its color corresponds to the video it belongs to. The ArcFace embeddings are not as well-separated, failing to distinguish between certain videos. In contrast, the CLIP embeddings resulted in clear separation of clusters, indicating that they effectively distinguish between different scenes and movies, highlighting their potential for such tasks.

\subsection{Additional Visual Results}
\begin{figure*}[ht!]
	\centering
	\captionsetup[subfigure]{labelformat=empty,justification=centering,aboveskip=1pt,belowskip=1pt}
    
    \begin{tabular}[c]{c c}
    
    \vspace{-2.5mm}
    
    \begin{subfigure}[t]{0.1\textwidth}
    \hspace{-.3mm}
	\centering
    Source
	\end{subfigure}
	
	&
	
	\begin{subfigure}[t]{0.8\textwidth}
    \hspace{-.3mm}
	\centering
    Driving video
	\end{subfigure}
	
	\\
	
    \vspace{-2.5mm}
    
    \begin{subfigure}[t]{0.1\textwidth}
    \hspace{-.3mm}
	\centering
    \includegraphics[width=1\linewidth]{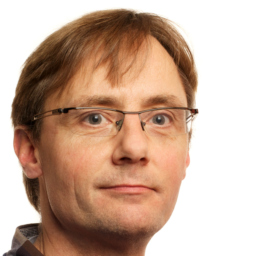}
	\end{subfigure}
	
	&
	
	\begin{subfigure}[t]{0.1\textwidth}
	\hspace{-.3mm}
    \centering
    \includegraphics[width=1\linewidth]{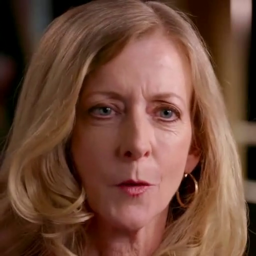}
	\end{subfigure}
	
	\begin{subfigure}[t]{0.1\textwidth}
	\hspace{-.3mm}
    \centering
    \includegraphics[width=1\linewidth]{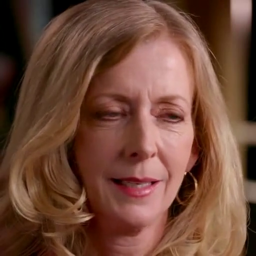}
	\end{subfigure}
	
	\begin{subfigure}[t]{0.1\textwidth}
	\hspace{-.3mm}
    \centering
    \includegraphics[width=1\linewidth]{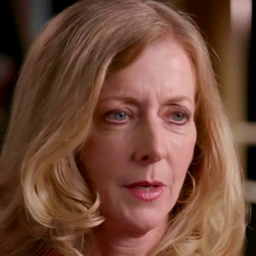}
	\end{subfigure}
	
	\begin{subfigure}[t]{0.1\textwidth}
	\hspace{-.3mm}
    \centering
    \includegraphics[width=1\linewidth]{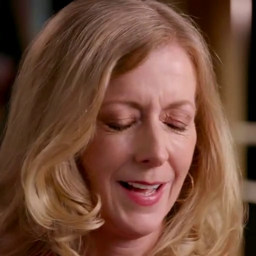}
	\end{subfigure}
	
	\begin{subfigure}[t]{0.1\textwidth}
	\hspace{-.3mm}
    \centering
    \includegraphics[width=1\linewidth]{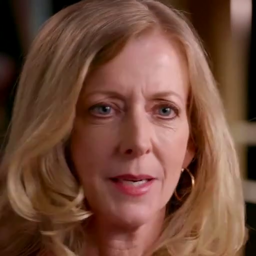}
	\end{subfigure}
	
	\begin{subfigure}[t]{0.1\textwidth}
	\hspace{-.3mm}
    \centering
    \includegraphics[width=1\linewidth]{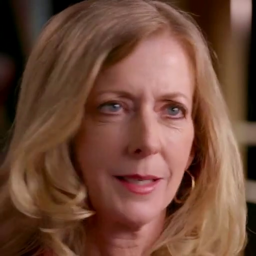}
	\end{subfigure}
	
	\begin{subfigure}[t]{0.1\textwidth}
	\hspace{-.3mm}
    \centering
    \includegraphics[width=1\linewidth]{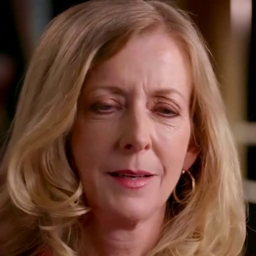}
	\end{subfigure}
	
	\begin{subfigure}[t]{0.1\textwidth}
	\hspace{-.3mm}
    \centering
    \includegraphics[width=1\linewidth]{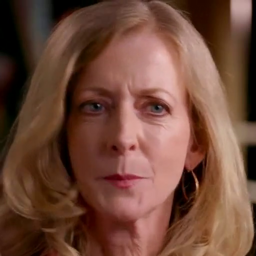}
	\end{subfigure}

	\\
	
	\vspace{-2.5mm}
	
	\begin{subfigure}[t]{0.1\textwidth}
    \hspace{-.3mm}
	\centering
    \includegraphics[width=1\linewidth]{figures/reenactment/samples/FOMM.png}
	\end{subfigure}
	
	&
	
	\begin{subfigure}[t]{0.1\textwidth}
	\hspace{-.3mm}
    \centering
    \includegraphics[width=1\linewidth]{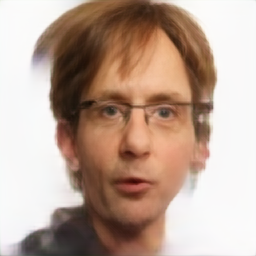}
	\end{subfigure}
	
	\begin{subfigure}[t]{0.1\textwidth}
	\hspace{-.3mm}
    \centering
    \includegraphics[width=1\linewidth]{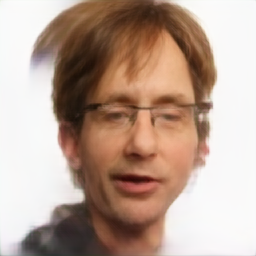}
	\end{subfigure}
	
	\begin{subfigure}[t]{0.1\textwidth}
	\hspace{-.3mm}
    \centering
    \includegraphics[width=1\linewidth]{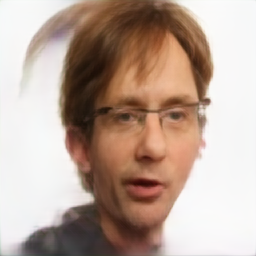}
	\end{subfigure}
	
	\begin{subfigure}[t]{0.1\textwidth}
	\hspace{-.3mm}
    \centering
    \includegraphics[width=1\linewidth]{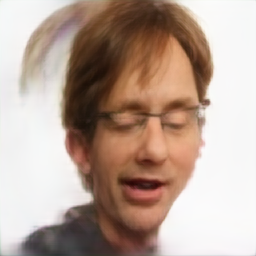}
	\end{subfigure}
	
	\begin{subfigure}[t]{0.1\textwidth}
	\hspace{-.3mm}
    \centering
    \includegraphics[width=1\linewidth]{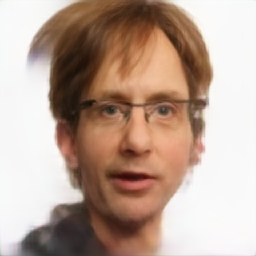}
	\end{subfigure}
	
	\begin{subfigure}[t]{0.1\textwidth}
	\hspace{-.3mm}
    \centering
    \includegraphics[width=1\linewidth]{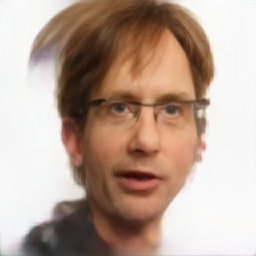}
	\end{subfigure}
	
	\begin{subfigure}[t]{0.1\textwidth}
	\hspace{-.3mm}
    \centering
    \includegraphics[width=1\linewidth]{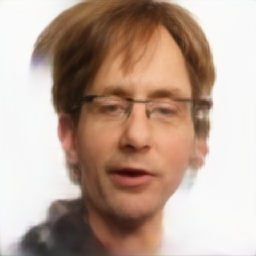}
	\end{subfigure}
	
	\begin{subfigure}[t]{0.1\textwidth}
	\hspace{-.3mm}
    \centering
    \includegraphics[width=1\linewidth]{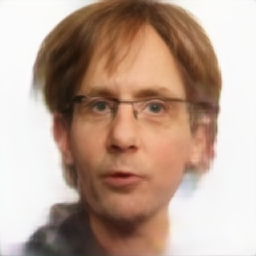}
	\end{subfigure}

	\\
	
	\vspace{-2.5mm}
	
	\begin{subfigure}[t]{0.1\textwidth}
    \hspace{-.3mm}
	\centering
    \includegraphics[width=1\linewidth]{figures/reenactment/samples/DaGAN.png}
	\end{subfigure}
	
	&
	
	\begin{subfigure}[t]{0.1\textwidth}
	\hspace{-.3mm}
    \centering
    \includegraphics[width=1\linewidth]{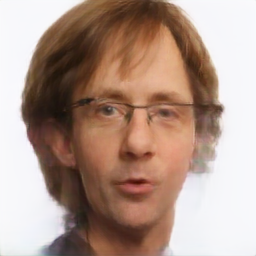}
	\end{subfigure}
	
	\begin{subfigure}[t]{0.1\textwidth}
	\hspace{-.3mm}
    \centering
    \includegraphics[width=1\linewidth]{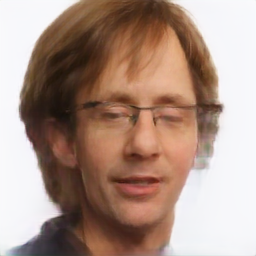}
	\end{subfigure}
	
	\begin{subfigure}[t]{0.1\textwidth}
	\hspace{-.3mm}
    \centering
    \includegraphics[width=1\linewidth]{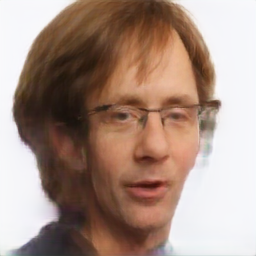}
	\end{subfigure}
	
	\begin{subfigure}[t]{0.1\textwidth}
	\hspace{-.3mm}
    \centering
    \includegraphics[width=1\linewidth]{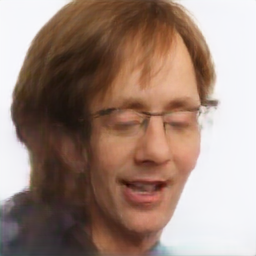}
	\end{subfigure}
	
	\begin{subfigure}[t]{0.1\textwidth}
	\hspace{-.3mm}
    \centering
    \includegraphics[width=1\linewidth]{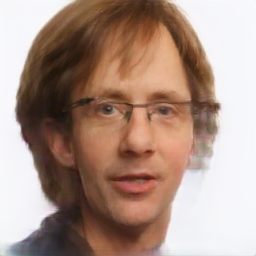}
	\end{subfigure}
	
	\begin{subfigure}[t]{0.1\textwidth}
	\hspace{-.3mm}
    \centering
    \includegraphics[width=1\linewidth]{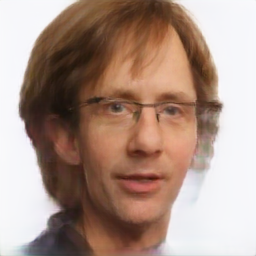}
	\end{subfigure}
	
	\begin{subfigure}[t]{0.1\textwidth}
	\hspace{-.3mm}
    \centering
    \includegraphics[width=1\linewidth]{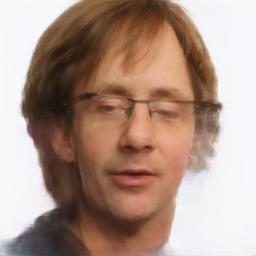}
	\end{subfigure}
	
	\begin{subfigure}[t]{0.1\textwidth}
	\hspace{-.3mm}
    \centering
    \includegraphics[width=1\linewidth]{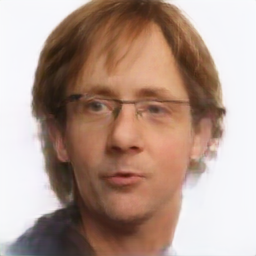}
	\end{subfigure}

	\\
	
	\vspace{-2.5mm}
	
	\begin{subfigure}[t]{0.1\textwidth}
    \hspace{-.3mm}
	\centering
    \includegraphics[width=1\linewidth]{figures/reenactment/samples/MCNET.png}
	\end{subfigure}
	
	&
	
	\begin{subfigure}[t]{0.1\textwidth}
	\hspace{-.3mm}
    \centering
    \includegraphics[width=1\linewidth]{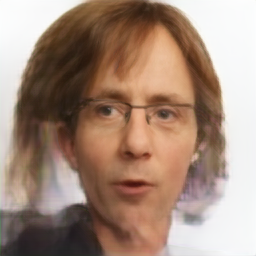}
	\end{subfigure}
	
	\begin{subfigure}[t]{0.1\textwidth}
	\hspace{-.3mm}
    \centering
    \includegraphics[width=1\linewidth]{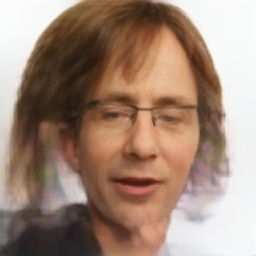}
	\end{subfigure}
	
	\begin{subfigure}[t]{0.1\textwidth}
	\hspace{-.3mm}
    \centering
    \includegraphics[width=1\linewidth]{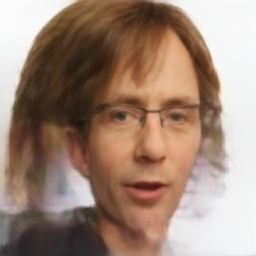}
	\end{subfigure}
	
	\begin{subfigure}[t]{0.1\textwidth}
	\hspace{-.3mm}
    \centering
    \includegraphics[width=1\linewidth]{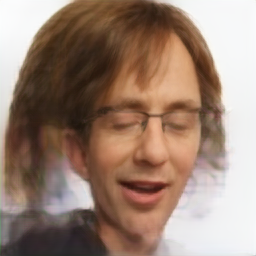}
	\end{subfigure}
	
	\begin{subfigure}[t]{0.1\textwidth}
	\hspace{-.3mm}
    \centering
    \includegraphics[width=1\linewidth]{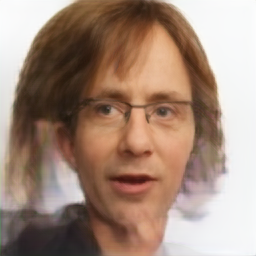}
	\end{subfigure}
	
	\begin{subfigure}[t]{0.1\textwidth}
	\hspace{-.3mm}
    \centering
    \includegraphics[width=1\linewidth]{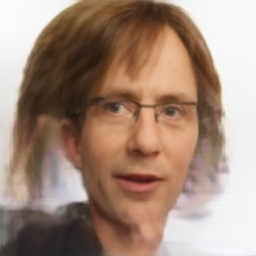}
	\end{subfigure}
	
	\begin{subfigure}[t]{0.1\textwidth}
	\hspace{-.3mm}
    \centering
    \includegraphics[width=1\linewidth]{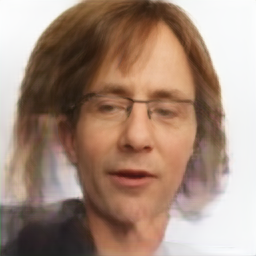}
	\end{subfigure}
	
	\begin{subfigure}[t]{0.1\textwidth}
	\hspace{-.3mm}
    \centering
    \includegraphics[width=1\linewidth]{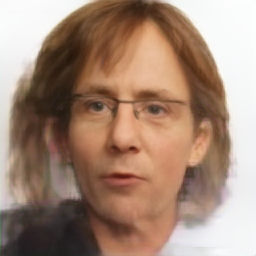}
	\end{subfigure}

	\\
	
	\begin{subfigure}[t]{0.1\textwidth}
    \hspace{-.3mm}
	\centering
    \includegraphics[width=1\linewidth]{figures/reenactment/samples/Ours.png}
	\end{subfigure}
	
	&
	
	\begin{subfigure}[t]{0.1\textwidth}
	\hspace{-.3mm}
    \centering
    \includegraphics[width=1\linewidth]{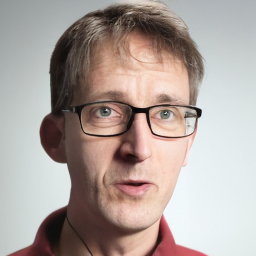}
	\end{subfigure}
	
	\begin{subfigure}[t]{0.1\textwidth}
	\hspace{-.3mm}
    \centering
    \includegraphics[width=1\linewidth]{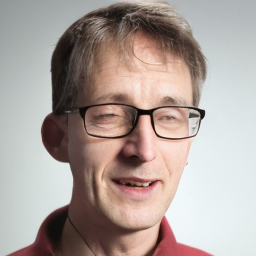}
	\end{subfigure}
	
	\begin{subfigure}[t]{0.1\textwidth}
	\hspace{-.3mm}
    \centering
    \includegraphics[width=1\linewidth]{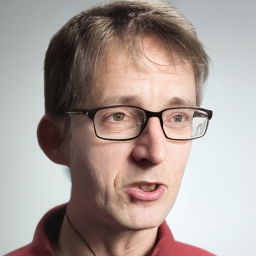}
	\end{subfigure}
	
	\begin{subfigure}[t]{0.1\textwidth}
	\hspace{-.3mm}
    \centering
    \includegraphics[width=1\linewidth]{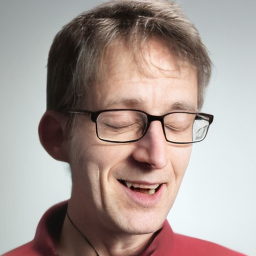}
	\end{subfigure}
	
	\begin{subfigure}[t]{0.1\textwidth}
	\hspace{-.3mm}
    \centering
    \includegraphics[width=1\linewidth]{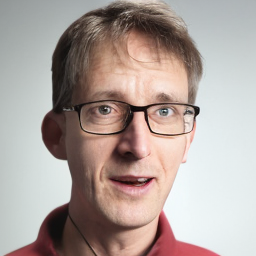}
	\end{subfigure}
	
	\begin{subfigure}[t]{0.1\textwidth}
	\hspace{-.3mm}
    \centering
    \includegraphics[width=1\linewidth]{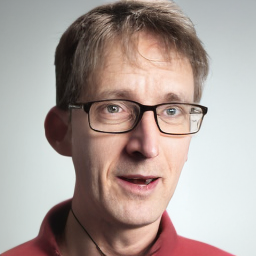}
	\end{subfigure}
	
	\begin{subfigure}[t]{0.1\textwidth}
	\hspace{-.3mm}
    \centering
    \includegraphics[width=1\linewidth]{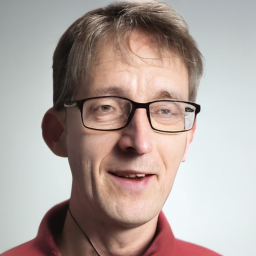}
	\end{subfigure}
	
	\begin{subfigure}[t]{0.1\textwidth}
	\hspace{-.3mm}
    \centering
    \includegraphics[width=1\linewidth]{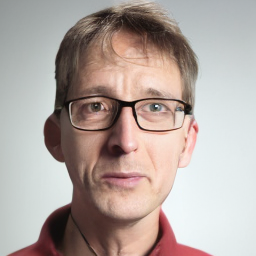}
	\end{subfigure}
	
	\\

\end{tabular}
	
\caption{\textbf{Cross-identity Reenactment.} Comparisons with the competing methods~\cite{siarohin2019first, hong2022depth, hong2023implicit}.}
\label{fig:reenactment_sm6}
\end{figure*}

\begin{figure*}[ht!]
	\centering
	\captionsetup[subfigure]{labelformat=empty,justification=centering,aboveskip=1pt,belowskip=1pt}
    
    \begin{tabular}[c]{c c}
    
    \vspace{-2.5mm}
    
    \begin{subfigure}[t]{0.1\textwidth}
    \hspace{-.3mm}
	\centering
    Source
	\end{subfigure}
	
	&
	
	\begin{subfigure}[t]{0.8\textwidth}
    \hspace{-.3mm}
	\centering
    Driving video
	\end{subfigure}
	
	\\
	
    \vspace{-2.5mm}
    
    \begin{subfigure}[t]{0.1\textwidth}
    \hspace{-.3mm}
	\centering
    \includegraphics[width=1\linewidth]{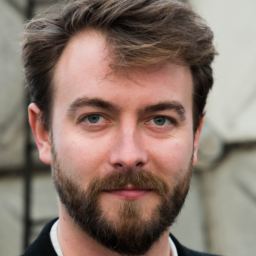}
	\end{subfigure}
	
	&
	
	\begin{subfigure}[t]{0.1\textwidth}
	\hspace{-.3mm}
    \centering
    \includegraphics[width=1\linewidth]{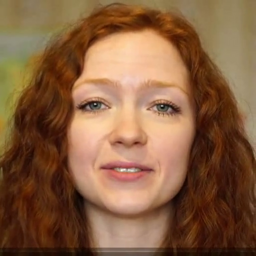}
	\end{subfigure}
	
	\begin{subfigure}[t]{0.1\textwidth}
	\hspace{-.3mm}
    \centering
    \includegraphics[width=1\linewidth]{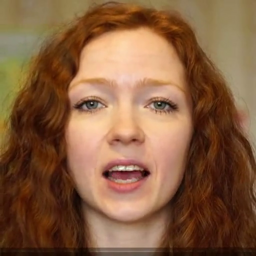}
	\end{subfigure}
	
	\begin{subfigure}[t]{0.1\textwidth}
	\hspace{-.3mm}
    \centering
    \includegraphics[width=1\linewidth]{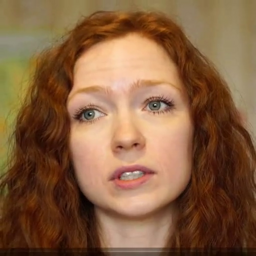}
	\end{subfigure}
	
	\begin{subfigure}[t]{0.1\textwidth}
	\hspace{-.3mm}
    \centering
    \includegraphics[width=1\linewidth]{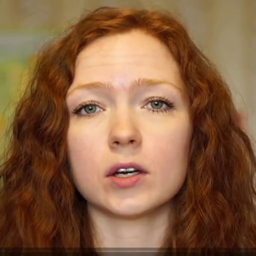}
	\end{subfigure}
	
	\begin{subfigure}[t]{0.1\textwidth}
	\hspace{-.3mm}
    \centering
    \includegraphics[width=1\linewidth]{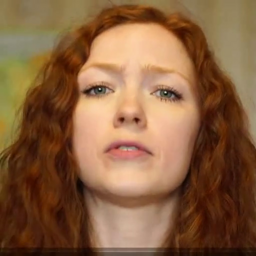}
	\end{subfigure}
	
	\begin{subfigure}[t]{0.1\textwidth}
	\hspace{-.3mm}
    \centering
    \includegraphics[width=1\linewidth]{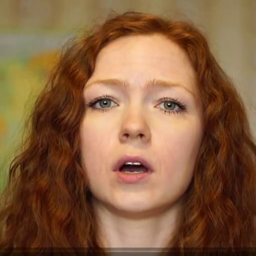}
	\end{subfigure}
	
	\begin{subfigure}[t]{0.1\textwidth}
	\hspace{-.3mm}
    \centering
    \includegraphics[width=1\linewidth]{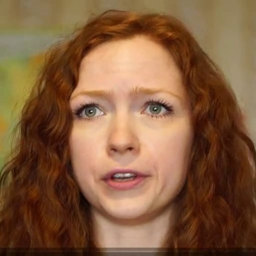}
	\end{subfigure}
	
	\begin{subfigure}[t]{0.1\textwidth}
	\hspace{-.3mm}
    \centering
    \includegraphics[width=1\linewidth]{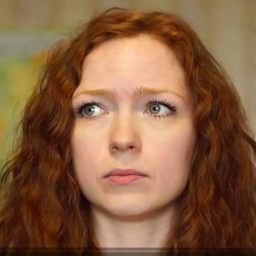}
	\end{subfigure}

	\\
	
	\vspace{-2.5mm}
	
	\begin{subfigure}[t]{0.1\textwidth}
    \hspace{-.3mm}
	\centering
    \includegraphics[width=1\linewidth]{figures/reenactment/samples/FOMM.png}
	\end{subfigure}
	
	&
	
	\begin{subfigure}[t]{0.1\textwidth}
	\hspace{-.3mm}
    \centering
    \includegraphics[width=1\linewidth]{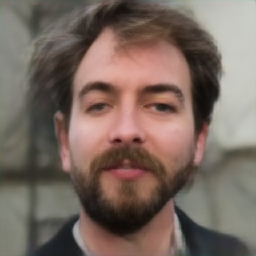}
	\end{subfigure}
	
	\begin{subfigure}[t]{0.1\textwidth}
	\hspace{-.3mm}
    \centering
    \includegraphics[width=1\linewidth]{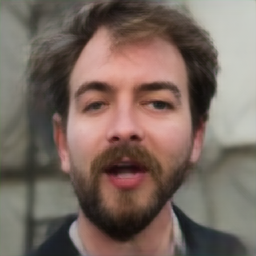}
	\end{subfigure}
	
	\begin{subfigure}[t]{0.1\textwidth}
	\hspace{-.3mm}
    \centering
    \includegraphics[width=1\linewidth]{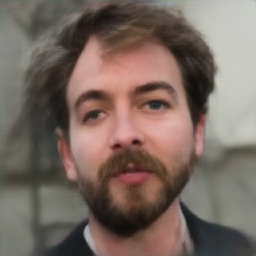}
	\end{subfigure}
	
	\begin{subfigure}[t]{0.1\textwidth}
	\hspace{-.3mm}
    \centering
    \includegraphics[width=1\linewidth]{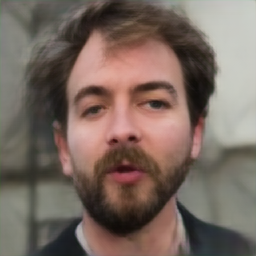}
	\end{subfigure}
	
	\begin{subfigure}[t]{0.1\textwidth}
	\hspace{-.3mm}
    \centering
    \includegraphics[width=1\linewidth]{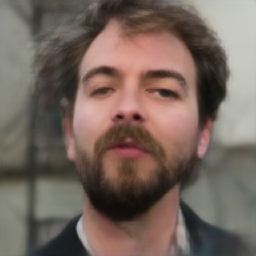}
	\end{subfigure}
	
	\begin{subfigure}[t]{0.1\textwidth}
	\hspace{-.3mm}
    \centering
    \includegraphics[width=1\linewidth]{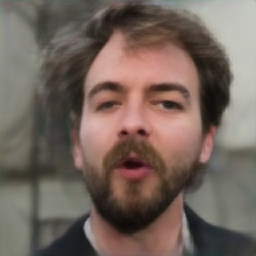}
	\end{subfigure}
	
	\begin{subfigure}[t]{0.1\textwidth}
	\hspace{-.3mm}
    \centering
    \includegraphics[width=1\linewidth]{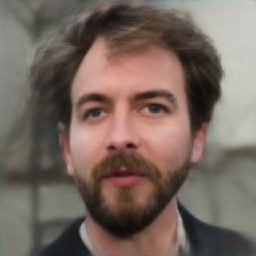}
	\end{subfigure}
	
	\begin{subfigure}[t]{0.1\textwidth}
	\hspace{-.3mm}
    \centering
    \includegraphics[width=1\linewidth]{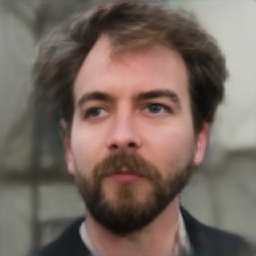}
	\end{subfigure}

	\\
	
	\vspace{-2.5mm}
	
	\begin{subfigure}[t]{0.1\textwidth}
    \hspace{-.3mm}
	\centering
    \includegraphics[width=1\linewidth]{figures/reenactment/samples/DaGAN.png}
	\end{subfigure}
	
	&
	
	\begin{subfigure}[t]{0.1\textwidth}
	\hspace{-.3mm}
    \centering
    \includegraphics[width=1\linewidth]{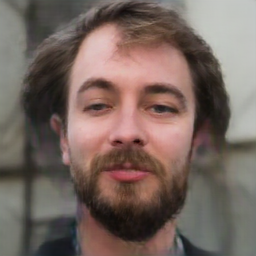}
	\end{subfigure}
	
	\begin{subfigure}[t]{0.1\textwidth}
	\hspace{-.3mm}
    \centering
    \includegraphics[width=1\linewidth]{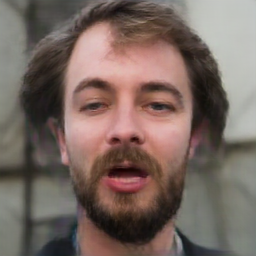}
	\end{subfigure}
	
	\begin{subfigure}[t]{0.1\textwidth}
	\hspace{-.3mm}
    \centering
    \includegraphics[width=1\linewidth]{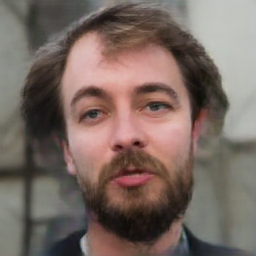}
	\end{subfigure}
	
	\begin{subfigure}[t]{0.1\textwidth}
	\hspace{-.3mm}
    \centering
    \includegraphics[width=1\linewidth]{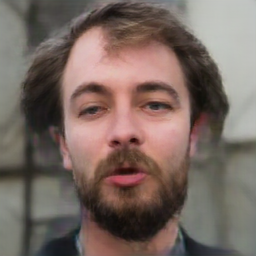}
	\end{subfigure}
	
	\begin{subfigure}[t]{0.1\textwidth}
	\hspace{-.3mm}
    \centering
    \includegraphics[width=1\linewidth]{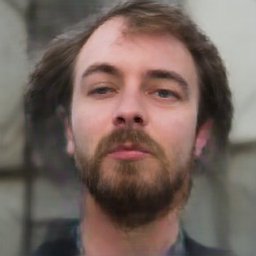}
	\end{subfigure}
	
	\begin{subfigure}[t]{0.1\textwidth}
	\hspace{-.3mm}
    \centering
    \includegraphics[width=1\linewidth]{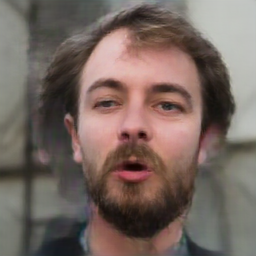}
	\end{subfigure}
	
	\begin{subfigure}[t]{0.1\textwidth}
	\hspace{-.3mm}
    \centering
    \includegraphics[width=1\linewidth]{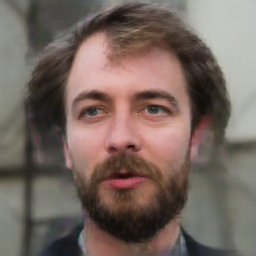}
	\end{subfigure}
	
	\begin{subfigure}[t]{0.1\textwidth}
	\hspace{-.3mm}
    \centering
    \includegraphics[width=1\linewidth]{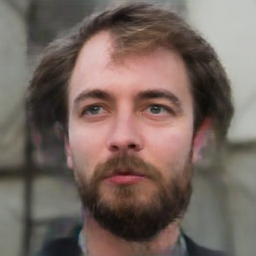}
	\end{subfigure}

	\\
	
	\vspace{-2.5mm}
	
	\begin{subfigure}[t]{0.1\textwidth}
    \hspace{-.3mm}
	\centering
    \includegraphics[width=1\linewidth]{figures/reenactment/samples/MCNET.png}
	\end{subfigure}
	
	&
	
	\begin{subfigure}[t]{0.1\textwidth}
	\hspace{-.3mm}
    \centering
    \includegraphics[width=1\linewidth]{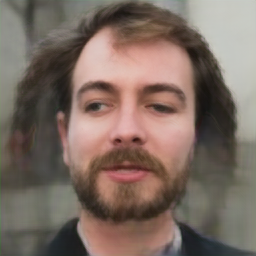}
	\end{subfigure}
	
	\begin{subfigure}[t]{0.1\textwidth}
	\hspace{-.3mm}
    \centering
    \includegraphics[width=1\linewidth]{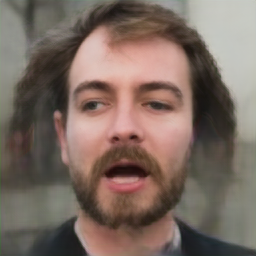}
	\end{subfigure}
	
	\begin{subfigure}[t]{0.1\textwidth}
	\hspace{-.3mm}
    \centering
    \includegraphics[width=1\linewidth]{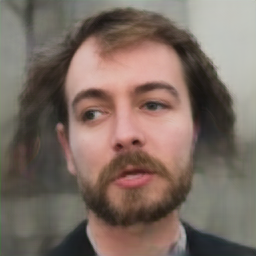}
	\end{subfigure}
	
	\begin{subfigure}[t]{0.1\textwidth}
	\hspace{-.3mm}
    \centering
    \includegraphics[width=1\linewidth]{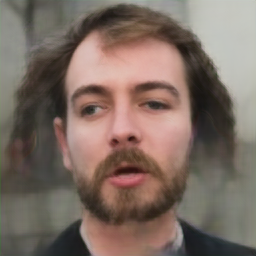}
	\end{subfigure}
	
	\begin{subfigure}[t]{0.1\textwidth}
	\hspace{-.3mm}
    \centering
    \includegraphics[width=1\linewidth]{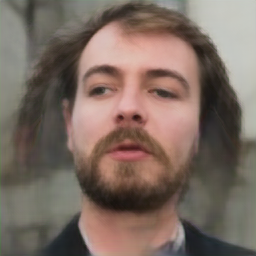}
	\end{subfigure}
	
	\begin{subfigure}[t]{0.1\textwidth}
	\hspace{-.3mm}
    \centering
    \includegraphics[width=1\linewidth]{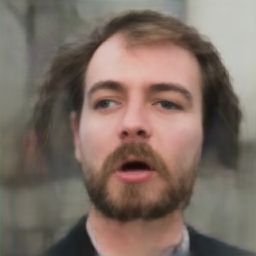}
	\end{subfigure}
	
	\begin{subfigure}[t]{0.1\textwidth}
	\hspace{-.3mm}
    \centering
    \includegraphics[width=1\linewidth]{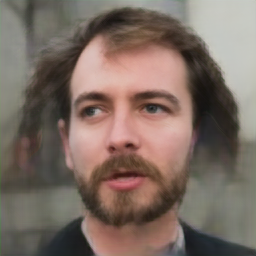}
	\end{subfigure}
	
	\begin{subfigure}[t]{0.1\textwidth}
	\hspace{-.3mm}
    \centering
    \includegraphics[width=1\linewidth]{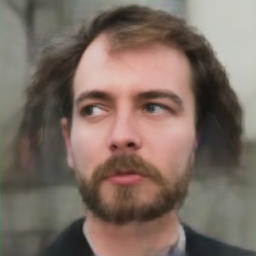}
	\end{subfigure}

	\\
	
	\begin{subfigure}[t]{0.1\textwidth}
    \hspace{-.3mm}
	\centering
    \includegraphics[width=1\linewidth]{figures/reenactment/samples/Ours.png}
	\end{subfigure}
	
	&
	
	\begin{subfigure}[t]{0.1\textwidth}
	\hspace{-.3mm}
    \centering
    \includegraphics[width=1\linewidth]{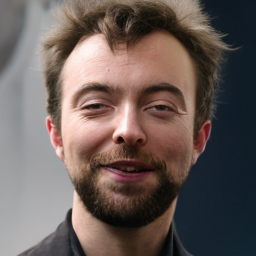}
	\end{subfigure}
	
	\begin{subfigure}[t]{0.1\textwidth}
	\hspace{-.3mm}
    \centering
    \includegraphics[width=1\linewidth]{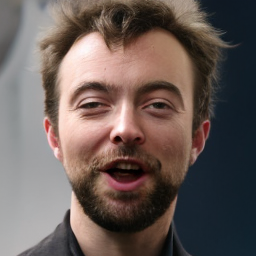}
	\end{subfigure}
	
	\begin{subfigure}[t]{0.1\textwidth}
	\hspace{-.3mm}
    \centering
    \includegraphics[width=1\linewidth]{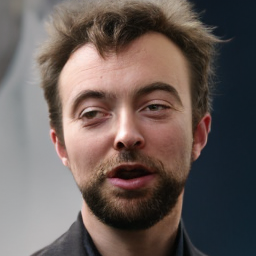}
	\end{subfigure}
	
	\begin{subfigure}[t]{0.1\textwidth}
	\hspace{-.3mm}
    \centering
    \includegraphics[width=1\linewidth]{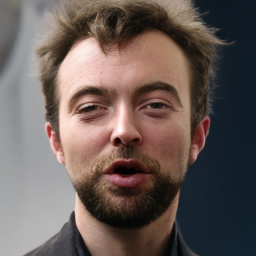}
	\end{subfigure}
	
	\begin{subfigure}[t]{0.1\textwidth}
	\hspace{-.3mm}
    \centering
    \includegraphics[width=1\linewidth]{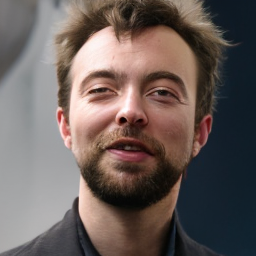}
	\end{subfigure}
	
	\begin{subfigure}[t]{0.1\textwidth}
	\hspace{-.3mm}
    \centering
    \includegraphics[width=1\linewidth]{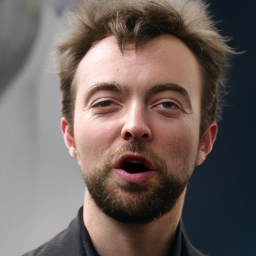}
	\end{subfigure}
	
	\begin{subfigure}[t]{0.1\textwidth}
	\hspace{-.3mm}
    \centering
    \includegraphics[width=1\linewidth]{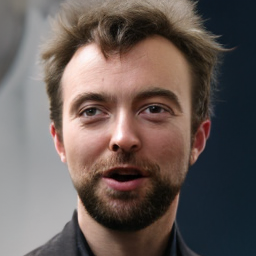}
	\end{subfigure}
	
	\begin{subfigure}[t]{0.1\textwidth}
	\hspace{-.3mm}
    \centering
    \includegraphics[width=1\linewidth]{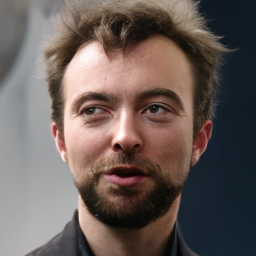}
	\end{subfigure}
	
	\\

\end{tabular}
	
\caption{\textbf{Cross-identity Reenactment.} Comparisons with the competing methods~\cite{siarohin2019first, hong2022depth, hong2023implicit}.}
\label{fig:reenactment_sm7}
\end{figure*}

\begin{figure*}[ht!]
	\centering
	\captionsetup[subfigure]{labelformat=empty,justification=centering,aboveskip=1pt,belowskip=1pt}
    
    \begin{tabular}[c]{c c}
    
    \vspace{-2.5mm}
    
    \begin{subfigure}[t]{0.1\textwidth}
    \hspace{-.3mm}
	\centering
    Source
	\end{subfigure}
	
	&
	
	\begin{subfigure}[t]{0.8\textwidth}
    \hspace{-.3mm}
	\centering
    Driving video
	\end{subfigure}
	
	\\
	
    \vspace{-2.5mm}
    
    \begin{subfigure}[t]{0.1\textwidth}
    \hspace{-.3mm}
	\centering
    \includegraphics[width=1\linewidth]{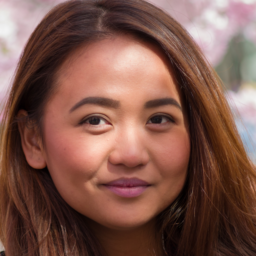}
	\end{subfigure}
	
	&
	
	\begin{subfigure}[t]{0.1\textwidth}
	\hspace{-.3mm}
    \centering
    \includegraphics[width=1\linewidth]{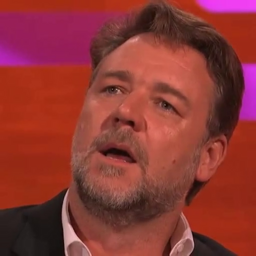}
	\end{subfigure}
	
	\begin{subfigure}[t]{0.1\textwidth}
	\hspace{-.3mm}
    \centering
    \includegraphics[width=1\linewidth]{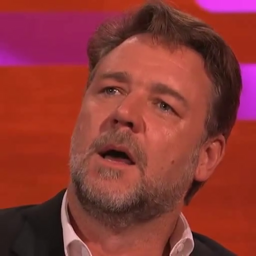}
	\end{subfigure}
	
	\begin{subfigure}[t]{0.1\textwidth}
	\hspace{-.3mm}
    \centering
    \includegraphics[width=1\linewidth]{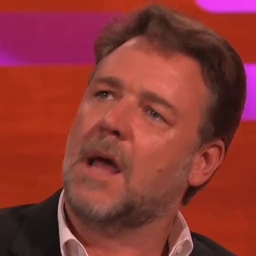}
	\end{subfigure}
	
	\begin{subfigure}[t]{0.1\textwidth}
	\hspace{-.3mm}
    \centering
    \includegraphics[width=1\linewidth]{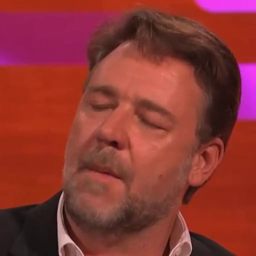}
	\end{subfigure}
	
	\begin{subfigure}[t]{0.1\textwidth}
	\hspace{-.3mm}
    \centering
    \includegraphics[width=1\linewidth]{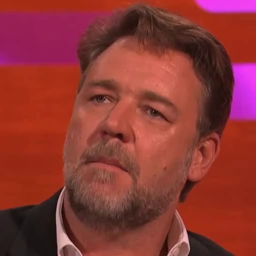}
	\end{subfigure}
	
	\begin{subfigure}[t]{0.1\textwidth}
	\hspace{-.3mm}
    \centering
    \includegraphics[width=1\linewidth]{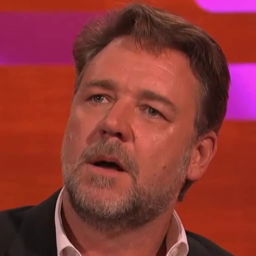}
	\end{subfigure}
	
	\begin{subfigure}[t]{0.1\textwidth}
	\hspace{-.3mm}
    \centering
    \includegraphics[width=1\linewidth]{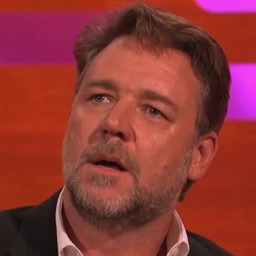}
	\end{subfigure}
	
	\begin{subfigure}[t]{0.1\textwidth}
	\hspace{-.3mm}
    \centering
    \includegraphics[width=1\linewidth]{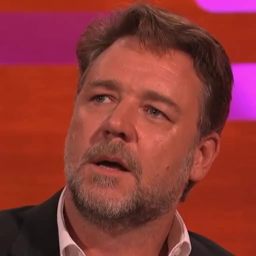}
	\end{subfigure}

	\\
	
	\vspace{-2.5mm}
	
	\begin{subfigure}[t]{0.1\textwidth}
    \hspace{-.3mm}
	\centering
    \includegraphics[width=1\linewidth]{figures/reenactment/samples/FOMM.png}
	\end{subfigure}
	
	&
	
	\begin{subfigure}[t]{0.1\textwidth}
	\hspace{-.3mm}
    \centering
    \includegraphics[width=1\linewidth]{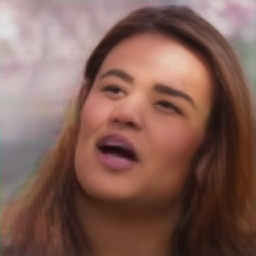}
	\end{subfigure}
	
	\begin{subfigure}[t]{0.1\textwidth}
	\hspace{-.3mm}
    \centering
    \includegraphics[width=1\linewidth]{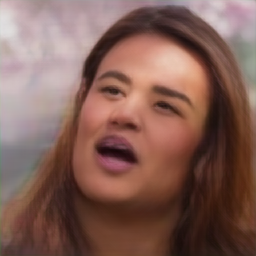}
	\end{subfigure}
	
	\begin{subfigure}[t]{0.1\textwidth}
	\hspace{-.3mm}
    \centering
    \includegraphics[width=1\linewidth]{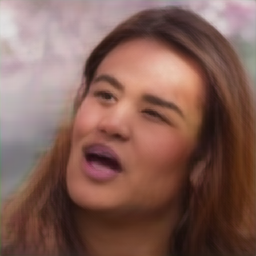}
	\end{subfigure}
	
	\begin{subfigure}[t]{0.1\textwidth}
	\hspace{-.3mm}
    \centering
    \includegraphics[width=1\linewidth]{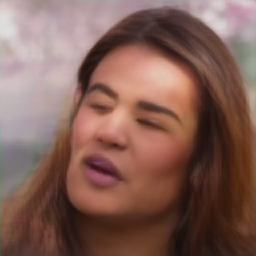}
	\end{subfigure}
	
	\begin{subfigure}[t]{0.1\textwidth}
	\hspace{-.3mm}
    \centering
    \includegraphics[width=1\linewidth]{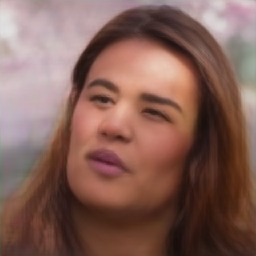}
	\end{subfigure}
	
	\begin{subfigure}[t]{0.1\textwidth}
	\hspace{-.3mm}
    \centering
    \includegraphics[width=1\linewidth]{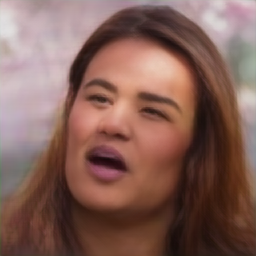}
	\end{subfigure}
	
	\begin{subfigure}[t]{0.1\textwidth}
	\hspace{-.3mm}
    \centering
    \includegraphics[width=1\linewidth]{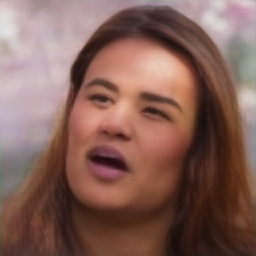}
	\end{subfigure}
	
	\begin{subfigure}[t]{0.1\textwidth}
	\hspace{-.3mm}
    \centering
    \includegraphics[width=1\linewidth]{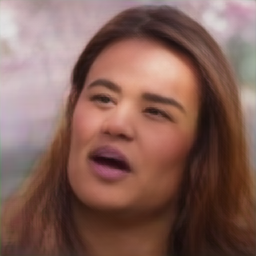}
	\end{subfigure}

	\\
	
	\vspace{-2.5mm}
	
	\begin{subfigure}[t]{0.1\textwidth}
    \hspace{-.3mm}
	\centering
    \includegraphics[width=1\linewidth]{figures/reenactment/samples/DaGAN.png}
	\end{subfigure}
	
	&
	
	\begin{subfigure}[t]{0.1\textwidth}
	\hspace{-.3mm}
    \centering
    \includegraphics[width=1\linewidth]{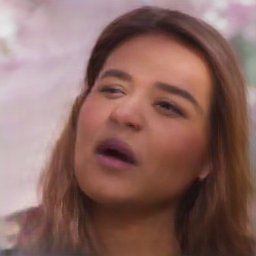}
	\end{subfigure}
	
	\begin{subfigure}[t]{0.1\textwidth}
	\hspace{-.3mm}
    \centering
    \includegraphics[width=1\linewidth]{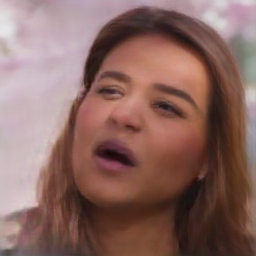}
	\end{subfigure}
	
	\begin{subfigure}[t]{0.1\textwidth}
	\hspace{-.3mm}
    \centering
    \includegraphics[width=1\linewidth]{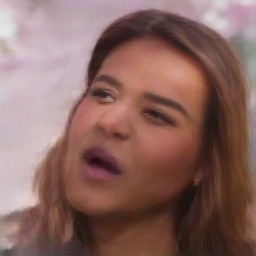}
	\end{subfigure}
	
	\begin{subfigure}[t]{0.1\textwidth}
	\hspace{-.3mm}
    \centering
    \includegraphics[width=1\linewidth]{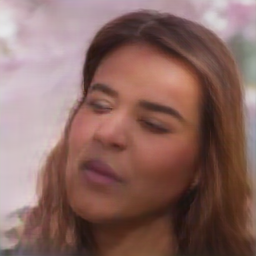}
	\end{subfigure}
	
	\begin{subfigure}[t]{0.1\textwidth}
	\hspace{-.3mm}
    \centering
    \includegraphics[width=1\linewidth]{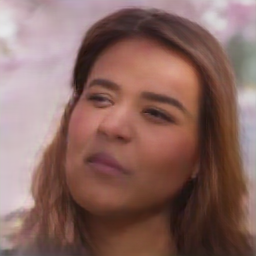}
	\end{subfigure}
	
	\begin{subfigure}[t]{0.1\textwidth}
	\hspace{-.3mm}
    \centering
    \includegraphics[width=1\linewidth]{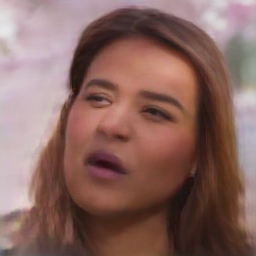}
	\end{subfigure}
	
	\begin{subfigure}[t]{0.1\textwidth}
	\hspace{-.3mm}
    \centering
    \includegraphics[width=1\linewidth]{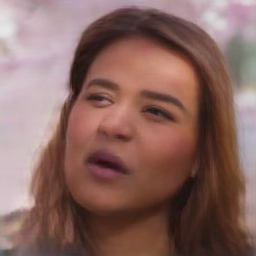}
	\end{subfigure}
	
	\begin{subfigure}[t]{0.1\textwidth}
	\hspace{-.3mm}
    \centering
    \includegraphics[width=1\linewidth]{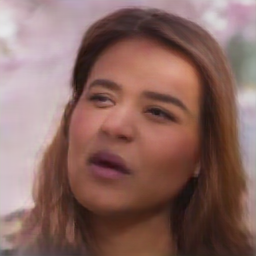}
	\end{subfigure}

	\\
	
	\vspace{-2.5mm}
	
	\begin{subfigure}[t]{0.1\textwidth}
    \hspace{-.3mm}
	\centering
    \includegraphics[width=1\linewidth]{figures/reenactment/samples/MCNET.png}
	\end{subfigure}
	
	&
	
	\begin{subfigure}[t]{0.1\textwidth}
	\hspace{-.3mm}
    \centering
    \includegraphics[width=1\linewidth]{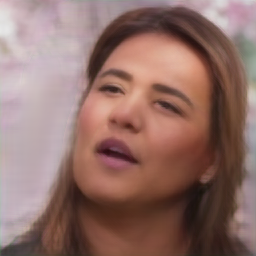}
	\end{subfigure}
	
	\begin{subfigure}[t]{0.1\textwidth}
	\hspace{-.3mm}
    \centering
    \includegraphics[width=1\linewidth]{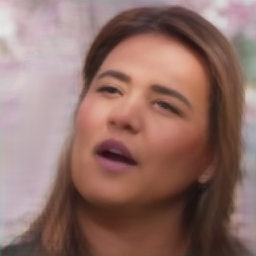}
	\end{subfigure}
	
	\begin{subfigure}[t]{0.1\textwidth}
	\hspace{-.3mm}
    \centering
    \includegraphics[width=1\linewidth]{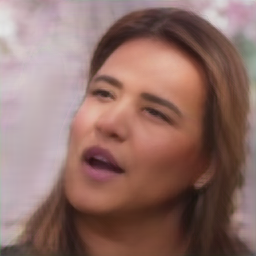}
	\end{subfigure}
	
	\begin{subfigure}[t]{0.1\textwidth}
	\hspace{-.3mm}
    \centering
    \includegraphics[width=1\linewidth]{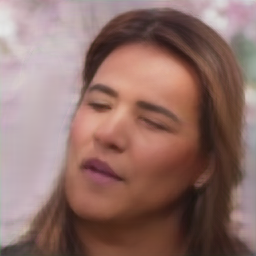}
	\end{subfigure}
	
	\begin{subfigure}[t]{0.1\textwidth}
	\hspace{-.3mm}
    \centering
    \includegraphics[width=1\linewidth]{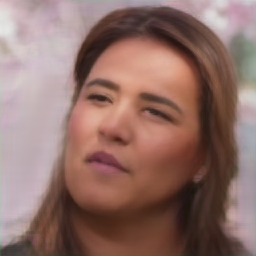}
	\end{subfigure}
	
	\begin{subfigure}[t]{0.1\textwidth}
	\hspace{-.3mm}
    \centering
    \includegraphics[width=1\linewidth]{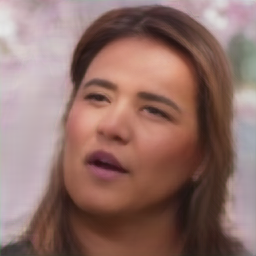}
	\end{subfigure}
	
	\begin{subfigure}[t]{0.1\textwidth}
	\hspace{-.3mm}
    \centering
    \includegraphics[width=1\linewidth]{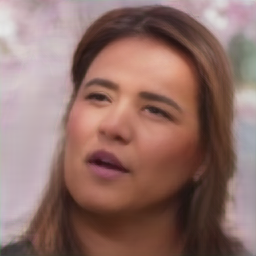}
	\end{subfigure}
	
	\begin{subfigure}[t]{0.1\textwidth}
	\hspace{-.3mm}
    \centering
    \includegraphics[width=1\linewidth]{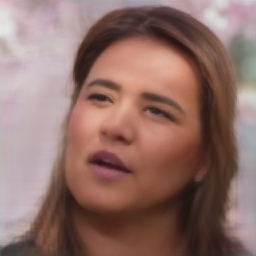}
	\end{subfigure}

	\\
	
	\begin{subfigure}[t]{0.1\textwidth}
    \hspace{-.3mm}
	\centering
    \includegraphics[width=1\linewidth]{figures/reenactment/samples/Ours.png}
	\end{subfigure}
	
	&
	
	\begin{subfigure}[t]{0.1\textwidth}
	\hspace{-.3mm}
    \centering
    \includegraphics[width=1\linewidth]{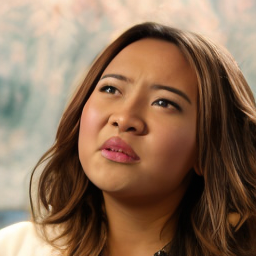}
	\end{subfigure}
	
	\begin{subfigure}[t]{0.1\textwidth}
	\hspace{-.3mm}
    \centering
    \includegraphics[width=1\linewidth]{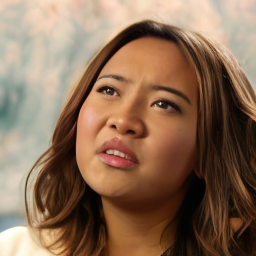}
	\end{subfigure}
	
	\begin{subfigure}[t]{0.1\textwidth}
	\hspace{-.3mm}
    \centering
    \includegraphics[width=1\linewidth]{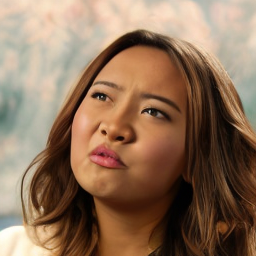}
	\end{subfigure}
	
	\begin{subfigure}[t]{0.1\textwidth}
	\hspace{-.3mm}
    \centering
    \includegraphics[width=1\linewidth]{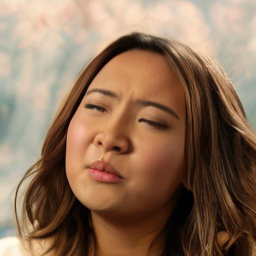}
	\end{subfigure}
	
	\begin{subfigure}[t]{0.1\textwidth}
	\hspace{-.3mm}
    \centering
    \includegraphics[width=1\linewidth]{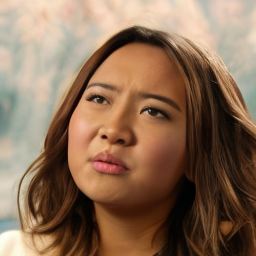}
	\end{subfigure}
	
	\begin{subfigure}[t]{0.1\textwidth}
	\hspace{-.3mm}
    \centering
    \includegraphics[width=1\linewidth]{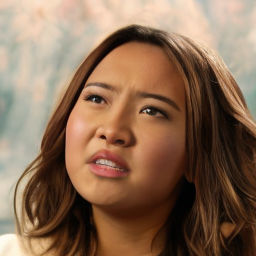}
	\end{subfigure}
	
	\begin{subfigure}[t]{0.1\textwidth}
	\hspace{-.3mm}
    \centering
    \includegraphics[width=1\linewidth]{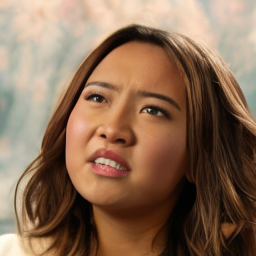}
	\end{subfigure}
	
	\begin{subfigure}[t]{0.1\textwidth}
	\hspace{-.3mm}
    \centering
    \includegraphics[width=1\linewidth]{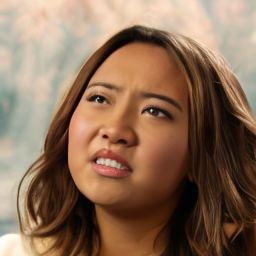}
	\end{subfigure}
	
	\\

\end{tabular}
	
\caption{\textbf{Cross-identity Reenactment.} Comparisons with the competing methods~\cite{siarohin2019first, hong2022depth, hong2023implicit}.}
\label{fig:reenactment_sm2}
\end{figure*}

\begin{figure*}[ht!]
	\centering
	\captionsetup[subfigure]{labelformat=empty,justification=centering,aboveskip=1pt,belowskip=1pt}
    
    \begin{tabular}[c]{c c}
    
    \vspace{-2.5mm}
    
    \begin{subfigure}[t]{0.1\textwidth}
    \hspace{-.3mm}
	\centering
    Source
	\end{subfigure}
	
	&
	
	\begin{subfigure}[t]{0.8\textwidth}
    \hspace{-.3mm}
	\centering
    Driving video
	\end{subfigure}
	
	\\
	
    \vspace{-2.5mm}
    
    \begin{subfigure}[t]{0.1\textwidth}
    \hspace{-.3mm}
	\centering
    \includegraphics[width=1\linewidth]{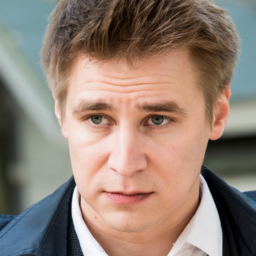}
	\end{subfigure}
	
	&
	
	\begin{subfigure}[t]{0.1\textwidth}
	\hspace{-.3mm}
    \centering
    \includegraphics[width=1\linewidth]{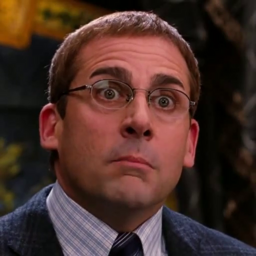}
	\end{subfigure}
	
	\begin{subfigure}[t]{0.1\textwidth}
	\hspace{-.3mm}
    \centering
    \includegraphics[width=1\linewidth]{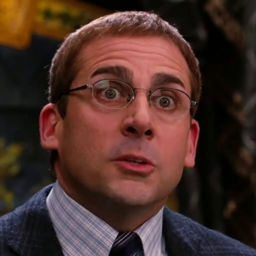}
	\end{subfigure}
	
	\begin{subfigure}[t]{0.1\textwidth}
	\hspace{-.3mm}
    \centering
    \includegraphics[width=1\linewidth]{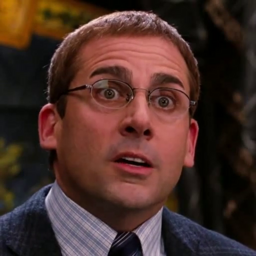}
	\end{subfigure}
	
	\begin{subfigure}[t]{0.1\textwidth}
	\hspace{-.3mm}
    \centering
    \includegraphics[width=1\linewidth]{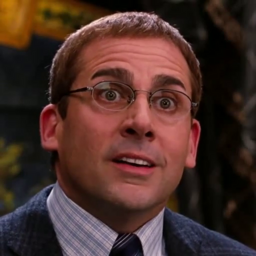}
	\end{subfigure}
	
	\begin{subfigure}[t]{0.1\textwidth}
	\hspace{-.3mm}
    \centering
    \includegraphics[width=1\linewidth]{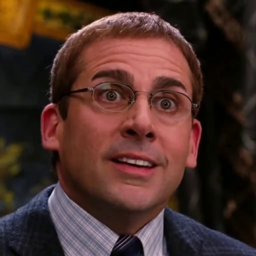}
	\end{subfigure}
	
	\begin{subfigure}[t]{0.1\textwidth}
	\hspace{-.3mm}
    \centering
    \includegraphics[width=1\linewidth]{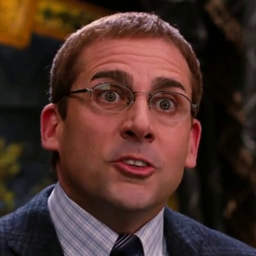}
	\end{subfigure}
	
	\begin{subfigure}[t]{0.1\textwidth}
	\hspace{-.3mm}
    \centering
    \includegraphics[width=1\linewidth]{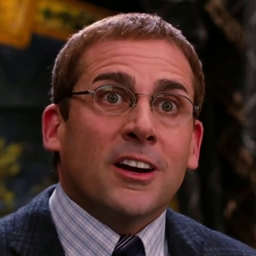}
	\end{subfigure}
	
	\begin{subfigure}[t]{0.1\textwidth}
	\hspace{-.3mm}
    \centering
    \includegraphics[width=1\linewidth]{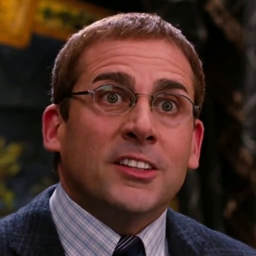}
	\end{subfigure}

	\\
	
	\vspace{-2.5mm}
	
	\begin{subfigure}[t]{0.1\textwidth}
    \hspace{-.3mm}
	\centering
    \includegraphics[width=1\linewidth]{figures/reenactment/samples/FOMM.png}
	\end{subfigure}
	
	&
	
	\begin{subfigure}[t]{0.1\textwidth}
	\hspace{-.3mm}
    \centering
    \includegraphics[width=1\linewidth]{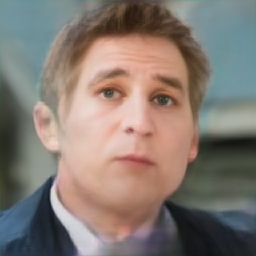}
	\end{subfigure}
	
	\begin{subfigure}[t]{0.1\textwidth}
	\hspace{-.3mm}
    \centering
    \includegraphics[width=1\linewidth]{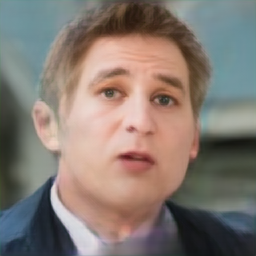}
	\end{subfigure}
	
	\begin{subfigure}[t]{0.1\textwidth}
	\hspace{-.3mm}
    \centering
    \includegraphics[width=1\linewidth]{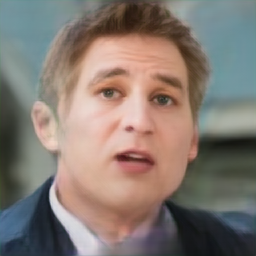}
	\end{subfigure}
	
	\begin{subfigure}[t]{0.1\textwidth}
	\hspace{-.3mm}
    \centering
    \includegraphics[width=1\linewidth]{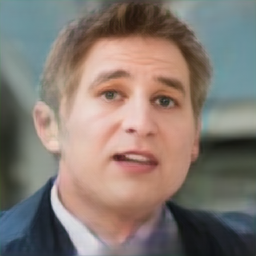}
	\end{subfigure}
	
	\begin{subfigure}[t]{0.1\textwidth}
	\hspace{-.3mm}
    \centering
    \includegraphics[width=1\linewidth]{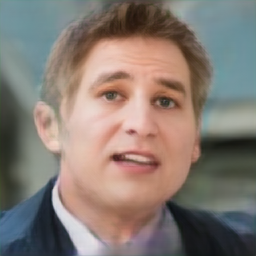}
	\end{subfigure}
	
	\begin{subfigure}[t]{0.1\textwidth}
	\hspace{-.3mm}
    \centering
    \includegraphics[width=1\linewidth]{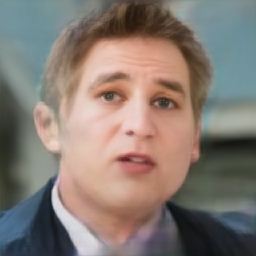}
	\end{subfigure}
	
	\begin{subfigure}[t]{0.1\textwidth}
	\hspace{-.3mm}
    \centering
    \includegraphics[width=1\linewidth]{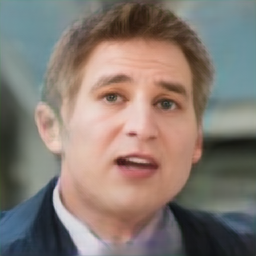}
	\end{subfigure}
	
	\begin{subfigure}[t]{0.1\textwidth}
	\hspace{-.3mm}
    \centering
    \includegraphics[width=1\linewidth]{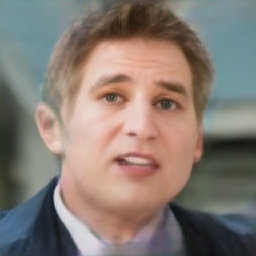}
	\end{subfigure}

	\\
	
	\vspace{-2.5mm}
	
	\begin{subfigure}[t]{0.1\textwidth}
    \hspace{-.3mm}
	\centering
    \includegraphics[width=1\linewidth]{figures/reenactment/samples/DaGAN.png}
	\end{subfigure}
	
	&
	
	\begin{subfigure}[t]{0.1\textwidth}
	\hspace{-.3mm}
    \centering
    \includegraphics[width=1\linewidth]{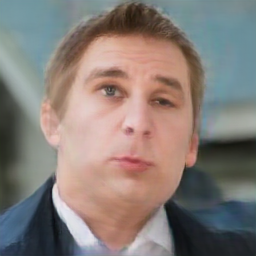}
	\end{subfigure}
	
	\begin{subfigure}[t]{0.1\textwidth}
	\hspace{-.3mm}
    \centering
    \includegraphics[width=1\linewidth]{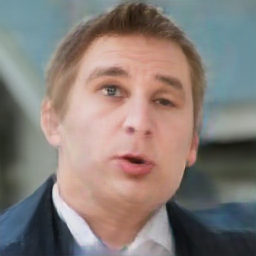}
	\end{subfigure}
	
	\begin{subfigure}[t]{0.1\textwidth}
	\hspace{-.3mm}
    \centering
    \includegraphics[width=1\linewidth]{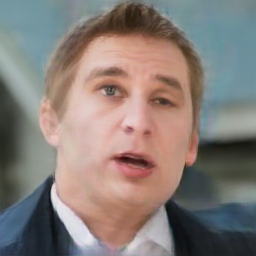}
	\end{subfigure}
	
	\begin{subfigure}[t]{0.1\textwidth}
	\hspace{-.3mm}
    \centering
    \includegraphics[width=1\linewidth]{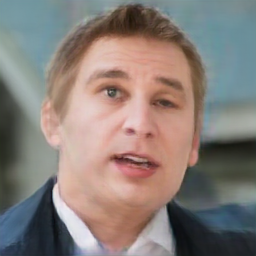}
	\end{subfigure}
	
	\begin{subfigure}[t]{0.1\textwidth}
	\hspace{-.3mm}
    \centering
    \includegraphics[width=1\linewidth]{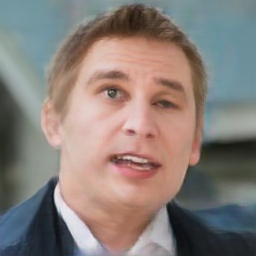}
	\end{subfigure}
	
	\begin{subfigure}[t]{0.1\textwidth}
	\hspace{-.3mm}
    \centering
    \includegraphics[width=1\linewidth]{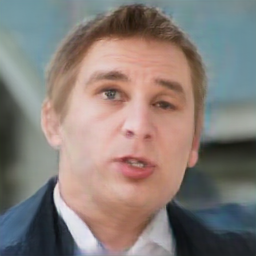}
	\end{subfigure}
	
	\begin{subfigure}[t]{0.1\textwidth}
	\hspace{-.3mm}
    \centering
    \includegraphics[width=1\linewidth]{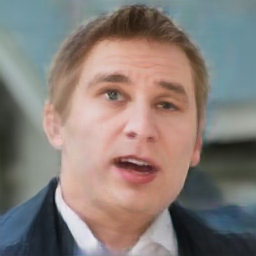}
	\end{subfigure}
	
	\begin{subfigure}[t]{0.1\textwidth}
	\hspace{-.3mm}
    \centering
    \includegraphics[width=1\linewidth]{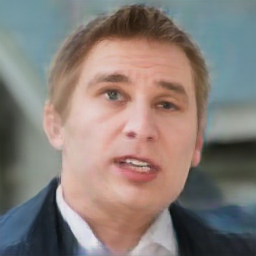}
	\end{subfigure}

	\\
	
	\vspace{-2.5mm}
	
	\begin{subfigure}[t]{0.1\textwidth}
    \hspace{-.3mm}
	\centering
    \includegraphics[width=1\linewidth]{figures/reenactment/samples/MCNET.png}
	\end{subfigure}
	
	&
	
	\begin{subfigure}[t]{0.1\textwidth}
	\hspace{-.3mm}
    \centering
    \includegraphics[width=1\linewidth]{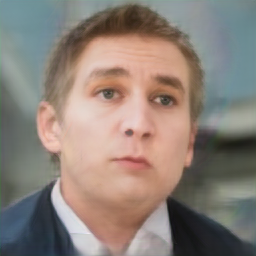}
	\end{subfigure}
	
	\begin{subfigure}[t]{0.1\textwidth}
	\hspace{-.3mm}
    \centering
    \includegraphics[width=1\linewidth]{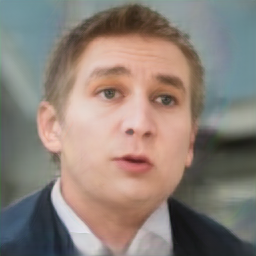}
	\end{subfigure}
	
	\begin{subfigure}[t]{0.1\textwidth}
	\hspace{-.3mm}
    \centering
    \includegraphics[width=1\linewidth]{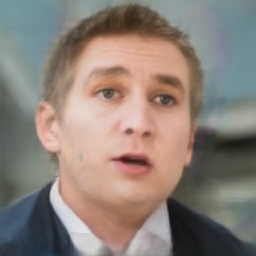}
	\end{subfigure}
	
	\begin{subfigure}[t]{0.1\textwidth}
	\hspace{-.3mm}
    \centering
    \includegraphics[width=1\linewidth]{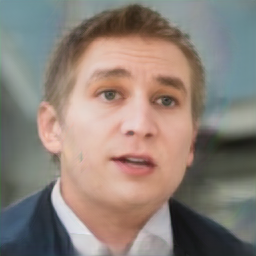}
	\end{subfigure}
	
	\begin{subfigure}[t]{0.1\textwidth}
	\hspace{-.3mm}
    \centering
    \includegraphics[width=1\linewidth]{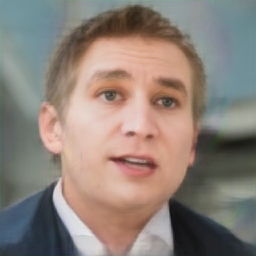}
	\end{subfigure}
	
	\begin{subfigure}[t]{0.1\textwidth}
	\hspace{-.3mm}
    \centering
    \includegraphics[width=1\linewidth]{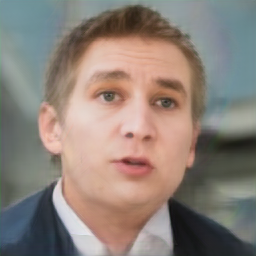}
	\end{subfigure}
	
	\begin{subfigure}[t]{0.1\textwidth}
	\hspace{-.3mm}
    \centering
    \includegraphics[width=1\linewidth]{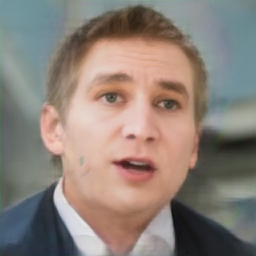}
	\end{subfigure}
	
	\begin{subfigure}[t]{0.1\textwidth}
	\hspace{-.3mm}
    \centering
    \includegraphics[width=1\linewidth]{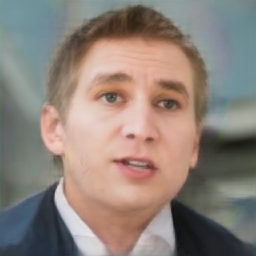}
	\end{subfigure}

	\\
	
	\begin{subfigure}[t]{0.1\textwidth}
    \hspace{-.3mm}
	\centering
    \includegraphics[width=1\linewidth]{figures/reenactment/samples/Ours.png}
	\end{subfigure}
	
	&
	
	\begin{subfigure}[t]{0.1\textwidth}
	\hspace{-.3mm}
    \centering
    \includegraphics[width=1\linewidth]{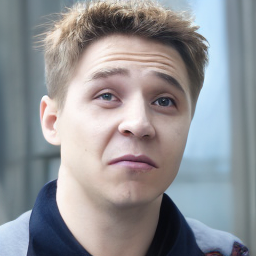}
	\end{subfigure}
	
	\begin{subfigure}[t]{0.1\textwidth}
	\hspace{-.3mm}
    \centering
    \includegraphics[width=1\linewidth]{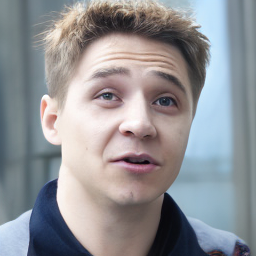}
	\end{subfigure}
	
	\begin{subfigure}[t]{0.1\textwidth}
	\hspace{-.3mm}
    \centering
    \includegraphics[width=1\linewidth]{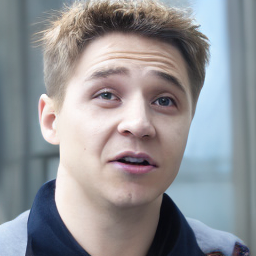}
	\end{subfigure}
	
	\begin{subfigure}[t]{0.1\textwidth}
	\hspace{-.3mm}
    \centering
    \includegraphics[width=1\linewidth]{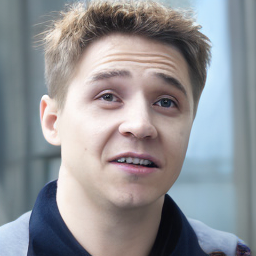}
	\end{subfigure}
	
	\begin{subfigure}[t]{0.1\textwidth}
	\hspace{-.3mm}
    \centering
    \includegraphics[width=1\linewidth]{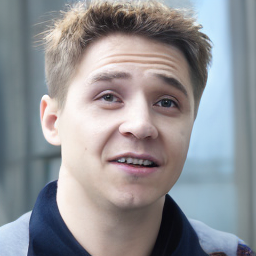}
	\end{subfigure}
	
	\begin{subfigure}[t]{0.1\textwidth}
	\hspace{-.3mm}
    \centering
    \includegraphics[width=1\linewidth]{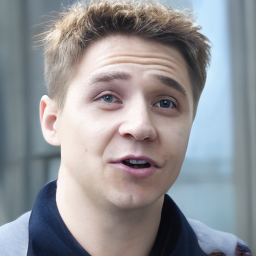}
	\end{subfigure}
	
	\begin{subfigure}[t]{0.1\textwidth}
	\hspace{-.3mm}
    \centering
    \includegraphics[width=1\linewidth]{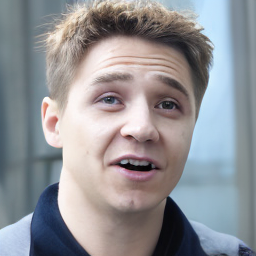}
	\end{subfigure}
	
	\begin{subfigure}[t]{0.1\textwidth}
	\hspace{-.3mm}
    \centering
    \includegraphics[width=1\linewidth]{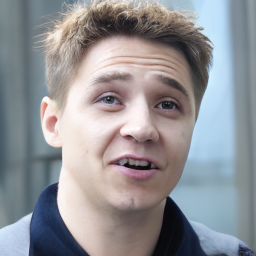}
	\end{subfigure}
	
	\\

\end{tabular}
	
\caption{\textbf{Cross-identity Reenactment.} Comparisons with the competing methods~\cite{siarohin2019first, hong2022depth, hong2023implicit}.}
\label{fig:reenactment_sm5}
\end{figure*}

\begin{figure*}[ht!]
	\centering
	\captionsetup[subfigure]{labelformat=empty,justification=centering,aboveskip=1pt,belowskip=1pt}
    
    \begin{tabular}[c]{c c}
    
    \vspace{-2.5mm}
    
    \begin{subfigure}[t]{0.1\textwidth}
    \hspace{-.3mm}
	\centering
    Source
	\end{subfigure}
	
	&
	
	\begin{subfigure}[t]{0.8\textwidth}
    \hspace{-.3mm}
	\centering
    Driving video
	\end{subfigure}
	
	\\
	
    \vspace{-2.5mm}
    
    \begin{subfigure}[t]{0.1\textwidth}
    \hspace{-.3mm}
	\centering
    \includegraphics[width=1\linewidth]{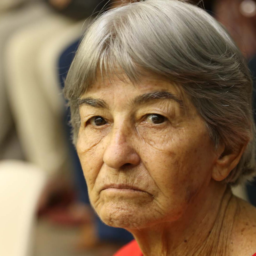}
	\end{subfigure}
	
	&
	
	\begin{subfigure}[t]{0.1\textwidth}
	\hspace{-.3mm}
    \centering
    \includegraphics[width=1\linewidth]{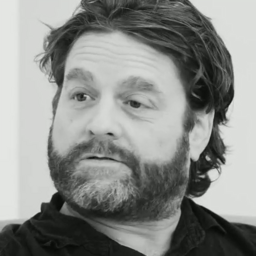}
	\end{subfigure}
	
	\begin{subfigure}[t]{0.1\textwidth}
	\hspace{-.3mm}
    \centering
    \includegraphics[width=1\linewidth]{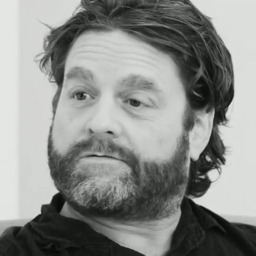}
	\end{subfigure}
	
	\begin{subfigure}[t]{0.1\textwidth}
	\hspace{-.3mm}
    \centering
    \includegraphics[width=1\linewidth]{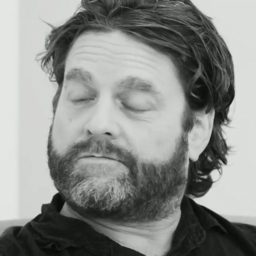}
	\end{subfigure}
	
	\begin{subfigure}[t]{0.1\textwidth}
	\hspace{-.3mm}
    \centering
    \includegraphics[width=1\linewidth]{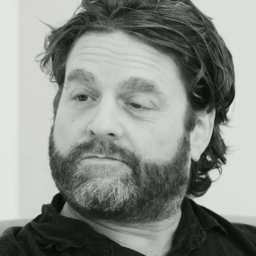}
	\end{subfigure}
	
	\begin{subfigure}[t]{0.1\textwidth}
	\hspace{-.3mm}
    \centering
    \includegraphics[width=1\linewidth]{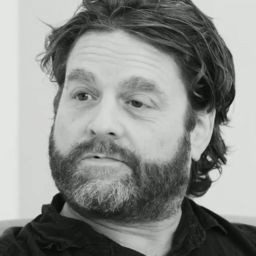}
	\end{subfigure}
	
	\begin{subfigure}[t]{0.1\textwidth}
	\hspace{-.3mm}
    \centering
    \includegraphics[width=1\linewidth]{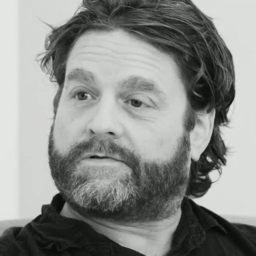}
	\end{subfigure}
	
	\begin{subfigure}[t]{0.1\textwidth}
	\hspace{-.3mm}
    \centering
    \includegraphics[width=1\linewidth]{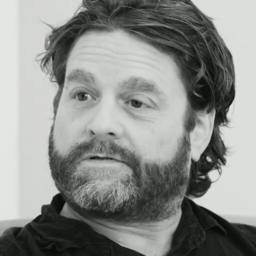}
	\end{subfigure}
	
	\begin{subfigure}[t]{0.1\textwidth}
	\hspace{-.3mm}
    \centering
    \includegraphics[width=1\linewidth]{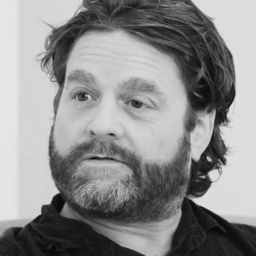}
	\end{subfigure}

	\\
	
	\vspace{-2.5mm}
	
	\begin{subfigure}[t]{0.1\textwidth}
    \hspace{-.3mm}
	\centering
    \includegraphics[width=1\linewidth]{figures/reenactment/samples/FOMM.png}
	\end{subfigure}
	
	&
	
	\begin{subfigure}[t]{0.1\textwidth}
	\hspace{-.3mm}
    \centering
    \includegraphics[width=1\linewidth]{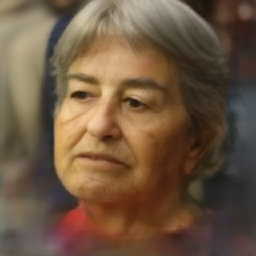}
	\end{subfigure}
	
	\begin{subfigure}[t]{0.1\textwidth}
	\hspace{-.3mm}
    \centering
    \includegraphics[width=1\linewidth]{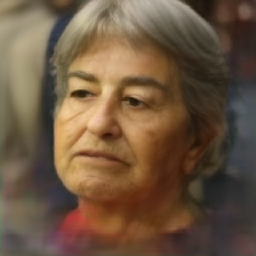}
	\end{subfigure}
	
	\begin{subfigure}[t]{0.1\textwidth}
	\hspace{-.3mm}
    \centering
    \includegraphics[width=1\linewidth]{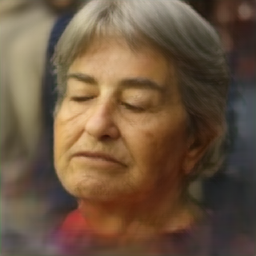}
	\end{subfigure}
	
	\begin{subfigure}[t]{0.1\textwidth}
	\hspace{-.3mm}
    \centering
    \includegraphics[width=1\linewidth]{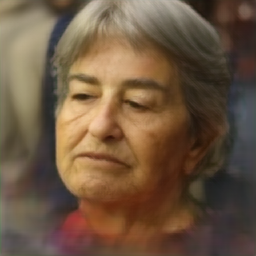}
	\end{subfigure}
	
	\begin{subfigure}[t]{0.1\textwidth}
	\hspace{-.3mm}
    \centering
    \includegraphics[width=1\linewidth]{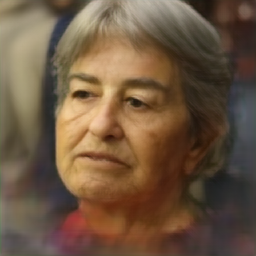}
	\end{subfigure}
	
	\begin{subfigure}[t]{0.1\textwidth}
	\hspace{-.3mm}
    \centering
    \includegraphics[width=1\linewidth]{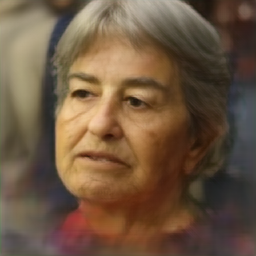}
	\end{subfigure}
	
	\begin{subfigure}[t]{0.1\textwidth}
	\hspace{-.3mm}
    \centering
    \includegraphics[width=1\linewidth]{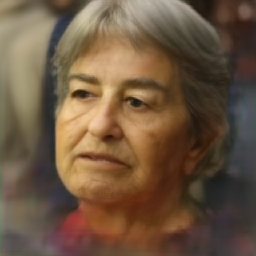}
	\end{subfigure}
	
	\begin{subfigure}[t]{0.1\textwidth}
	\hspace{-.3mm}
    \centering
    \includegraphics[width=1\linewidth]{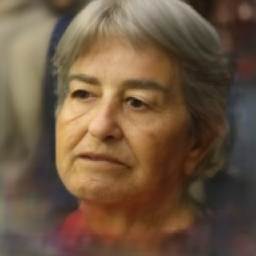}
	\end{subfigure}

	\\
	
	\vspace{-2.5mm}
	
	\begin{subfigure}[t]{0.1\textwidth}
    \hspace{-.3mm}
	\centering
    \includegraphics[width=1\linewidth]{figures/reenactment/samples/DaGAN.png}
	\end{subfigure}
	
	&
	
	\begin{subfigure}[t]{0.1\textwidth}
	\hspace{-.3mm}
    \centering
    \includegraphics[width=1\linewidth]{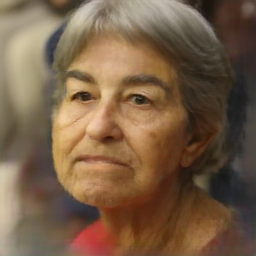}
	\end{subfigure}
	
	\begin{subfigure}[t]{0.1\textwidth}
	\hspace{-.3mm}
    \centering
    \includegraphics[width=1\linewidth]{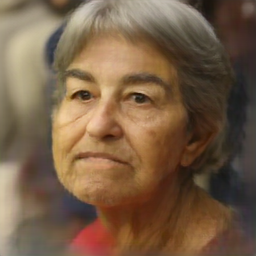}
	\end{subfigure}
	
	\begin{subfigure}[t]{0.1\textwidth}
	\hspace{-.3mm}
    \centering
    \includegraphics[width=1\linewidth]{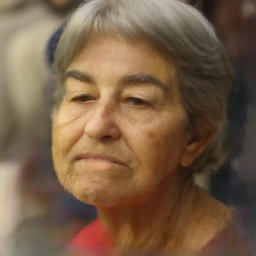}
	\end{subfigure}
	
	\begin{subfigure}[t]{0.1\textwidth}
	\hspace{-.3mm}
    \centering
    \includegraphics[width=1\linewidth]{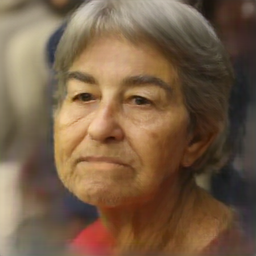}
	\end{subfigure}
	
	\begin{subfigure}[t]{0.1\textwidth}
	\hspace{-.3mm}
    \centering
    \includegraphics[width=1\linewidth]{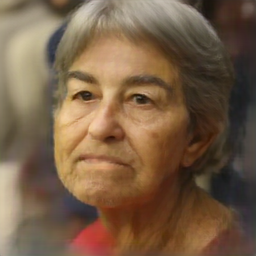}
	\end{subfigure}
	
	\begin{subfigure}[t]{0.1\textwidth}
	\hspace{-.3mm}
    \centering
    \includegraphics[width=1\linewidth]{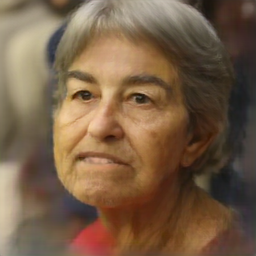}
	\end{subfigure}
	
	\begin{subfigure}[t]{0.1\textwidth}
	\hspace{-.3mm}
    \centering
    \includegraphics[width=1\linewidth]{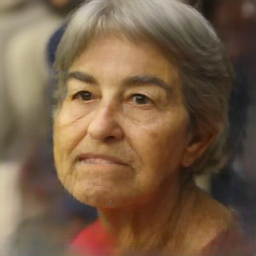}
	\end{subfigure}
	
	\begin{subfigure}[t]{0.1\textwidth}
	\hspace{-.3mm}
    \centering
    \includegraphics[width=1\linewidth]{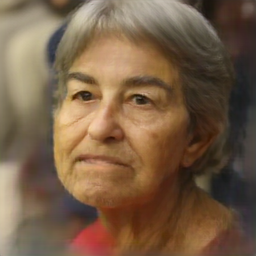}
	\end{subfigure}

	\\
	
	\vspace{-2.5mm}
	
	\begin{subfigure}[t]{0.1\textwidth}
    \hspace{-.3mm}
	\centering
    \includegraphics[width=1\linewidth]{figures/reenactment/samples/MCNET.png}
	\end{subfigure}
	
	&
	
	\begin{subfigure}[t]{0.1\textwidth}
	\hspace{-.3mm}
    \centering
    \includegraphics[width=1\linewidth]{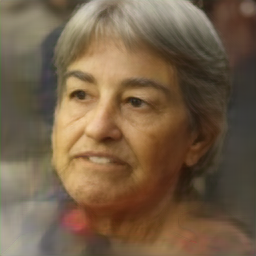}
	\end{subfigure}
	
	\begin{subfigure}[t]{0.1\textwidth}
	\hspace{-.3mm}
    \centering
    \includegraphics[width=1\linewidth]{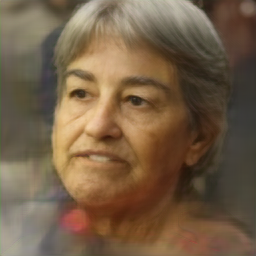}
	\end{subfigure}
	
	\begin{subfigure}[t]{0.1\textwidth}
	\hspace{-.3mm}
    \centering
    \includegraphics[width=1\linewidth]{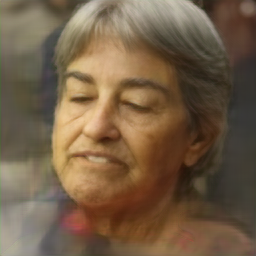}
	\end{subfigure}
	
	\begin{subfigure}[t]{0.1\textwidth}
	\hspace{-.3mm}
    \centering
    \includegraphics[width=1\linewidth]{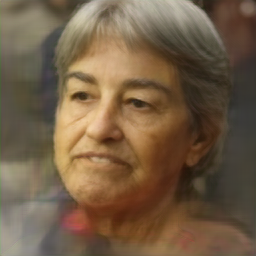}
	\end{subfigure}
	
	\begin{subfigure}[t]{0.1\textwidth}
	\hspace{-.3mm}
    \centering
    \includegraphics[width=1\linewidth]{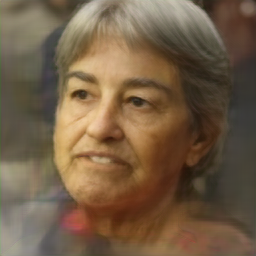}
	\end{subfigure}
	
	\begin{subfigure}[t]{0.1\textwidth}
	\hspace{-.3mm}
    \centering
    \includegraphics[width=1\linewidth]{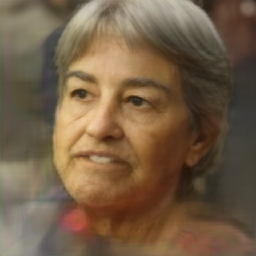}
	\end{subfigure}
	
	\begin{subfigure}[t]{0.1\textwidth}
	\hspace{-.3mm}
    \centering
    \includegraphics[width=1\linewidth]{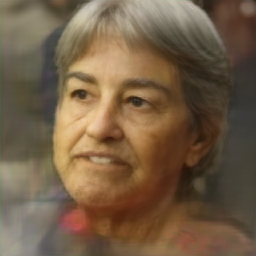}
	\end{subfigure}
	
	\begin{subfigure}[t]{0.1\textwidth}
	\hspace{-.3mm}
    \centering
    \includegraphics[width=1\linewidth]{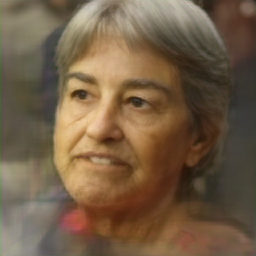}
	\end{subfigure}

	\\
	
	\begin{subfigure}[t]{0.1\textwidth}
    \hspace{-.3mm}
	\centering
    \includegraphics[width=1\linewidth]{figures/reenactment/samples/Ours.png}
	\end{subfigure}
	
	&
	
	\begin{subfigure}[t]{0.1\textwidth}
	\hspace{-.3mm}
    \centering
    \includegraphics[width=1\linewidth]{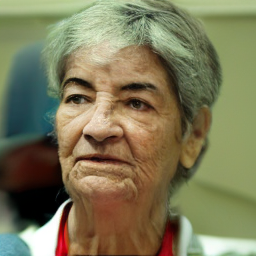}
	\end{subfigure}
	
	\begin{subfigure}[t]{0.1\textwidth}
	\hspace{-.3mm}
    \centering
    \includegraphics[width=1\linewidth]{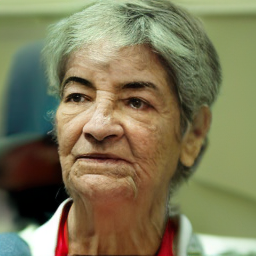}
	\end{subfigure}
	
	\begin{subfigure}[t]{0.1\textwidth}
	\hspace{-.3mm}
    \centering
    \includegraphics[width=1\linewidth]{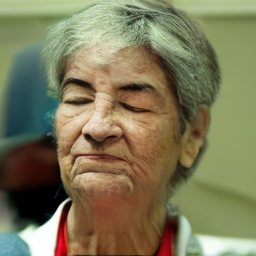}
	\end{subfigure}
	
	\begin{subfigure}[t]{0.1\textwidth}
	\hspace{-.3mm}
    \centering
    \includegraphics[width=1\linewidth]{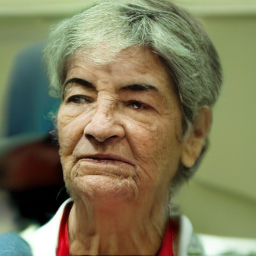}
	\end{subfigure}
	
	\begin{subfigure}[t]{0.1\textwidth}
	\hspace{-.3mm}
    \centering
    \includegraphics[width=1\linewidth]{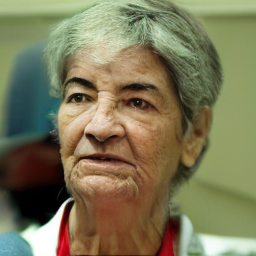}
	\end{subfigure}
	
	\begin{subfigure}[t]{0.1\textwidth}
	\hspace{-.3mm}
    \centering
    \includegraphics[width=1\linewidth]{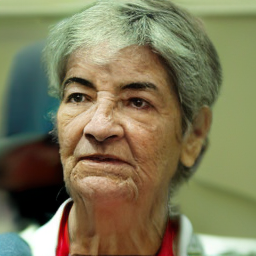}
	\end{subfigure}
	
	\begin{subfigure}[t]{0.1\textwidth}
	\hspace{-.3mm}
    \centering
    \includegraphics[width=1\linewidth]{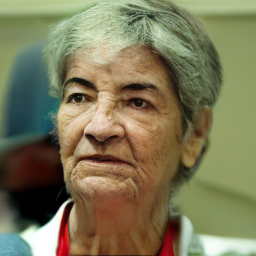}
	\end{subfigure}
	
	\begin{subfigure}[t]{0.1\textwidth}
	\hspace{-.3mm}
    \centering
    \includegraphics[width=1\linewidth]{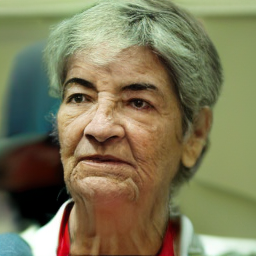}
	\end{subfigure}
	
	\\

\end{tabular}
	
\caption{\textbf{Cross-identity Reenactment.} Comparisons with the competing methods~\cite{siarohin2019first, hong2022depth, hong2023implicit}.}
\label{fig:reenactment_sm0}
\end{figure*}

\begin{figure*}[ht!]
	\centering
	\captionsetup[subfigure]{labelformat=empty,justification=centering,aboveskip=1pt,belowskip=1pt}
    
    \begin{tabular}[c]{c c}
    
        \vspace{-2.5mm}
    
        \begin{subfigure}[t]{0.15\textwidth}
            \centering
            \vspace{2.3cm}
            \small{Driving video}
        \end{subfigure} 
	
	&
	
	\begin{subfigure}[t]{0.8\textwidth}
            \hspace{-.3mm}
            \centering
            \includegraphics[width=1\linewidth]{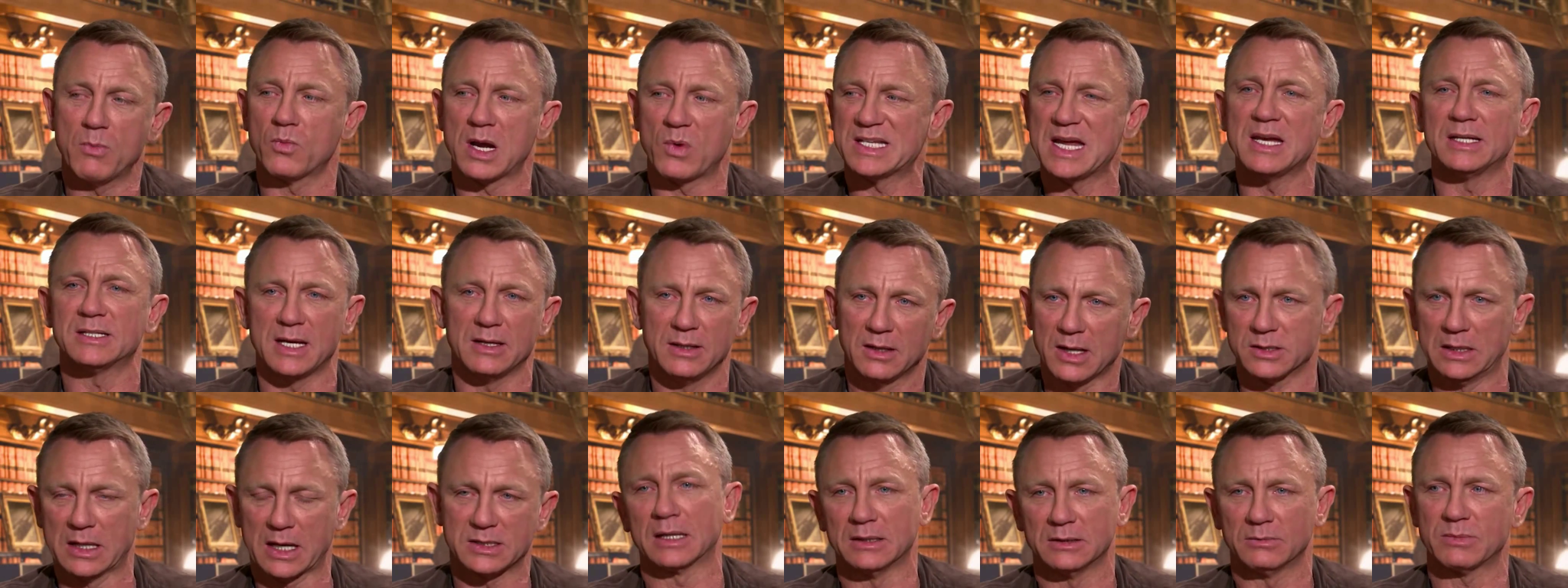}
	\end{subfigure}
	
	\\
    
        \vspace{-2.5mm}
    
        \begin{subfigure}[t]{0.1\textwidth}
            \hspace{-.3mm}
            \centering
            \includegraphics[width=1\linewidth]{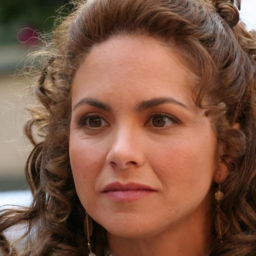}
	\end{subfigure}
	
	&
	
	\begin{subfigure}[t]{0.8\textwidth}
            \hspace{-.3mm}
            \centering
            \includegraphics[width=1\linewidth]{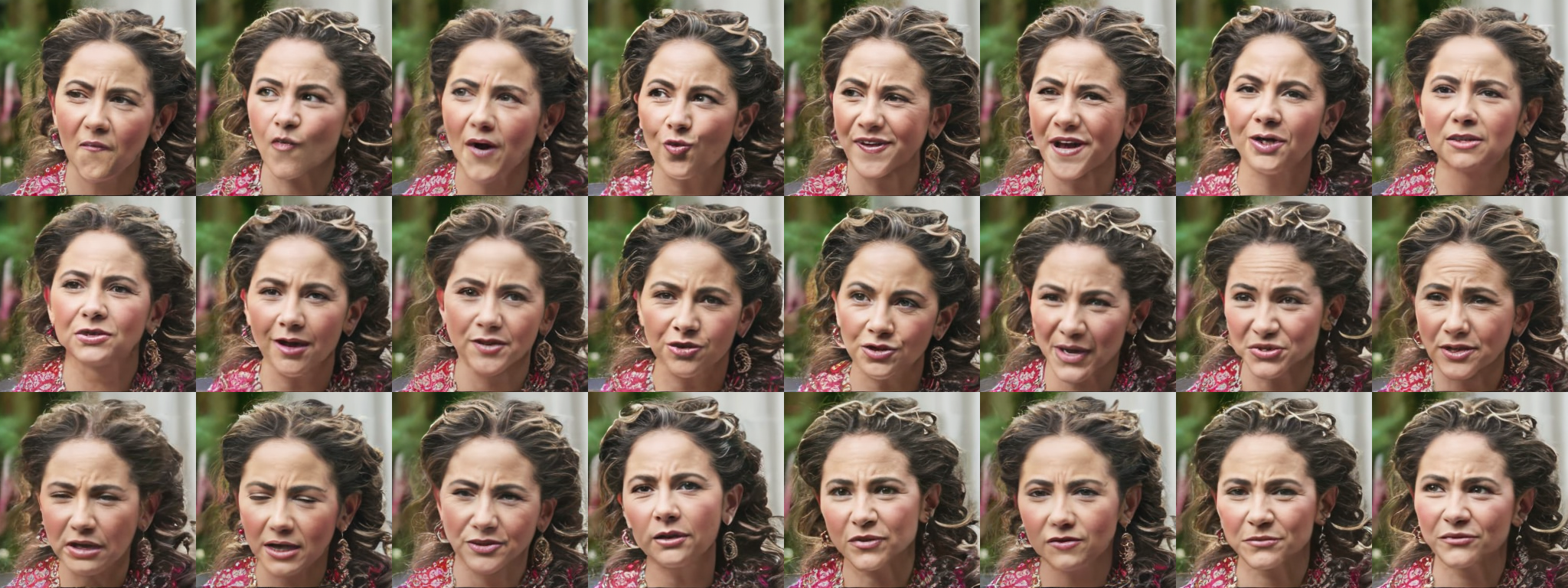}
	\end{subfigure}

        \\

        \vspace{-2.5mm}
    
        \begin{subfigure}[t]{0.15\textwidth}
            \centering
            \vspace{2.3cm}
            \small{Driving video}
        \end{subfigure} 
	
	&
	
	\begin{subfigure}[t]{0.8\textwidth}
            \hspace{-.3mm}
            \centering
            \includegraphics[width=1\linewidth]{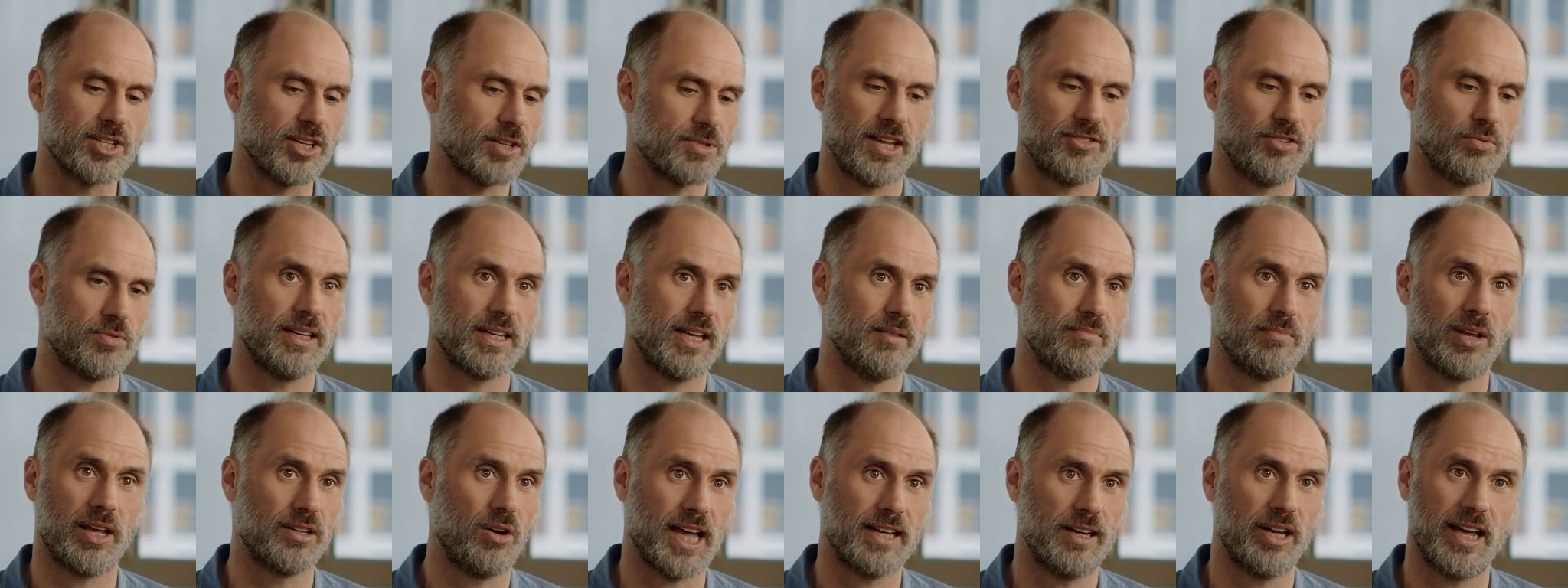}
	\end{subfigure}
	
	\\
    
        \vspace{-2.5mm}
    
        \begin{subfigure}[t]{0.1\textwidth}
            \hspace{-.3mm}
            \centering
            \includegraphics[width=1\linewidth]{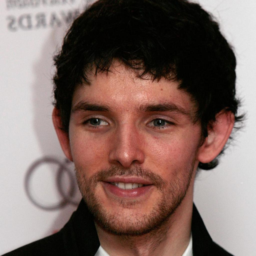}
	\end{subfigure}
	
	&
	
	\begin{subfigure}[t]{0.8\textwidth}
            \hspace{-.3mm}
            \centering
            \includegraphics[width=1\linewidth]{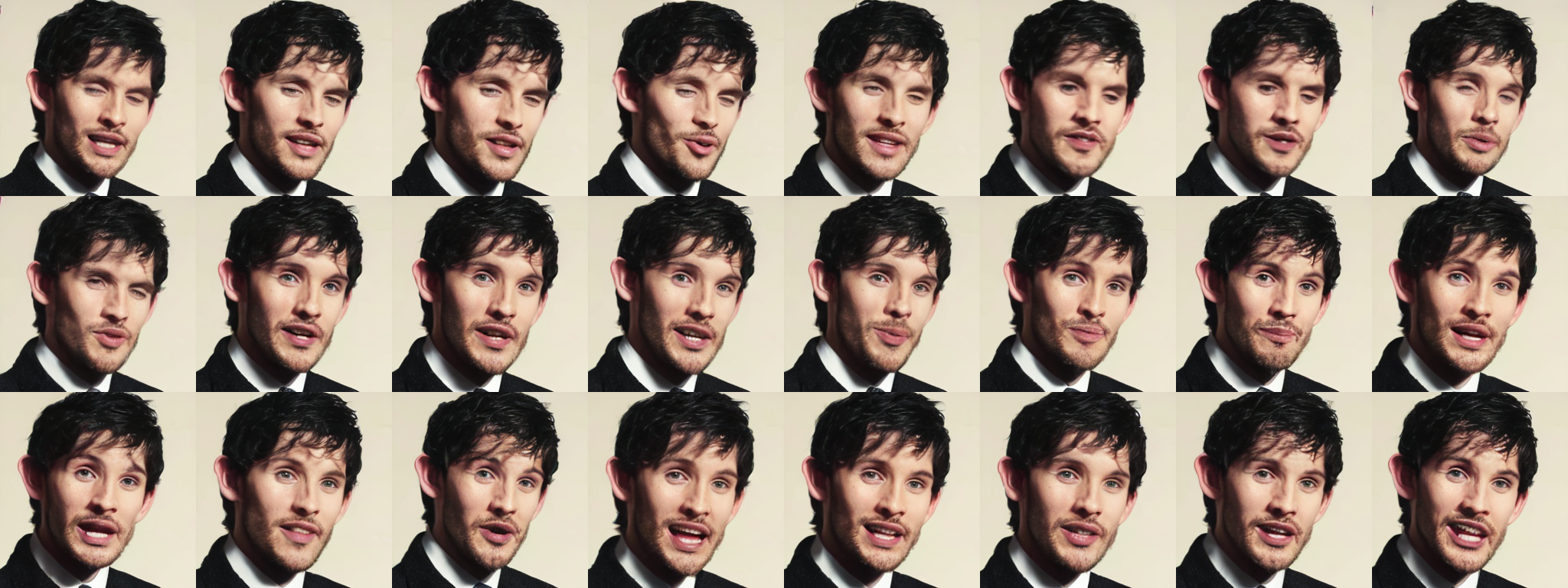}
	\end{subfigure}

	\end{tabular}
	
\caption{\textbf{Cross-identity Reenactment.} Enabling Identity-Swapping in 24-Frame Video Clips.}
\label{fig:long_c24}
\end{figure*}

To further illustrate the capabilities of our face reenactment approach, we present a broader range of visual results in Figures \ref{fig:reenactment_sm6}-\ref{fig:reenactment_sm0}. 
These examples highlight the model's ability to handle challenging conditions such as extreme poses and varying facial attributes, while maintaining visual fidelity and temporal consistency.
Additionally, Figure \ref{fig:long_c24} highlights the model's effectiveness in generating coherent and extended video sequences, further demonstrating its versatility and potential applications.

\section{Additional Experimental Details}
We train two base sDiT-XL models at a resolution of 256x256 pixels, each with a patch size of 2x2. These models are capable of generating sequences of $4$ and $8$ frames, respectively. The mapping network consists of $4$ residual blocks, and we use standard weight initialization techniques from ViT~\cite{dosovitskiy2020image}. All models are trained with AdamW~\cite{loshchilov2017decoupled}, using default parameter values and a cosine learning rate scheduler. The initial learning rates are set to $6.4 \times 10^{-5}$ for the denoiser and $6.4 \times 10^{-6}$ for the mapping network.

For the VAE model, we use a pre-trained model from Stable Diffusion~\cite{rombach2022high}. The VAE encoder downscales the spatial dimensions by a factor of x8 while producing a 4-channel output for a 3-channel RGB input. We retain diffusion hyperparameters from DiT~\cite{peebles2023scalable}, including $t_{max}=1000$ and a learned sigma routine.

Our training loss function is a weighted MSE, designed to prioritize accurate reconstruction of facial expressions, specifically targeting facial landmarks around the mouth and eyes. These expressive landmark pixels are assigned a weight of $(1 + \lambda_{\text{ex}})$, with $\lambda_{\text{ex}}$ set to 1, while other pixels are weighted at 1.

All models are trained for $1$ million steps using a global batch size of $16$ samples. We implement our models in Pytorch~\cite{paszke2019pytorch} and train them using four Nvidia A100-SXM4-80GB GPUs. The most compute-intensive model achieves a training speed of approximately 1.8 iterations per second.

\FloatBarrier
{\small
\bibliographystyle{ieee_fullname}
\bibliography{references}
}

\end{document}